\pdfoutput=1
\documentclass[11pt]{article}

\usepackage[preprint]{acl}

\usepackage{times}
\usepackage{latexsym}

\usepackage[T1]{fontenc}
\usepackage[utf8]{inputenc}

\usepackage{microtype}

\usepackage{inconsolata}

\usepackage{multirow} 
\usepackage{float}
\usepackage{amssymb}
\usepackage{graphics}
\usepackage{amsmath}
\usepackage{booktabs}
\usepackage{comment}
\usepackage{cleveref}
\usepackage{tabularx}
\usepackage{graphicx}

\usepackage{wrapfig}
\usepackage{caption}
\usepackage{subfigure}

\title{Exploring Large Language Models for Detecting Mental Disorders}

\author{\bf Gleb Kuzmin\textsuperscript{1,2,3} \qquad
  Petr Strepetov\textsuperscript{4} \qquad
  Maksim Stankevich\textsuperscript{3} \\
  \bf Natalia Chudova\textsuperscript{3} \qquad
  Artem Shelmanov\textsuperscript{6} \qquad
  Ivan Smirnov\textsuperscript{3,5} \\
\textsuperscript{1}AIRI ~~
\textsuperscript{2}ISP RAS Research Center for Trusted Artificial Intelligence \\
\textsuperscript{3}FRC CSC RAS ~~
\textsuperscript{4}MIPT ~~
\textsuperscript{5}RUDN University ~~
\textsuperscript{6}MBZUAI ~~
\\
\href{mailto:kuzmin@airi.net}{kuzmin@airi.net} ~~ \href{mailto:strepetov.pa@phystech.edu}{strepetov.pa@phystech.edu} ~~ \href{mailto:stankevich@isa.ru}{stankevich@isa.ru}\\
\href{mailto:nchudova@gmail.com}{nchudova@gmail.com} ~~
\href{mailto:artem.shelmanov@mbzuai.ac.ae}{artem.shelmanov@mbzuai.ac.ae} ~~
\href{mailto:ivs@isa.ru}{ivs@isa.ru}
}

\begin{document}
\maketitle

\begin{abstract}
This paper compares the effectiveness of traditional machine learning methods, encoder-based models, and large language models (LLMs) on the task of detecting depression and anxiety. Five Russian-language datasets were considered, each differing in format and in the method used to define the target pathology class. We tested AutoML models based on linguistic features, several variations of encoder-based Transformers such as BERT, and state-of-the-art LLMs as pathology classification models. The results demonstrated that LLMs outperform traditional methods, particularly on noisy and small datasets where training examples vary significantly in text length and genre. However, psycholinguistic features and encoder-based models can achieve performance comparable to language models when trained on texts from individuals with clinically confirmed depression, highlighting their potential effectiveness in targeted clinical applications.\footnote{\url{https://github.com/glkuzi/llm-mental-disorders-detection}}

\end{abstract}

\section{Introduction} \label{sec:introduction}

The problem of detecting mental disorders and patient emotions through text analysis and machine learning has been of increasing interest to researchers over the past decade~\citep{graham2019artificial, zhang2022natural, calixto2022natural, mayer2024predicting}. In fact, advances in data science and natural language processing methods offer promising opportunities for screening, monitoring, early detection, and prevention of negative outcomes of mental disorders. Although there are studies that work with interviews~\citep{morales2016speech, ringeval2017avec} and offline texts~\citep{lynn2018clpsych, stankevich2019predicting}, in most cases the material for such research comes from social media~\citep{guntuku2017detecting, garg2023mental}. These studies tend to focus on, but are not limited to, conditions such as depression, anxiety, stress, suicidality, post-traumatic stress disorder, and anorexia. Unsurprisingly, the methods considered for predicting mental state from text fell into traditional machine learning, using hand-crafted linguistic features, and various forms of deep learning~\citep{zhang2022natural}. The deep learning approach is often more accurate, especially when there are enough data samples, while traditional machine learning produces more interpretable results.

In this paper, we compare the performance of linguistic features, encoder-based models, and large language models (LLMs) on the task of identifying mental disorders. We consider two types of mental states, depression and anxiety, and several datasets in Russian that differ in text format and in the way a pathology is detected. To the best of our knowledge, this is the first comprehensive study on mental disorder detection in Russian texts, involving a wide range of techniques, including LLMs.

For our study, we involve a clinically verified dataset of essays, which contains texts written by patients with clinically diagnosed depression as well as texts from healthy volunteers~\citep{stankevich2019predicting}.
Most studies based on social media rely on self-reports, group affiliations, or questionnaire responses to determine mental health status. In contrast, only a few works use clinically validated data, which can differ substantially in quality and reliability~\citep{chancellor2020methods, ernala2019methodological}.

This paper addresses the following research questions:

\begin{itemize}
\item RQ1: What is the most prominent technique for predicting depression and anxiety: traditional machine learning, encoder-based models, or recent LLMs?
\item RQ2: Do models trained on a dataset of essays in which depression was defined by a clinical diagnosis generalize to social media, where the depression status is defined by a questionnaire?
\item RQ3: How do the LLM-generated explanations for detected depression align with the clinician's perspective?
\end{itemize}

Our main contributions are the following:
\begin{itemize}
    \item We outperformed the existing state-of-the-art depression detection method on one dataset and established classification baselines on three previously unexamined anxiety datasets.
    \item We conducted a thorough comparison of various groups of models on the depression and anxiety detection tasks in Russian, which could be used by practitioners in this field for future experiments.
    \item We explored the transferability of models from tasks using clinical diagnoses as targets to those based on questionnaire-derived labels, aiming to mitigate the scarcity of clinically validated data in mental disorder detection.
    \item We evaluated and categorized LLM-generated explanations for detected depression from the point of view of clinicians, which could be used for future improvement of LLM-assisted systems in this field.
\end{itemize}

\section{Related Work} \label{sec:relatedwork}
\subsection{Traditional and Advanced ML Methods}

Researchers have employed various methods for detecting depression and anxiety across social media platforms. \citet{tadesse2019depression} consider the Reddit users' dataset, comparing single and combined feature learning for depression detection. N-gram features, Linguistic Inquiry and Word Count (LIWC) dictionary features, and topics from Latent Dirichlet Allocation are considered, showcasing the effectiveness of combined features on the multi-layer perceptron.

In~\citep{shan2020depression}, NLP methods are applied for depression detection of Reddit users based on their posts. GloVe, Word2Vec, and FastText embeddings, as well as handcrafted statistical metadata features and LIWC features, are used for text representation. The two-headed model, combining BiLSTM for embeddings and a fully connected layer for meta-features, demonstrates superior results with Word2Vec embeddings and meta-features. Additionally, the authors use Early Risk Detection Error and Latency metrics to take the time of classification into account.

\citet{owen-etal-2020-towards} consider depression and anxiety detection for tweets, using SVM on TF-IDF vectors and GloVe embeddings, as well as BERT~\citep{devlin-etal-2019-bert} and ALBERT~\citep{lan2019albert}. Results show that BERT is better on a balanced dataset, while SVM excels on an unbalanced one.

\citet{Babu2021SentimentAI} underscores the importance of emoticons in texts in sentiment analysis for depression detection. The study covers 101 publications, emphasizing the effectiveness of combining deep learning algorithms, with CNN$+$LSTM yielding the highest precision. Moreover, multi-class sentiment classification provides more precise results than binary and ternary classifications.

The FCL (Fasttext$+$CNN$+$LSTM) model proposed in~\citep{Tejaswini2022DepressionDF} outperforms LSTM and CNN models, based on GloVe and Word2vec embeddings, on datasets for depression detection with Reddit and Twitter posts.

Assessing the anxious Twitter posts caused by the COVID-19 pandemic,~\citep{jeong2023reading} uses BERT trained on the Korean language, achieving strong accuracy and establishing a correlation between the anxiety index and COVID-19 waves.

The study~\citep{Ansari2023EnsembleHL} focuses on identifying depression in social media datasets (CLPsych, Reddit, eRisk) through various text classification methods, combining sentiment lexicons with deep learning pipelines. Authors utilize sentiment lexicons with logistic regression and LSTM with attention on GloVe embeddings, comparing them and combining them into ensembles.

\subsection{Large Language Models}

According to the systematic review~\citep{OMAR2025234}, most studies about depression detection focus on BERT-based models, indicating the field's early stages in adopting newer technologies like GPT-4 and Google's Gemini. However, LLMs are demonstrating significant potential in advancing depression detection systems.

The Chat-Diagnose approach~\citep{qin2023read} integrates diagnostic criteria from the Diagnostic and Statistical Manual of Mental Disorders (DSM-5) into prompts and uses the Chain of Thoughts technique to deliver explainable diagnoses via an LLMs-augmented system based on ChatGPT/GPT-3. This method demonstrates state-of-the-art results on Twitter and Weibo depression datasets by employing zero-shot and few-shot learning.

Another study~\citep{doi:10.1080/18824889.2024.2342624} compares the effectiveness of fine-tuned BERT with GPT-3.5 and GPT-4 in the depression detection task. The authors use Patient Health Questionnaire-8 scores for classifying transcribed audio data from the Distress Analysis Interview Corpus, KID, and a simulated dataset. With scores separated into depressive and non-depressive groups, the zero-shot method for GPT-4 outperforms GPT-3.5 and BERT across all datasets, highlighting the potential of LLMs in depression detection.

Additionally, \citet{wang-etal-2024-explainable} investigates depression symptom detection and severity classification using LLMs on the eRisk 2021 and eRisk 2023 datasets. Utilizing Beck’s Depression Inventory to form queries related to depression symptoms and the Universal Sentence Encoder for text embeddings, the study creates two datasets containing top-1 and top-5 ranked texts for each query. LLMs fine-tuned with QLoRA are used for classification into four levels of depression severity.

The DORIS~\citep{lan2024depression} system addresses the challenges of detecting depression through social media posts from the Sina Weibo Depression Dataset. The authors use GPT3.5-Turbo-1103 for annotating high-risk texts according to the DSM-5 depression scale; also, LLM is used to summarize critical information from users' historical mood records (mood courses). The final model based on XGBoost is learned on features from annotations and gte-small-zh model vector representations of post histories and mood courses and shows an improvement over the baseline.

We are the first to examine and compare three generations of the discussed models for depression and anxiety detection tasks in Russian, namely, traditional ML models, encoder-decoder models, and LLMs. Unlike other works, we used various models from each group and carefully compared the results of the models between the groups on five datasets, aiming for a general recommendation on the best models to use in practice.

\section{Data} \label{sec:data}

This paper considers five Russian-language datasets: 2 for depression and 3 for anxiety. Classes in all datasets were represented in the binary format: a healthy class (no signs of mental disorders) and a pathology class (depression or anxiety). The general description of the datasets used in our study is shown in Table~\ref{tab:dataset_descriptions}.
\begin{table*}
\centering
\resizebox{\linewidth}{!}{%
    % \begin{tabular}{l|lllll}
    % \toprule
    % Reference &
    % \citep{stankevich2019predicting} &
    % \citep{ignatiev2022predicting} &
    % \multicolumn{2}{l}{\citep{litvinova2018rusneuropsych}} 
    % &
    % \citep{medvedeva2021lexical} \\
    % Label & DE & DSM & AL & AD & AC \\
    % Condition                                                         & Depression         & Depression            & Anxiety & Anxiety           & Anxiety        \\
    % Text format                                                       & Essay              & Social media messages & Letter  & Description of the picture & Short comments \\
    % Class criteria                                                    & Clinical diagnosis & BDI & HADS                   & HADS    & SCL-90-R       \\
    % \# healthy samples                                             & 447                & 135                   & 109     & 101               & 222            \\
    % \# pathology samples                                           & 110                & 89                    & 93      & 89                & 191            \\
    % \begin{tabular}[c]{@{}l@{}}Median words\\ per sample\end{tabular} & 309                  & 122                     & 144       & 84                 & 7     \\
    % \bottomrule
    % \end{tabular}

    \begin{tabular}{l|l|l|l|l|l}
    \toprule
    \textbf{Name} & Depression-Essays (DE) & Depression-Social Media (DSM) & Anxiety-Letter (AL) & Anxiety-Description (AD) & Anxiety-COVID Comments (AC) \\
    \textbf{Reference} &
    \citep{stankevich2019predicting} &
    \citep{ignatiev2022predicting} &
    \multicolumn{2}{l|}{\citep{litvinova2018rusneuropsych}} 
    &
    \citep{medvedeva2021lexical} \\
    \textbf{Condition} & Depression & Depression & Anxiety & Anxiety & Anxiety \\
    \textbf{Text format} & Essay & Social media messages & Letter & Description of the picture & Short comments \\
    \textbf{Class criteria} & Clinical diagnosis & BDI & HADS & HADS & SCL-90-R \\
    \textbf{\# healthy samples} & 447 & 135 & 109 & 101 & 222 \\
    \textbf{\# pathology samples} & 110 & 89 & 93 & 89 & 191 \\
    \textbf{\begin{tabular}[c]{@{}l@{}}Median words\\ per sample\end{tabular}} & 309 & 122 & 144 & 84 & 7 \\
    \bottomrule
\end{tabular}
    
}
\caption{Summary of datasets used in this study
\vspace{-0.3cm}
}
\label{tab:dataset_descriptions}
\end{table*}

\subsection{Depression-Essays (DE)} \label{subsec:DE}
To compile this dataset, subjects were asked to write a short essay (from 1,500 to 5,000 characters with spaces and punctuation) on the topic of ``Myself, others, world''~\citep{stankevich2019predicting}. 	In total, 557 essays were collected, including 110 authored by patients with clinical depression. The essays written by people with clinically validated depression were provided by the Mental Health Research Center, Moscow, Russia. The collection of essays was done on a voluntary basis, under conditions of anonymity and for research purposes only. 
The best-reported performance on this data reaches 73\% F1 for the depression class in the cross-validation evaluation with the random forest model trained on n-grams and psycholinguistic features~\citep{stankevich2019predicting}. For this dataset, we provided several anonymized examples in \Cref{tab:de_examples} in \Cref{app:de_examples}.

\subsection{Depression-Social Media (DSM)} \label{subsec:DSM}
The Depression-Social Media dataset was developed to support research on detecting depression in social network users~\citep{ignatiev2022predicting}. It consists of text messages from the VKontakte platform, accompanied by results from the Russian-language adaptation of the 21-item Beck Depression Inventory (BDI) questionnaire~\citep{beck1961inventory}. The healthy class included users with a scale score of 10 or less, and the pathological class between 30 and 63. The best-reported classification performance using textual data reaches only 65\% F1 for the depression class with a logistic regression model trained on psycholinguistic features~\citep{ignatiev2022predicting}.

We have made some changes to the original dataset. For each subject, all messages were combined for a period of 170 days prior to the date of the questionnaire screening, and the total text was limited to 6,000 characters (symbols), resulting in a median length per sample of 122 words. Such restrictions were imposed to bring the final texts from this dataset closer to the format of the essay dataset in terms of total text length for each subject and to account for the fact that the relevance of depression screening becomes less over a period of more than six months. 

\subsection{Anxiety-Letter and Anxiety Picture Description (AL and AD)} \label{subsec:AL_AD}
For anxiety detection experiments, we use the RusNeuroPsych corpus \citep{litvinova2018rusneuropsych}. This corpus was prepared to study the relationships between a person's text, personal traits, mental status, and demographic characteristics. To compile the corpus, participants were asked to write an informal letter to a friend, provide a textual description of a picture, and complete a series of psychological questionnaires. Among these, the Hospital Anxiety and Depression Scale (HADS)~\citep{bjelland2002validity} was employed to assess anxiety levels. Based on HADS scores, subjects with a score of 7 or below were assigned to the healthy class, while those with a score of 8 or higher were assigned to the pathology class. Because the corpus contains two distinct text types, it was further split into two separate datasets for the experiments.
We denote them AL (Anxiety-Letter) for letters and AD (Anxiety-Description) for picture descriptions. To the best of our knowledge, no classification experiments have been performed on this data before.

\subsection{Anxiety-COVID Comments (AC)} \label{subsec:AC}

In addition, we used a corpus of subjects' comments on the COVID-19 pandemic situation~\citep{medvedeva2021lexical}. Subjects were asked to complete a series of questionnaires and write a free-form commentary describing their attitudes towards the world situation around the pandemic and self-isolation. Among the questionnaires used was the SCL-90-R symptom questionnaire~\citep{derogatis1983scl}, the anxiety scale from which was used to form 2 groups: the healthy group, with an anxiety scale score below the 33rd percentile, and the pathology group, with an anxiety scale score above the 66th percentile. 
To the best of our knowledge, no classification experiments have been performed on this data before.

\section{Methods} \label{sec:method}
\subsection{Linguistic Features} \label{subsec:linguistic_features}

\subsubsection{Psycholinguistic Features} 
\label{subsubsec:psycholinguistic_features}

The linguistic features used in our research were extracted using the tool described in~\citep{smirnov2021titanis}. This tool extracts morphological, syntactic, and vocabulary parameters of texts, including various psycholinguistic coefficients. A total of 113 features were used. A detailed description of the features utilized and those extracted with this tool on our data is available in the Hugging Face repository.\footnote{\url{https://huggingface.co/datasets/anonymizedauthor/paper_data}}

\subsubsection{Classification Setup} \label{subsubsec:ngram_and_psy_features_classification}
As a classification baseline, we use the AutoML system auto-sklearn \citep{feurer-neurips15a} trained on psycholinguistic features and n-grams. The auto-sklearn classifier employs a Bayesian optimizer that considers 15 classification algorithms, 14 feature preprocessing methods, and 4 data preprocessing techniques. Additionally, the classifier utilizes a meta-learning approach and automated ensemble construction to speed up optimization. 

For psycholinguistic features, we consider several feature selection methods: filter method, wrapper method (forward selection and backward elimination), and embedding method. Selected features are examined for pairwise linear correlation, and those with absolute Pearson coefficient values of more than 0.95 are deleted.

We also train models on TF-IDF vectors on unigrams and unigrams with bigrams. In addition to full vectors, we consider different subsets of features for vectors on unigrams: from 20\% to 100\% of the features are selected with a 20\% increment. For each subset, correlated features are deleted in the same way as with psycholinguistic features.

On each set of features, we launch the classifier 6 times with different seeds and then calculate the mean and standard deviation values of the metrics based on the results of the launches. The chosen AutoML models are presented in \Cref{app:hyperpars}.

\subsection{Encoder-Based Models} \label{subsec:enc_models}

Encoder-based classification models eliminate the need for tedious feature engineering in favor of a deep multi-layer neural architecture.
As we target classification on Russian corpora, we consider models pretrained on multilingual or Russian datasets. As a simple baseline, we used a base version of multilingual BERT, a model with 110 million parameters, first introduced in \citep{devlin-etal-2019-bert}. Another considered baseline is RuBERT \citep{kuratov2019adaptation}, a pretrained Russian version of BERT-base with the same amount of parameters. We also finetuned several more recent models, such as RuBioRoBERTa \citep{yalunin2022rubioroberta}, which is a RoBERTa pretrained on Russian language biomedical texts, and RuRoBERTa-large -- a bigger version of RoBERTa, also pretrained on Russian language datasets.
For all of the four models, we optimized hyperparameters using Bayesian optimization from the HuggingFace framework \citep{wolf-etal-2020-transformers}.

The used hyperparameter grid and optimal parameters, along with the used checkpoint of models, are presented in \Cref{app:hyperpars}.

\subsection{Large Language Models} \label{subsec:llm}
We conducted experiments with LLMs in various settings. First of all, we evaluated models by 0-shot and 5-shot prompting, considering only normalized probabilities for tokens ``0'' or ``1'' in the first generated token, as it was done in MMLU \citep{hendryckstest2021}. We will refer to these settings as ``0-shot MMLU'' and ``5-shot MMLU'' correspondingly. Secondly, we employed 0-shot and 5-shot prompting with bigger generation lengths and matched the generated answer to one of the possible classes with string matching. Finally, we conducted the fine-tuning of the models using LoRA \citep{hu2022lora}.

We selected a set of relatively small (less than 9B parameters) self-hosted open-source models, either fine-tuned for the Russian language or showing good multilingual capabilities. To the first group belongs SaigaLlama3 8B, a version of Llama 3 8B Instruct \citep{dubey2024llama} fine-tuned on several Russian datasets, as well as models from the Vikhr family \citep{nikolich2024vikhr}. We used Vikhr 7B Instruct 0.4, Vikhr 7B Instruct 5.4, and Vikhr Gemma 2B Instruct. The former two are based on Mistral 7B \citep{jiang2023mistral} with vocabulary adaptation for the Russian language, followed by additional pretraining and instruction tuning. The latter one is based on Gemma2 2B Instruct \citep{team2024gemma}, additionally trained on Russian data. We also used multilingual models, such as Gemma2 2B Instruct and Gemma2 9B Instruct, as well as Qwen2 7B Instruct \citep{yang2024qwen2}. As we are conducting experiments on sensitive data and with private datasets, we did not consider remotely-hosted models (such as GPT-4, Claude) due to the possibility of data leakage.

The full information about used prompts, training hyperparameters, and versions of used models is presented in \Cref{app:hyperpars}.

\section{Results} \label{sec:results}

\subsection{Classification Results on Depression and Anxiety Datasets} \label{subsec:results_table_2}
The data were divided into training and test samples in an 80\% by 20\% ratio, with stratification by the target variable reduced to binary form. All classification reports in this study show results on the test data. The classification report for the best models from each group is presented in \Cref{tab:results_main_best_exp}.

\begin{table*}[t]
\centering
\resizebox{\linewidth}{!}{%
    \begin{tabular}{lllccccccc}
    \toprule
    % Corpus & Mode & Model & \begin{tabular}[c]{@{}l@{}} Precision \\ healthy\end{tabular} & \begin{tabular}[c]{@{}l@{}} Recall \\ healthy\end{tabular} & F1-healthy & \begin{tabular}[c]{@{}l@{}} Precision \\ pathology\end{tabular} & \begin{tabular}[c]{@{}l@{}} Recall \\ pathology\end{tabular} & F1-pathology & F1-macro  \\
    \textbf{Corpus} & 
\textbf{Mode} & 
\textbf{Model} & 
\textbf{\begin{tabular}[c]{@{}l@{}} Precision \\ healthy\end{tabular}} & 
\textbf{\begin{tabular}[c]{@{}l@{}} Recall \\ healthy\end{tabular}} & 
\textbf{F1-healthy} & 
\textbf{\begin{tabular}[c]{@{}l@{}} Precision \\ pathology\end{tabular}} & 
\textbf{\begin{tabular}[c]{@{}l@{}} Recall \\ pathology\end{tabular}} & 
\textbf{F1-pathology} & 
\textbf{F1-macro} \\
    \hline
    \multirow{7}{*}{DE} & SFT & Linguistic features & 94.00\tiny{$\pm$ 0.80} & 95.20\tiny{$\pm$ 1.00} & 94.60\tiny{$\pm$ 0.80} & 79.30\tiny{$\pm$ 4.00} & 75.00\tiny{$\pm$ 3.50} & 77.00\tiny{$\pm$ 3.30} & 85.80\tiny{$\pm$ 2.10}  \\ 
    ~ & SFT & TF-IDF & 91.60\tiny{$\pm$ 2.10} & 94.40\tiny{$\pm$ 2.20} & 93.00\tiny{$\pm$ 1.20} & 74.80\tiny{$\pm$ 7.00} & 64.40\tiny{$\pm$ 10.30} & 68.50\tiny{$\pm$ 6.60} & 80.70\tiny{$\pm$ 3.80}  \\
    ~ & 5-shot & Vikhr 7B IT 5.4 & 87.06\tiny{$\pm$ 0.00} & 82.22\tiny{$\pm$ 0.00} & 84.57\tiny{$\pm$ 0.00} & 40.74\tiny{$\pm$ 0.00} & 50.00\tiny{$\pm$ 0.00} & 44.90\tiny{$\pm$ 0.00} & 64.73\tiny{$\pm$ 0.00} \\
~ & 5-shot MMLU & Gemma2 9B IT & 93.11\tiny{$\pm$ 0.08} & 75.11\tiny{$\pm$ 0.89} & 83.15\tiny{$\pm$ 0.58} & 43.16\tiny{$\pm$ 0.85} & 77.27\tiny{$\pm$ 0.00} & 55.38\tiny{$\pm$ 0.71} & 69.26\tiny{$\pm$ 0.64} \\
~ & LoRA & VikhrGemma 2B IT & 94.74\tiny{$\pm$ 0.88} & 96.48\tiny{$\pm$ 1.62} & 95.59\tiny{$\pm$ 0.66} & 84.96\tiny{$\pm$ 5.72} & 78.03\tiny{$\pm$ 4.08} & \textbf{81.13\tiny{$\pm$ 2.42}} & \textbf{88.36\tiny{$\pm$ 1.52}} \\
~ & SFT & RuBERT & 94.45\tiny{$\pm$ 1.16} & 91.11\tiny{$\pm$ 1.70} & 92.74\tiny{$\pm$ 1.04} & 68.41\tiny{$\pm$ 4.04} & 78.03\tiny{$\pm$ 4.85} & 72.80\tiny{$\pm$ 3.43} & 82.77\tiny{$\pm$ 2.21} \\
~ & 0-shot & Gemma2 9B IT & 93.33\tiny{$\pm$ 0.00} & 62.22\tiny{$\pm$ 0.00} & 74.67\tiny{$\pm$ 0.00} & 34.62\tiny{$\pm$ 0.00} & 81.82\tiny{$\pm$ 0.00} & 48.65\tiny{$\pm$ 0.00} & 61.66\tiny{$\pm$ 0.00} \\
    \hline
    \multirow{8}{*}{DSM} & SFT & Linguistic features & 62.80\tiny{$\pm$ 1.80} & 59.30\tiny{$\pm$ 7.40} & 60.70\tiny{$\pm$ 4.00} & 43.70\tiny{$\pm$ 2.70} & 47.20\tiny{$\pm$ 7.70} & 45.10\tiny{$\pm$ 3.80} & 52.90\tiny{$\pm$ 2.00}  \\ 
    ~ & SFT & TF-IDF & 62.20\tiny{$\pm$ 2.60} & 51.20\tiny{$\pm$ 13.90} & 55.30\tiny{$\pm$ 8.50} & 42.90\tiny{$\pm$ 3.60} & 53.70\tiny{$\pm$ 11.40} & 46.90\tiny{$\pm$ 4.70} & 51.10\tiny{$\pm$ 3.50}  \\
    ~ & 5-shot & SaigaLlama3 8B & 100.00\tiny{$\pm$ 0.00} & 3.70\tiny{$\pm$ 0.00} & 7.14\tiny{$\pm$ 0.00} & 40.91\tiny{$\pm$ 0.00} & 100.00\tiny{$\pm$ 0.00} & \textbf{58.06\tiny{$\pm$ 0.00}} & 32.60\tiny{$\pm$ 0.00} \\
~ & 5-shot & Vikhr 7B IT 0.4 & 69.19\tiny{$\pm$ 0.89} & 81.48\tiny{$\pm$ 0.00} & 74.83\tiny{$\pm$ 0.51} & 62.09\tiny{$\pm$ 1.10} & 45.56\tiny{$\pm$ 2.22} & 52.54\tiny{$\pm$ 1.85} & \textbf{63.69\tiny{$\pm$ 1.18}} \\
~ & 5-shot MMLU & SaigaLlama3 8B & 100.00\tiny{$\pm$ 0.00} & 3.70\tiny{$\pm$ 0.00} & 7.14\tiny{$\pm$ 0.00} & 40.91\tiny{$\pm$ 0.00} & 100.00\tiny{$\pm$ 0.00} & \textbf{58.06\tiny{$\pm$ 0.00}} & 32.60\tiny{$\pm$ 0.00} \\
~ & 5-shot MMLU & Vikhr 7B IT 5.4 & 68.75\tiny{$\pm$ 0.00} & 81.48\tiny{$\pm$ 0.00} & 74.58\tiny{$\pm$ 0.00} & 61.54\tiny{$\pm$ 0.00} & 44.44\tiny{$\pm$ 0.00} & 51.61\tiny{$\pm$ 0.00} & 63.09\tiny{$\pm$ 0.00} \\
~ & LoRA & Qwen2 7B IT & 68.87\tiny{$\pm$ 6.39} & 72.22\tiny{$\pm$ 8.21} & 70.40\tiny{$\pm$ 6.75} & 55.33\tiny{$\pm$ 11.08} & 50.93\tiny{$\pm$ 10.84} & 52.81\tiny{$\pm$ 10.26} & 61.61\tiny{$\pm$ 8.32} \\
~ & SFT & RuBioRoBERTa & 62.10\tiny{$\pm$ 2.98} & 62.96\tiny{$\pm$ 6.05} & 62.46\tiny{$\pm$ 4.30} & 43.66\tiny{$\pm$ 4.43} & 42.59\tiny{$\pm$ 4.14} & 43.01\tiny{$\pm$ 3.70} & 52.74\tiny{$\pm$ 3.67} \\
    \hline
    \multirow{8}{*}{AL} & SFT & Linguistic features & 47.40\tiny{$\pm$ 21.40} & 36.40\tiny{$\pm$ 17.80} & 40.90\tiny{$\pm$ 19.20} & 48.50\tiny{$\pm$ 2.70} & 68.40\tiny{$\pm$ 14.60} & 56.10\tiny{$\pm$ 3.70} & 48.50\tiny{$\pm$ 8.20}  \\ 
    ~ & SFT & TF-IDF & 61.50\tiny{$\pm$ 5.00} & 46.20\tiny{$\pm$ 3.10} & 52.60\tiny{$\pm$ 2.30} & 51.20\tiny{$\pm$ 2.50} & 65.80\tiny{$\pm$ 7.90} & 57.50\tiny{$\pm$ 4.50} & 55.00\tiny{$\pm$ 2.90}  \\ 
    ~ & 5-shot & Gemma2 9B IT & 75.00\tiny{$\pm$ 0.00} & 27.27\tiny{$\pm$ 0.00} & 40.00\tiny{$\pm$ 0.00} & 51.52\tiny{$\pm$ 0.00} & 89.47\tiny{$\pm$ 0.00} & \textbf{65.38\tiny{$\pm$ 0.00}} & 52.69\tiny{$\pm$ 0.00} \\
~ & LoRA & Gemma2 2B IT & 60.18\tiny{$\pm$ 4.51} & 60.61\tiny{$\pm$ 13.55} & 59.66\tiny{$\pm$ 8.17} & 54.87\tiny{$\pm$ 7.30} & 53.51\tiny{$\pm$ 10.71} & 53.45\tiny{$\pm$ 6.06} & 56.56\tiny{$\pm$ 5.40} \\
~ & SFT & RuRoBERTa & 56.82\tiny{$\pm$ 5.65} & 75.00\tiny{$\pm$ 7.76} & 64.49\tiny{$\pm$ 5.59} & 52.54\tiny{$\pm$ 10.76} & 33.33\tiny{$\pm$ 12.77} & 40.17\tiny{$\pm$ 12.48} & 52.33\tiny{$\pm$ 8.31} \\
~ & 0-shot & Gemma2 9B IT & 66.67\tiny{$\pm$ 0.00} & 63.64\tiny{$\pm$ 0.00} & 65.12\tiny{$\pm$ 0.00} & 60.00\tiny{$\pm$ 0.00} & 63.16\tiny{$\pm$ 0.00} & 61.54\tiny{$\pm$ 0.00} & \textbf{63.33\tiny{$\pm$ 0.00}} \\
~ & 0-shot MMLU & Gemma2 2B IT & 0.00\tiny{$\pm$ 0.00} & 0.00\tiny{$\pm$ 0.00} & 0.00\tiny{$\pm$ 0.00} & 46.34\tiny{$\pm$ 0.00} & 100.00\tiny{$\pm$ 0.00} & 63.33\tiny{$\pm$ 0.00} & 31.67\tiny{$\pm$ 0.00} \\
~ & 0-shot MMLU & Qwen2 7B IT & 61.03\tiny{$\pm$ 0.52} & 85.45\tiny{$\pm$ 1.82} & 71.20\tiny{$\pm$ 0.99} & 68.73\tiny{$\pm$ 2.55} & 36.84\tiny{$\pm$ 0.00} & 47.95\tiny{$\pm$ 0.64} & 59.58\tiny{$\pm$ 0.82} \\
    \hline 
    \multirow{8}{*}{AD} & SFT & Linguistic features & 50.80\tiny{$\pm$ 3.30} & 43.30\tiny{$\pm$ 12.50} & 45.30\tiny{$\pm$ 8.10} & 44.70\tiny{$\pm$ 2.60} & 51.90\tiny{$\pm$ 15.90} & 47.30\tiny{$\pm$ 7.40} & 46.30\tiny{$\pm$ 1.80}  \\ 
    ~ & SFT & TF-IDF & 54.90\tiny{$\pm$ 15.10} & 37.50\tiny{$\pm$ 4.80} & 43.20\tiny{$\pm$ 2.50} & 44.60\tiny{$\pm$ 7.80} & 59.30\tiny{$\pm$ 21.00} & 50.40\tiny{$\pm$ 12.70} & 46.80\tiny{$\pm$ 7.10}\\ 
    ~ & LoRA & Vikhr 7B IT 0.4 & 59.81\tiny{$\pm$ 4.42} & 51.67\tiny{$\pm$ 12.13} & 54.66\tiny{$\pm$ 7.59} & 53.47\tiny{$\pm$ 4.23} & 61.11\tiny{$\pm$ 11.56} & 56.43\tiny{$\pm$ 6.19} & 55.55\tiny{$\pm$ 4.55} \\
~ & SFT & RuRoBERTa & 57.63\tiny{$\pm$ 2.52} & 56.67\tiny{$\pm$ 4.71} & 57.07\tiny{$\pm$ 3.19} & 52.80\tiny{$\pm$ 2.82} & 53.70\tiny{$\pm$ 4.14} & 53.17\tiny{$\pm$ 2.84} & 55.12\tiny{$\pm$ 2.58} \\
~ & 0-shot & Vikhr 7B IT 5.4 & 83.33\tiny{$\pm$ 0.00} & 25.00\tiny{$\pm$ 0.00} & 38.46\tiny{$\pm$ 0.00} & 53.12\tiny{$\pm$ 0.00} & 94.44\tiny{$\pm$ 0.00} & \textbf{68.00\tiny{$\pm$ 0.00}} & 53.23\tiny{$\pm$ 0.00} \\
~ & 0-shot & Qwen2 7B IT & 67.25\tiny{$\pm$ 1.16} & 80.00\tiny{$\pm$ 0.00} & 73.07\tiny{$\pm$ 0.68} & 71.81\tiny{$\pm$ 0.76} & 56.67\tiny{$\pm$ 2.22} & 63.33\tiny{$\pm$ 1.67} & 68.20\tiny{$\pm$ 1.17} \\
~ & 0-shot MMLU & Vikhr 7B IT 5.4 & 83.33\tiny{$\pm$ 0.00} & 25.00\tiny{$\pm$ 0.00} & 38.46\tiny{$\pm$ 0.00} & 53.12\tiny{$\pm$ 0.00} & 94.44\tiny{$\pm$ 0.00} & \textbf{68.00\tiny{$\pm$ 0.00}} & 53.23\tiny{$\pm$ 0.00} \\
~ & 0-shot MMLU & Qwen2 7B IT & 70.74\tiny{$\pm$ 1.47} & 70.00\tiny{$\pm$ 0.00} & 70.36\tiny{$\pm$ 0.72} & 67.02\tiny{$\pm$ 0.70} & 67.78\tiny{$\pm$ 2.22} & 67.39\tiny{$\pm$ 1.44} & \textbf{68.87\tiny{$\pm$ 1.08}} \\
    \hline 
    \multirow{8}{*}{AC} & SFT & Linguistic features & 64.80\tiny{$\pm$ 16.00} & 45.60\tiny{$\pm$ 21.00} & 47.00\tiny{$\pm$ 19.60} & 48.70\tiny{$\pm$ 3.60} & 60.50\tiny{$\pm$ 20.10} & 52.60\tiny{$\pm$ 7.30} & 49.80\tiny{$\pm$ 7.80}  \\ 
    ~ & SFT & TF-IDF & 56.30\tiny{$\pm$ 2.10} & 63.30\tiny{$\pm$ 4.20} & 59.50\tiny{$\pm$ 2.50} & 48.90\tiny{$\pm$ 3.30} & 41.70\tiny{$\pm$ 5.60} & 44.90\tiny{$\pm$ 4.30} & 52.20\tiny{$\pm$ 2.70}  \\
    ~ & 5-shot & Qwen2 7B IT & 84.89\tiny{$\pm$ 3.56} & 26.22\tiny{$\pm$ 5.33} & 39.85\tiny{$\pm$ 6.96} & 52.08\tiny{$\pm$ 1.72} & 94.74\tiny{$\pm$ 0.00} & \textbf{67.20\tiny{$\pm$ 1.46}} & 53.52\tiny{$\pm$ 4.21} \\
~ & 5-shot MMLU & SaigaLlama3 8B & 100.00\tiny{$\pm$ 0.00} & 13.33\tiny{$\pm$ 0.00} & 23.53\tiny{$\pm$ 0.00} & 49.35\tiny{$\pm$ 0.00} & 100.00\tiny{$\pm$ 0.00} & 66.09\tiny{$\pm$ 0.00} & 44.81\tiny{$\pm$ 0.00} \\
~ & LoRA & Vikhr 7B IT 5.4 & 61.87\tiny{$\pm$ 2.07} & 62.96\tiny{$\pm$ 8.18} & 62.13\tiny{$\pm$ 4.30} & 55.50\tiny{$\pm$ 3.03} & 53.95\tiny{$\pm$ 7.25} & 54.33\tiny{$\pm$ 3.58} & \textbf{58.23\tiny{$\pm$ 2.38}} \\
~ & SFT & BERT & 57.01\tiny{$\pm$ 5.15} & 62.96\tiny{$\pm$ 16.81} & 58.44\tiny{$\pm$ 6.53} & 46.07\tiny{$\pm$ 8.77} & 41.67\tiny{$\pm$ 22.14} & 41.28\tiny{$\pm$ 17.15} & 49.86\tiny{$\pm$ 6.84} \\
~ & 0-shot & SaigaLlama3 8B & 66.05\tiny{$\pm$ 0.85} & 46.67\tiny{$\pm$ 0.00} & 54.69\tiny{$\pm$ 0.29} & 53.12\tiny{$\pm$ 0.36} & 71.58\tiny{$\pm$ 1.05} & 60.98\tiny{$\pm$ 0.62} & 57.84\tiny{$\pm$ 0.45} \\
~ & 0-shot MMLU & Vikhr 7B IT 0.4 & 69.15\tiny{$\pm$ 2.43} & 41.78\tiny{$\pm$ 0.89} & 52.09\tiny{$\pm$ 1.39} & 53.04\tiny{$\pm$ 1.06} & 77.89\tiny{$\pm$ 2.11} & 63.11\tiny{$\pm$ 1.45} & 57.60\tiny{$\pm$ 1.42} \\
    \hline
    \end{tabular}
    }
    
\caption{Comparison of the results of the best models from each group (mean $\pm$ std)}
\label{tab:results_main_best_exp}
\end{table*}

\subsubsection{DE Dataset}
The best scores of the F1 for the pathology class and F1-macro in this experiment are achieved on the essay dataset (DE) by the fine-tuned LLM model with 88.4\% F1-macro and 81.1\% F1 score for the pathology class. The model trained on the linguistic features shows comparable results with 85.8\% F1-macro and 77.0\% F1-pathology.

In general, the three best results on the DE dataset were achieved by fine-tuned models, which can be linked to the bigger dataset size and to the longer texts in the dataset, which, in turn, are crucial for supervised fine-tuning. For all other datasets, various prompting methods without finetuning significantly outperform SFT and LoRA.

\begin{table*}[t]
\centering
\resizebox{\linewidth}{!}{%
    \begin{tabular}{lllcccccccc}
    \hline
    \textbf{Corpus} & \textbf{Mode} & \textbf{Model} & \begin{tabular}[c]{@{}l@{}} \textbf{Precision} \\ \textbf{healthy}\end{tabular} & \begin{tabular}[c]{@{}l@{}} \textbf{Recall} \\ \textbf{healthy}\end{tabular} & \textbf{F1-healthy} & \begin{tabular}[c]{@{}l@{}} \textbf{Precision} \\ \textbf{pathology}\end{tabular} & \begin{tabular}[c]{@{}l@{}} \textbf{Recall} \\ \textbf{pathology}\end{tabular} & \textbf{F1-pathology} & \textbf{F1-macro}  \\
    \hline
    \multirow{7}{*}{D-all} & 5-shot MMLU & Gemma2 9B IT & 88.83\tiny{$\pm$ 0.25} & 96.56\tiny{$\pm$ 0.49} & 92.53\tiny{$\pm$ 0.36} & 59.56\tiny{$\pm$ 4.84} & 29.29\tiny{$\pm$ 1.43} & 39.26\tiny{$\pm$ 2.32} & 65.90\tiny{$\pm$ 1.34} \\
~ & LoRA & Gemma2 2B IT & 92.33\tiny{$\pm$ 0.94} & 97.14\tiny{$\pm$ 1.04} & 94.67\tiny{$\pm$ 0.72} & 76.44\tiny{$\pm$ 6.36} & 52.98\tiny{$\pm$ 6.33} & \textbf{62.33\tiny{$\pm$ 5.60}} & \textbf{78.50\tiny{$\pm$ 3.13}} \\
~ & SFT & RuBERT & 92.26\tiny{$\pm$ 0.52} & 96.22\tiny{$\pm$ 1.56} & 94.19\tiny{$\pm$ 0.68} & 72.12\tiny{$\pm$ 10.10} & 52.98\tiny{$\pm$ 3.81} & 60.60\tiny{$\pm$ 2.87} & 77.39\tiny{$\pm$ 1.72} \\
~ & 0-shot & Gemma2 9B IT & 93.28\tiny{$\pm$ 0.00} & 76.69\tiny{$\pm$ 0.00} & 84.18\tiny{$\pm$ 0.00} & 33.33\tiny{$\pm$ 0.00} & 67.86\tiny{$\pm$ 0.00} & 44.71\tiny{$\pm$ 0.00} & 64.44\tiny{$\pm$ 0.00} \\
~ & 0-shot MMLU & Gemma2 9B IT & 94.87\tiny{$\pm$ 0.00} & 68.10\tiny{$\pm$ 0.00} & 79.29\tiny{$\pm$ 0.00} & 29.73\tiny{$\pm$ 0.00} & 78.57\tiny{$\pm$ 0.00} & 43.14\tiny{$\pm$ 0.00} & 61.21\tiny{$\pm$ 0.00} \\
    ~ & SFT & Linguistic features & 88.10\tiny{$\pm$ 0.60} & 88.10\tiny{$\pm$ 3.40} & 88.00\tiny{$\pm$ 1.60} & 51.60\tiny{$\pm$ 5.80} & 50.70\tiny{$\pm$ 4.10} & 50.80\tiny{$\pm$ 2.60} & 69.40\tiny{$\pm$ 1.90}  \\ 
    ~ & SFT & TF-IDF & 89.50\tiny{$\pm$ 1.00} & 84.20\tiny{$\pm$ 1.00} & 86.80\tiny{$\pm$ 0.60} & 47.50\tiny{$\pm$ 1.80} & 59.10\tiny{$\pm$ 4.20} & 52.60\tiny{$\pm$ 2.50} & 69.70\tiny{$\pm$ 1.50}  \\ \hline
    \multirow{9}{*}{A-all} & 5-shot MMLU & SaigaLlama3 8B & 80.51\tiny{$\pm$ 5.64} & 11.26\tiny{$\pm$ 0.46} & 19.76\tiny{$\pm$ 0.88} & 48.46\tiny{$\pm$ 0.41} & 96.80\tiny{$\pm$ 1.07} & \textbf{64.59\tiny{$\pm$ 0.60}} & 42.18\tiny{$\pm$ 0.74} \\
~ & LoRA & SaigaLlama3 8B & 58.56\tiny{$\pm$ 1.32} & 56.32\tiny{$\pm$ 4.83} & 57.33\tiny{$\pm$ 2.79} & 51.59\tiny{$\pm$ 2.03} & 53.78\tiny{$\pm$ 3.74} & 52.56\tiny{$\pm$ 1.83} & 54.95\tiny{$\pm$ 1.57} \\
~ & SFT & RuRoBERTa & 61.53\tiny{$\pm$ 3.69} & 60.15\tiny{$\pm$ 13.80} & 59.68\tiny{$\pm$ 4.47} & 54.39\tiny{$\pm$ 0.97} & 54.44\tiny{$\pm$ 18.08} & 52.27\tiny{$\pm$ 11.87} & 55.97\tiny{$\pm$ 3.88} \\
~ & SFT & BERT & 59.44\tiny{$\pm$ 1.74} & 46.93\tiny{$\pm$ 3.66} & 52.33\tiny{$\pm$ 2.10} & 50.43\tiny{$\pm$ 0.98} & 62.67\tiny{$\pm$ 4.81} & 55.82\tiny{$\pm$ 2.21} & 54.08\tiny{$\pm$ 1.08} \\
~ & 0-shot & Qwen2 7B IT & 60.01\tiny{$\pm$ 0.41} & 69.66\tiny{$\pm$ 0.92} & 64.47\tiny{$\pm$ 0.17} & 56.72\tiny{$\pm$ 0.10} & 46.13\tiny{$\pm$ 1.60} & 50.87\tiny{$\pm$ 0.99} & 57.67\tiny{$\pm$ 0.41} \\
~ & 0-shot & Gemma2 2B IT & 71.44\tiny{$\pm$ 0.87} & 26.44\tiny{$\pm$ 0.00} & 38.59\tiny{$\pm$ 0.13} & 50.69\tiny{$\pm$ 0.15} & 87.73\tiny{$\pm$ 0.53} & 64.26\tiny{$\pm$ 0.27} & 51.42\tiny{$\pm$ 0.20} \\
~ & 0-shot MMLU & Qwen2 7B IT & 62.67\tiny{$\pm$ 0.06} & 64.83\tiny{$\pm$ 1.38} & 63.72\tiny{$\pm$ 0.65} & 57.51\tiny{$\pm$ 0.47} & 55.20\tiny{$\pm$ 1.07} & 56.32\tiny{$\pm$ 0.32} & \textbf{60.02\tiny{$\pm$ 0.17}} \\
    ~ & SFT & Linguistic features & 53.40\tiny{$\pm$ 2.90} & 39.70\tiny{$\pm$ 19.30} & 42.00\tiny{$\pm$ 17.90} & 46.10\tiny{$\pm$ 2.40} & 60.20\tiny{$\pm$ 20.40} & 51.00\tiny{$\pm$ 7.80} & 46.50\tiny{$\pm$ 6.20}  \\ 
    ~ & SFT & TF-IDF & 51.80\tiny{$\pm$ 0.70} & 40.00\tiny{$\pm$ 15.60} & 43.30\tiny{$\pm$ 10.80} & 44.30\tiny{$\pm$ 2.10} & 56.40\tiny{$\pm$ 17.80} & 48.60\tiny{$\pm$ 8.50} & 46.00\tiny{$\pm$ 2.40}  \\
    \hline 
    \end{tabular}
    }
\caption{Results of classification on D-all and A-all datasets, the best models from each group (mean $\pm$ std)}
\label{tab:results_combined_datasets_best}
\end{table*}

\begin{table*}[t]
\centering
\resizebox{\linewidth}{!}{%
    \begin{tabular}{lllccccccc}
    \hline
    \textbf{Transfer} & \textbf{Mode} & \textbf{Model} & \begin{tabular}[c]{@{}l@{}} \textbf{Precision} \\ \textbf{healthy}\end{tabular} & \begin{tabular}[c]{@{}l@{}} \textbf{Recall} \\ \textbf{healthy}\end{tabular} & \textbf{F1-healthy} & \begin{tabular}[c]{@{}l@{}} \textbf{Precision} \\ \textbf{pathology}\end{tabular} & \begin{tabular}[c]{@{}l@{}} \textbf{Recall} \\ \textbf{pathology}\end{tabular} & \textbf{F1-pathology} & \textbf{F1-macro}  \\
    \hline
    \multirow{13}{*}{{DE to DSM test}} & LoRA & SaigaLlama3 8B & 58.49\tiny{$\pm$ 3.68} & 83.95\tiny{$\pm$ 13.97} & 68.73\tiny{$\pm$ 7.39} & 40.34\tiny{$\pm$ 13.85} & 12.04\tiny{$\pm$ 4.99} & 17.21\tiny{$\pm$ 5.85} & 42.97\tiny{$\pm$ 4.20} \\
    ~ & LoRA & Vikhr 7B IT 0.4 & 60.33\tiny{$\pm$ 5.19} & 61.73\tiny{$\pm$ 20.91} & 58.81\tiny{$\pm$ 10.42} & 37.58\tiny{$\pm$ 5.47} & 37.04\tiny{$\pm$ 24.57} & 33.98\tiny{$\pm$ 15.06} & 46.39\tiny{$\pm$ 4.16} \\
    ~ & LoRA & Gemma2 2B IT & 64.14\tiny{$\pm$ 3.85} & 71.60\tiny{$\pm$ 21.45} & 65.89\tiny{$\pm$ 10.13} & 51.07\tiny{$\pm$ 8.34} & 38.89\tiny{$\pm$ 22.91} & 38.67\tiny{$\pm$ 16.04} & 52.28\tiny{$\pm$ 6.44} \\
    ~ & LoRA & Gemma2 9B IT & 64.46\tiny{$\pm$ 5.24} & 66.67\tiny{$\pm$ 20.73} & 64.11\tiny{$\pm$ 11.57} & 51.24\tiny{$\pm$ 8.98} & 45.37\tiny{$\pm$ 18.54} & 44.36\tiny{$\pm$ 11.95} & \textbf{54.23\tiny{$\pm$ 7.12}} \\
    ~ & LoRA & Vikhr 7B IT 5.4 & 58.24\tiny{$\pm$ 5.42} & 67.28\tiny{$\pm$ 21.75} & 60.24\tiny{$\pm$ 11.51} & 31.33\tiny{$\pm$ 7.80} & 25.93\tiny{$\pm$ 24.36} & 25.09\tiny{$\pm$ 15.40} & 42.66\tiny{$\pm$ 4.75} \\
    ~ & LoRA & Qwen2 7B IT & 61.14\tiny{$\pm$ 3.36} & 61.11\tiny{$\pm$ 18.61} & 59.75\tiny{$\pm$ 11.42} & 43.62\tiny{$\pm$ 8.81} & 42.59\tiny{$\pm$ 15.93} & 41.25\tiny{$\pm$ 8.00} & 50.50\tiny{$\pm$ 5.32} \\
    ~ & LoRA & VikhrGemma 2B IT & 60.84\tiny{$\pm$ 2.46} & 71.60\tiny{$\pm$ 16.93} & 64.90\tiny{$\pm$ 8.82} & 42.54\tiny{$\pm$ 7.31} & 31.48\tiny{$\pm$ 15.27} & 33.97\tiny{$\pm$ 12.07} & 49.44\tiny{$\pm$ 5.25} \\
    ~ & SFT & RuRoBERTa & 60.96\tiny{$\pm$ 7.01} & 41.98\tiny{$\pm$ 26.59} & 46.07\tiny{$\pm$ 14.36} & 33.75\tiny{$\pm$ 15.28} & 59.26\tiny{$\pm$ 27.34} & 42.95\tiny{$\pm$ 19.50} & 44.51\tiny{$\pm$ 5.07} \\
    ~ & SFT & RuBioRoBERTa & 61.65\tiny{$\pm$ 2.51} & 50.00\tiny{$\pm$ 23.88} & 52.54\tiny{$\pm$ 12.11} & 34.54\tiny{$\pm$ 15.55} & 52.78\tiny{$\pm$ 24.59} & 41.57\tiny{$\pm$ 18.71} & 47.05\tiny{$\pm$ 5.29} \\
    ~ & SFT & RuBERT & 68.00\tiny{$\pm$ 10.29} & 27.78\tiny{$\pm$ 9.01} & 38.68\tiny{$\pm$ 10.16} & 42.66\tiny{$\pm$ 2.80} & 80.56\tiny{$\pm$ 9.49} & 55.64\tiny{$\pm$ 4.19} & 47.16\tiny{$\pm$ 5.80} \\
    ~ & SFT & BERT & 76.85\tiny{$\pm$ 5.46} & 29.63\tiny{$\pm$ 8.28} & 42.21\tiny{$\pm$ 9.33} & 45.34\tiny{$\pm$ 2.29} & 87.04\tiny{$\pm$ 4.14} & \textbf{59.54\tiny{$\pm$ 2.03}} & 50.88\tiny{$\pm$ 5.43} \\
    ~ & SFT & Linguistic features & 60.00\tiny{$\pm$ 2.50} & 57.40\tiny{$\pm$ 7.00} & 58.50\tiny{$\pm$ 4.50} & 40.00\tiny{$\pm$ 3.50} & 42.60\tiny{$\pm$ 6.90} & 41.10\tiny{$\pm$ 4.50} & 49.80\tiny{$\pm$ 3.10}  \\
    ~ & SFT & TF-IDF & 59.90\tiny{$\pm$ 3.40} & 30.90\tiny{$\pm$ 9.20} & 40.00\tiny{$\pm$ 8.50} & 40.20\tiny{$\pm$ 1.50} & 69.40\tiny{$\pm$ 8.30} & 50.70\tiny{$\pm$ 2.70} & 45.40\tiny{$\pm$ 3.80}  \\
    \hline
    \end{tabular}
    }
\caption{Transfer models from DE to DSM (mean $\pm$ std)}
\label{tab:results_DE_to_DSM}
\end{table*}

\subsubsection{DSM Dataset}
The best result achieved by the Vikhr 7B IT 0.4 model, evaluated in the 5-shot regime, with 63.69\% F1-macro. The traditional machine learning methods, as well as encoder models, performed poorly with nearly 53\% F1-macro on linguistic features.

Comparing the results between the DE and DSM datasets, a significant difference in classification quality can be observed. This may be due to several factors. First of all, the sample size is significantly smaller in the DSM dataset, as machine learning models in general can show low accuracy on limited data. On the other hand, if we refer to the study \citep{ignatiev2022predicting}, where less stringent restrictions on text volume and temporal proximity to questionnaire screening dates were applied to the same raw data, the results were still not very high: about 60\% F1-macro for psycholinguistic features and TF-IDF features.

Secondly, the text format in the DSM introduces some noise. The texts in DSM are concatenated together with a collection of posts from users' personal pages, and they mostly lack coherent logic and cohesion in the resulting text. Even with the 6,000-character limit, the standard deviation in word count is approximately 300 words, compared with 100 in DE. Thus, DSM is a much noisier dataset, which can strongly affect the quality of classification with psycholinguistic markers, where the values of some markers can be affected by the volume of text analyzed, even considering their normalization with respect to the text volume.

Finally, the way in which the target class of pathology is defined can be of great importance. Although a widely used and validated psychological questionnaire was used for the DSM dataset, its results cannot be compared with a clinically confirmed diagnosis. The same findings can be outlined in work that criticizes the way in which social media users are labeled for mental illness by indirectly affiliating or self-identifying mental ill-health \citep{ernala2019methodological}. In favor of the significance of this factor is also the fact that TF-IDF-based features performed significantly better on DE data than on DSM, although, unlike psycholinguistic markers, they do not have an initial specialization for detecting signs of mental ill-health.

To take into account these considerations, we applied LLMs to this task and showed that the bigger model is able to partially overcome these issues. While RuBERT and the model trained on linguistic features both show comparatively low results, a 5-shot LLM performs significantly better. The same outcomes are observed in the experiments with anxiety datasets.

For a better understanding of per-class performance of models, we also provided results for the best models on the DSM dataset for various splits within a class based on the initial BDI scores. The results are provided in \Cref{app:dsm_splits}.

\subsubsection{AL, AD, and AC Datasets}
Turning to the anxiety detection task, the difference between the AutoML-based and encoder-based models is only noticeable on the picture description (AD) dataset, where the accuracy of the RuRoBERTa model was 55\% F1-macro. Overall, it can be said that none of the non-LLM models performed well in this task. However, LLMs performed significantly better on these tasks. Models with LoRA, in general, perform better than encoder-based and AutoML-based models, but the best performance is achieved by 0-shot and 5-shot prompting. Again, this can be linked to the complexity of the domain and the small amount of training data, which makes it hard to fine-tune a model.

In general, the LLMs outperform other models on all datasets, but on the DE and DSM, the gap between various types of models is smaller, leaving the usage of non-LLM models reasonable for some specific cases.

\subsection{Classification on D-all and A-all Datasets} \label{subsec:results_table_3}

We combined the depression and anxiety datasets into pooled datasets D-all (DE and DSM) and A-all (AL, AD, and AC). Classification performance of the best models on D-all and A-all combined data is presented in \Cref{tab:results_combined_datasets_best}.

The LLMs performed better than other models on both D-all and A-all. It is also noticeable that methods with model tuning work better on D-all, while prompting shows better results on A-all. It can be explained with significant genre and length variations in A-all datasets, so the general-purpose models perform better than finetuned ones due to the complexity of the data.

\subsection{Classification of DSM Data with Best Models from DE} \label{subsec:results_table_4}

The classification results of models that demonstrated the best performance on the DE dataset applied to DSM test data are shown in \Cref{tab:results_DE_to_DSM}.

This experiment shows that models trained on essays from depressed subjects cannot be used to detect depression from social media texts from users who have taken a depression questionnaire and received a high score. Similar findings were shown in \citep{ernala2019methodological}, but with a reverse logic of the experiments.

Such results may also be just due to the fact that essay texts and collections of social network posts are very different in genre. A sample of social network users with clinically diagnosed depression would be needed to clarify this issue.

\subsection{LLM Results Interpretation} \label{subsec:results_table_5}
As shown in \Cref{subsec:results_table_4}, the overall best results were obtained on the DE dataset with LLMs. To further analyze the quality of these results from the clinical perspective, we conducted an additional experiment on LLM results interpretation.

We chose the best LLM in a 5-shot setting on the DE dataset (Vikhr 7B IT 5.4) and asked two psychologists with relevant experience to rate the LLM generation on a scale from 1 to 5. The LLM-generated answer in this setting consists of a predicted class label and a detailed explanation of why this label was chosen. Details of the rating scale and statistics are provided in \Cref{app:interpretation_setup}. The psychologists assessed only texts from patients with clinically diagnosed depression that the LLM correctly labeled.

The average score from expert psychologists for LLM explanations was assigned to 2.84 out of 5. Moreover, the psychologists also noted each explanation, which can be marked as true to some extent (e.g., with some true claims in the explanation) -- there are approximately 66\% of such explanations. These results show that the explanations for depression detection from texts contain both correct and erroneous parts. Overall, the LLM (in the described settings) does not generate enough explanations that meet the requirements of clinicians.

To further extend the error analysis in the LLM generations in the psychological domain, we asked psychologists to describe the most common errors in the generated explanations. To do so, psychologists annotated each explanation with the list of errors and categorized these mistakes into the following groups: (1) tautology, (2) groundless generalization, (3) false conclusion, (4) confabulation, (5) distortion of medical understanding of depression, (6) incompleteness of selected signs of depression. Description and examples of these error types are located in \Cref{tab:llm_error_types} in the \Cref{app:interpretation_setup}. It should also be noted that from the general NLP perspective, most of these types of errors can be defined as hallucinations; however, we suppose that a more precise definition is needed in this specific domain.

On average, each explanation contains two or more errors. The most common types of errors are groundless generalization, false conclusion, and confabulation, which occur in 56\%, 56\%, and 50\% of samples, respectively. However, these errors appear several times in each explanation. The incompleteness of selected signs of depression appears in 44\%, while distortion of medical understanding of depression and tautology are the rare errors, which appear in 19\% and 13\%. These results show that there exist various types of errors, specific to the mental health disorders domain. The additional results with detailed analysis of feature importance for the pathology and healthy class prediction are located in \Cref{app:feature_importance}.

\section{Conclusion} \label{sec:conclusion}
In this paper, we have investigated the effectiveness of traditional machine learning methods and LLMs on the task of detecting depression and anxiety. The results obtained in our work establish the new state-of-the-art on five Russian-language datasets.

Our investigation shows that psycholinguistic features can produce results at the level of encoder-based models when texts from individuals with clinically diagnosed depression are used for training. BERT models, in turn, perform better on noisy text data, where examples from the training sample may vary widely in text length or genre. LLM-based models performed best on all five different mental health datasets. Even without fine-tuning, LLMs usually demonstrate relatively high performance. In response to \textbf{RQ1}, the experimental results indicate that LLM-based models have high potential for detecting mental disorders from texts. 

In response to \textbf{RQ2}, the findings reveal that models trained on essays from depressed individuals are not effective for detecting depression in social media texts from users who have completed a depression questionnaire and scored high.

Finally, we evaluated LLM-generated explanations and showed that in the current state, these explanations do not meet the clinicians' requirements with an overall score of 2.84 out of 5, which answers \textbf{RQ3}. We also constructed a detailed classification of common errors in LLM explanations from the clinicians' perspective, to guide further improvements of LLMs in this domain.

\section*{Limitations} \label{sec:limitations}
The main limitation of this paper is that it is impossible to share all the raw texts from the used datasets, as they are all distributed under different terms. However, we present several anonymized textual examples and extracted psycholinguistic features.

The amount of sample data used for predicting anxiety is small, which does not allow us to adequately judge the possibility of predicting anxiety in Russian text using machine learning methods. Although the results on all the anxiety datasets used show poor accuracy, perhaps a very different scale of data is needed for this task.
The paper does not discuss the differences between the results within one group of the models in detail, as this was not the aim of the paper.

We conducted the experiments only for Russian due to the poor availability of related datasets for other languages. However, the used methods in general are language agnostic, so the results could be extended to other languages. The overall result about LLMs as the best-performing models matches similar studies for English on closed data.

The final limitation is the temporal validity of the results. Due to the fast growth of LLM-based solutions, future LLMs could outperform the obtained results. However, we suppose that the current state of the field already presents an interest and is therefore investigated in the paper.

\section*{Ethical Considerations} \label{sec:ethical}
The problem discussed in this paper is the sensitive issue of mental health. To avoid any possible harm, we did not fully open-source the used datasets. All shared data does not contain any information that names or uniquely identifies individuals, nor does it contain offensive content. The examined models for the detection of mental disorders do not aim to replace a professional physician; on the contrary, these models are intended to support a human expert.

\section*{Acknowledgements}
The work of Gleb Kuzmin was supported by a grant, provided by the MED of Russia (agreement identifier 000000C313925P4G0002) and the agreement with the ISP RAS dated June 20, 2025 No. 139-15-2025-011.

The research was carried out using the infrastructure of the shared research facilities «High Performance Computing and Big Data » of FRC CSC RAS (CKP «Informatics»)

\bibliography{custom}

\begin{thebibliography}{45}
\providecommand{\natexlab}[1]{#1}

\bibitem[{Ansari et~al.(2023)Ansari, Ji, Chen, and
  Cambria}]{Ansari2023EnsembleHL}
Luna Ansari, Shaoxiong Ji, Qian Chen, and Erik Cambria. 2023.
\newblock \href {https://doi.org/10.1109/TCSS.2022.3154442} {Ensemble hybrid
  learning methods for automated depression detection}.
\newblock \emph{IEEE Transactions on Computational Social Systems},
  10:211--219.

\bibitem[{Babu and Kanaga(2022)}]{Babu2021SentimentAI}
Nirmal~Varghese Babu and Edward Grace~Mary Kanaga. 2022.
\newblock \href {https://doi.org/10.1007/S42979-021-00958-1} {Sentiment
  analysis in social media data for depression detection using artificial
  intelligence: {A} review}.
\newblock \emph{{SN} Computer Science}, 3(1):74.

\bibitem[{Beck et~al.(1961)Beck, Ward, Mendelson, Mock, and
  Erbaugh}]{beck1961inventory}
Aaron~T. Beck, Calvin~H. Ward, Mock Mendelson, Jeremiah Mock, and John Erbaugh.
  1961.
\newblock An inventory for measuring depression.
\newblock \emph{Archives of general psychiatry}, 4(6):561--571.

\bibitem[{Bjelland et~al.(2002)Bjelland, Dahl, Haug, and
  Neckelmann}]{bjelland2002validity}
Ingvar Bjelland, Alv~A Dahl, Tone~Tangen Haug, and Dag Neckelmann. 2002.
\newblock \href {https://doi.org/10.1016/S0022-3999(01)00296-3} {The validity
  of the hospital anxiety and depression scale: an updated literature review}.
\newblock \emph{Journal of psychosomatic research}, 52(2):69--77.

\bibitem[{Calixto et~al.(2022)Calixto, Yaneva, and
  Cardoso}]{calixto2022natural}
Iacer Calixto, Victoria Yaneva, and Raphael~Moura Cardoso. 2022.
\newblock Natural language processing for mental disorders: an overview.
\newblock \emph{Natural Language Processing In Healthcare}, pages 37--59.

\bibitem[{Chancellor and De~Choudhury(2020)}]{chancellor2020methods}
Stevie Chancellor and Munmun De~Choudhury. 2020.
\newblock \href {https://doi.org/10.1038/S41746-020-0233-7} {Methods in
  predictive techniques for mental health status on social media: a critical
  review}.
\newblock \emph{NPJ digital medicine}, 3(1):43.

\bibitem[{Derogatis(1983)}]{derogatis1983scl}
Leonard~R. Derogatis. 1983.
\newblock {SCL}-90-{R}: Administration, scoring and procedures.
\newblock \emph{Manual II for the R (evised) Version and Other Instruments of
  the Psychopathology Rating Scale Series}.

\bibitem[{Devlin et~al.(2019)Devlin, Chang, Lee, and
  Toutanova}]{devlin-etal-2019-bert}
Jacob Devlin, Ming-Wei Chang, Kenton Lee, and Kristina Toutanova. 2019.
\newblock \href {https://doi.org/10.18653/v1/N19-1423} {{BERT}: Pre-training of
  deep bidirectional transformers for language understanding}.
\newblock In \emph{Proceedings of the 2019 Conference of the North {A}merican
  Chapter of the Association for Computational Linguistics: Human Language
  Technologies, Volume 1 (Long and Short Papers)}, pages 4171--4186,
  Minneapolis, Minnesota. Association for Computational Linguistics.

\bibitem[{Ernala et~al.(2019)Ernala, Birnbaum, Candan, Rizvi, Sterling, Kane,
  and Choudhury}]{ernala2019methodological}
Sindhu~Kiranmai Ernala, Michael~L. Birnbaum, Kristin~A. Candan, Asra~F. Rizvi,
  William~A. Sterling, John~M. Kane, and Munmun~De Choudhury. 2019.
\newblock \href {https://doi.org/10.1145/3290605.3300364} {Methodological gaps
  in predicting mental health states from social media: Triangulating
  diagnostic signals}.
\newblock In \emph{Proceedings of the 2019 {CHI} Conference on Human Factors in
  Computing Systems, {CHI} 2019, Glasgow, Scotland, UK, May 04-09, 2019}, page
  134. {ACM}.

\bibitem[{Feurer et~al.(2015)Feurer, Klein, Eggensperger, Springenberg, Blum,
  and Hutter}]{feurer-neurips15a}
Matthias Feurer, Aaron Klein, Katharina Eggensperger, Jost~Tobias Springenberg,
  Manuel Blum, and Frank Hutter. 2015.
\newblock \href
  {https://proceedings.neurips.cc/paper/2015/hash/11d0e6287202fced83f79975ec59a3a6-Abstract.html}
  {Efficient and robust automated machine learning}.
\newblock In \emph{Advances in Neural Information Processing Systems 28: Annual
  Conference on Neural Information Processing Systems 2015, December 7-12,
  2015, Montreal, Quebec, Canada}, pages 2962--2970.

\bibitem[{Garg(2023)}]{garg2023mental}
Muskan Garg. 2023.
\newblock \href {https://doi.org/10.1007/s11831-022-09863-z} {Mental health
  analysis in social media posts: a survey}.
\newblock \emph{Archives of Computational Methods in Engineering},
  30(3):1819--1842.

\bibitem[{Graham et~al.(2019)Graham, Depp, Lee, Nebeker, Tu, Kim, and
  Jeste}]{graham2019artificial}
Sarah Graham, Colin Depp, Ellen~E Lee, Camille Nebeker, Xin Tu, Ho-Cheol Kim,
  and Dilip~V Jeste. 2019.
\newblock \href {https://doi.org/10.1007/s11920-019-1094-0} {Artificial
  intelligence for mental health and mental illnesses: an overview}.
\newblock \emph{Current psychiatry reports}, 21:1--18.

\bibitem[{Grattafiori et~al.(2024)Grattafiori, Dubey, Jauhri, Pandey, Kadian,
  Al-Dahle, Letman, Mathur, Schelten, Vaughan, Yang, Fan, Goyal, Hartshorn,
  Yang, Mitra, Sravankumar, Korenev, Hinsvark, Rao, Zhang, Rodriguez,
  Gregerson, Spataru, Roziere, Biron, Tang, Chern, Caucheteux, Nayak, Bi,
  Marra, McConnell, Keller, Touret, Wu, Wong, Ferrer, Nikolaidis, Allonsius,
  Song, Pintz, Livshits, Wyatt, Esiobu, Choudhary, Mahajan, Garcia-Olano,
  Perino, Hupkes, Lakomkin, AlBadawy, Lobanova, Dinan, Smith, Radenovic,
  Guzmán, Zhang, Synnaeve, Lee, Anderson, Thattai, Nail, Mialon, Pang,
  Cucurell, Nguyen, Korevaar, Xu, Touvron, Zarov, Ibarra, Kloumann, Misra,
  Evtimov, Zhang, Copet, Lee, Geffert, Vranes, Park, Mahadeokar, Shah, van~der
  Linde, Billock, Hong, Lee, Fu, Chi, Huang, Liu, Wang, Yu, Bitton, Spisak,
  Park, Rocca, Johnstun, Saxe, Jia, Alwala, Prasad, Upasani, Plawiak, Li,
  Heafield, Stone, El-Arini, Iyer, Malik, Chiu, Bhalla, Lakhotia,
  Rantala-Yeary, van~der Maaten, Chen, Tan, Jenkins, Martin, Madaan, Malo,
  Blecher, Landzaat, de~Oliveira, Muzzi, Pasupuleti, Singh, Paluri, Kardas,
  Tsimpoukelli, Oldham, Rita, Pavlova, Kambadur, Lewis, Si, Singh, Hassan,
  Goyal, Torabi, Bashlykov, Bogoychev, Chatterji, Zhang, Duchenne, Çelebi,
  Alrassy, Zhang, Li, Vasic, Weng, Bhargava, Dubal, Krishnan, Koura, Xu, He,
  Dong, Srinivasan, Ganapathy, Calderer, Cabral, Stojnic, Raileanu, Maheswari,
  Girdhar, Patel, Sauvestre, Polidoro, Sumbaly, Taylor, Silva, Hou, Wang,
  Hosseini, Chennabasappa, Singh, Bell, Kim, Edunov, Nie, Narang, Raparthy,
  Shen, Wan, Bhosale, Zhang, Vandenhende, Batra, Whitman, Sootla, Collot,
  Gururangan, Borodinsky, Herman, Fowler, Sheasha, Georgiou, Scialom,
  Speckbacher, Mihaylov, Xiao, Karn, Goswami, Gupta, Ramanathan, Kerkez,
  Gonguet, Do, Vogeti, Albiero, Petrovic, Chu, Xiong, Fu, Meers, Martinet,
  Wang, Wang, Tan, Xia, Xie, Jia, Wang, Goldschlag, Gaur, Babaei, Wen, Song,
  Zhang, Li, Mao, Coudert, Yan, Chen, Papakipos, Singh, Srivastava, Jain,
  Kelsey, Shajnfeld, Gangidi, Victoria, Goldstand, Menon, Sharma, Boesenberg,
  Baevski, Feinstein, Kallet, Sangani, Teo, Yunus, Lupu, Alvarado, Caples, Gu,
  Ho, Poulton, Ryan, Ramchandani, Dong, Franco, Goyal, Saraf, Chowdhury,
  Gabriel, Bharambe, Eisenman, Yazdan, James, Maurer, Leonhardi, Huang, Loyd,
  Paola, Paranjape, Liu, Wu, Ni, Hancock, Wasti, Spence, Stojkovic, Gamido,
  Montalvo, Parker, Burton, Mejia, Liu, Wang, Kim, Zhou, Hu, Chu, Cai, Tindal,
  Feichtenhofer, Gao, Civin, Beaty, Kreymer, Li, Adkins, Xu, Testuggine, David,
  Parikh, Liskovich, Foss, Wang, Le, Holland, Dowling, Jamil, Montgomery,
  Presani, Hahn, Wood, Le, Brinkman, Arcaute, Dunbar, Smothers, Sun, Kreuk,
  Tian, Kokkinos, Ozgenel, Caggioni, Kanayet, Seide, Florez, Schwarz, Badeer,
  Swee, Halpern, Herman, Sizov, Guangyi, Zhang, Lakshminarayanan, Inan,
  Shojanazeri, Zou, Wang, Zha, Habeeb, Rudolph, Suk, Aspegren, Goldman, Zhan,
  Damlaj, Molybog, Tufanov, Leontiadis, Veliche, Gat, Weissman, Geboski, Kohli,
  Lam, Asher, Gaya, Marcus, Tang, Chan, Zhen, Reizenstein, Teboul, Zhong, Jin,
  Yang, Cummings, Carvill, Shepard, McPhie, Torres, Ginsburg, Wang, Wu, U,
  Saxena, Khandelwal, Zand, Matosich, Veeraraghavan, Michelena, Li, Jagadeesh,
  Huang, Chawla, Huang, Chen, Garg, A, Silva, Bell, Zhang, Guo, Yu, Moshkovich,
  Wehrstedt, Khabsa, Avalani, Bhatt, Mankus, Hasson, Lennie, Reso, Groshev,
  Naumov, Lathi, Keneally, Liu, Seltzer, Valko, Restrepo, Patel, Vyatskov,
  Samvelyan, Clark, Macey, Wang, Hermoso, Metanat, Rastegari, Bansal,
  Santhanam, Parks, White, Bawa, Singhal, Egebo, Usunier, Mehta, Laptev, Dong,
  Cheng, Chernoguz, Hart, Salpekar, Kalinli, Kent, Parekh, Saab, Balaji,
  Rittner, Bontrager, Roux, Dollar, Zvyagina, Ratanchandani, Yuvraj, Liang,
  Alao, Rodriguez, Ayub, Murthy, Nayani, Mitra, Parthasarathy, Li, Hogan,
  Battey, Wang, Howes, Rinott, Mehta, Siby, Bondu, Datta, Chugh, Hunt, Dhillon,
  Sidorov, Pan, Mahajan, Verma, Yamamoto, Ramaswamy, Lindsay, Lindsay, Feng,
  Lin, Zha, Patil, Shankar, Zhang, Zhang, Wang, Agarwal, Sajuyigbe, Chintala,
  Max, Chen, Kehoe, Satterfield, Govindaprasad, Gupta, Deng, Cho, Virk,
  Subramanian, Choudhury, Goldman, Remez, Glaser, Best, Koehler, Robinson, Li,
  Zhang, Matthews, Chou, Shaked, Vontimitta, Ajayi, Montanez, Mohan, Kumar,
  Mangla, Ionescu, Poenaru, Mihailescu, Ivanov, Li, Wang, Jiang, Bouaziz,
  Constable, Tang, Wu, Wang, Wu, Gao, Kleinman, Chen, Hu, Jia, Qi, Li, Zhang,
  Zhang, Adi, Nam, Yu, Wang, Zhao, Hao, Qian, Li, He, Rait, DeVito, Rosnbrick,
  Wen, Yang, Zhao, and Ma}]{dubey2024llama}
Aaron Grattafiori, Abhimanyu Dubey, Abhinav Jauhri, Abhinav Pandey, Abhishek
  Kadian, Ahmad Al-Dahle, Aiesha Letman, Akhil Mathur, Alan Schelten, Alex
  Vaughan, Amy Yang, Angela Fan, Anirudh Goyal, Anthony Hartshorn, Aobo Yang,
  Archi Mitra, Archie Sravankumar, Artem Korenev, Arthur Hinsvark, and 542
  others. 2024.
\newblock \href {https://arxiv.org/abs/2407.21783} {The {L}lama 3 {H}erd of
  {M}odels}.
\newblock \emph{arXiv preprint arXiv:2407.21783}.

\bibitem[{Guntuku et~al.(2017)Guntuku, Yaden, Kern, Ungar, and
  Eichstaedt}]{guntuku2017detecting}
Sharath~Chandra Guntuku, David~B. Yaden, Margaret~L Kern, Lyle~H Ungar, and
  Johannes~C Eichstaedt. 2017.
\newblock \href {https://doi.org/10.1016/j.cobeha.2017.07.005} {Detecting
  depression and mental illness on social media: an integrative review}.
\newblock \emph{Current Opinion in Behavioral Sciences}, 18:43--49.

\bibitem[{Hadzic et~al.(2024)Hadzic, Mohammed, Danner, Ohse, Zhang, Shiban, and
  Rätsch}]{doi:10.1080/18824889.2024.2342624}
Bakir Hadzic, Parvez Mohammed, Michael Danner, Julia Ohse, Yihong Zhang,
  Youssef Shiban, and Matthias Rätsch. 2024.
\newblock \href {https://doi.org/10.1080/18824889.2024.2342624} {Enhancing
  early depression detection with ai: a comparative use of nlp models}.
\newblock \emph{SICE Journal of Control, Measurement, and System Integration},
  17(1):135--143.

\bibitem[{Hendrycks et~al.(2021)Hendrycks, Burns, Basart, Zou, Mazeika, Song,
  and Steinhardt}]{hendryckstest2021}
Dan Hendrycks, Collin Burns, Steven Basart, Andy Zou, Mantas Mazeika, Dawn
  Song, and Jacob Steinhardt. 2021.
\newblock \href {https://openreview.net/forum?id=d7KBjmI3GmQ} {Measuring
  massive multitask language understanding}.
\newblock In \emph{9th International Conference on Learning Representations,
  {ICLR} 2021, Virtual Event, Austria, May 3-7, 2021}.

\bibitem[{Hu et~al.(2022)Hu, yelong shen, Wallis, Allen-Zhu, Li, Wang, Wang,
  and Chen}]{hu2022lora}
Edward~J Hu, yelong shen, Phillip Wallis, Zeyuan Allen-Zhu, Yuanzhi Li, Shean
  Wang, Lu~Wang, and Weizhu Chen. 2022.
\newblock \href {https://openreview.net/forum?id=nZeVKeeFYf9} {Lo{RA}: Low-rank
  adaptation of large language models}.
\newblock In \emph{The Tenth International Conference on Learning
  Representations, {ICLR} 2022, Virtual Event, April 25-29, 2022}.

\bibitem[{Ignatiev et~al.(2022)Ignatiev, Smirnov, and
  Stankevich}]{ignatiev2022predicting}
Nikolay Ignatiev, Ivan~V. Smirnov, and Maxim Stankevich. 2022.
\newblock \href {https://doi.org/10.5220/0010986100003122} {Predicting
  depression with text, image, and profile data from social media}.
\newblock In \emph{Proceedings of the 11th International Conference on Pattern
  Recognition Applications and Methods, {ICPRAM} 2022, Online Streaming,
  February 3-5, 2022}, pages 753--760. {SCITEPRESS}.

\bibitem[{Jeong et~al.(2023)Jeong, Yoon, Sohn, and Choi}]{jeong2023reading}
Jinwoo Jeong, Sujin Yoon, Dongyoung Sohn, and Yong~Suk Choi. 2023.
\newblock \href {https://ijoc.org/index.php/ijoc/article/view/20241} {Reading
  emotions in the digital age: A deep learning approach to detecting anxiety
  during the covid-19 pandemic through social media}.
\newblock \emph{International Journal of Communication}, 17:22.

\bibitem[{Jiang et~al.(2023)Jiang, Sablayrolles, Mensch, Bamford, Chaplot,
  de~las Casas, Bressand, Lengyel, Lample, Saulnier, Lavaud, Lachaux, Stock,
  Scao, Lavril, Wang, Lacroix, and Sayed}]{jiang2023mistral}
Albert~Q. Jiang, Alexandre Sablayrolles, Arthur Mensch, Chris Bamford,
  Devendra~Singh Chaplot, Diego de~las Casas, Florian Bressand, Gianna Lengyel,
  Guillaume Lample, Lucile Saulnier, Lélio~Renard Lavaud, Marie-Anne Lachaux,
  Pierre Stock, Teven~Le Scao, Thibaut Lavril, Thomas Wang, Timothée Lacroix,
  and William~El Sayed. 2023.
\newblock \href {https://arxiv.org/abs/2310.06825} {Mistral 7b}.
\newblock \emph{arXiv preprint arXiv:2310.06825}.

\bibitem[{Kokhlikyan et~al.(2020)Kokhlikyan, Miglani, Martin, Wang, Alsallakh,
  Reynolds, Melnikov, Kliushkina, Araya, Yan, and
  Reblitz-Richardson}]{kokhlikyan2020captum}
Narine Kokhlikyan, Vivek Miglani, Miguel Martin, Edward Wang, Bilal Alsallakh,
  Jonathan Reynolds, Alexander Melnikov, Natalia Kliushkina, Carlos Araya, Siqi
  Yan, and Orion Reblitz-Richardson. 2020.
\newblock \href {https://arxiv.org/abs/2009.07896} {Captum: A unified and
  generic model interpretability library for pytorch}.
\newblock \emph{arXiv preprint arXiv:2009.07896}.

\bibitem[{Kuratov and Arkhipov(2019)}]{kuratov2019adaptation}
Yuri Kuratov and Mikhail Arkhipov. 2019.
\newblock \href {http://arxiv.org/abs/1905.07213} {Adaptation of deep
  bidirectional multilingual transformers for russian language}.
\newblock \emph{arXiv preprint arXiv:1905.07213}.

\bibitem[{Lan et~al.(2024)Lan, Cheng, Sheng, Gao, and Li}]{lan2024depression}
Xiaochong Lan, Yiming Cheng, Li~Sheng, Chen Gao, and Yong Li. 2024.
\newblock \href {https://arxiv.org/abs/2403.10750} {Depression detection on
  social media with large language models}.
\newblock \emph{arXiv preprint arXiv:2403.10750}.

\bibitem[{Lan et~al.(2020)Lan, Chen, Goodman, Gimpel, Sharma, and
  Soricut}]{lan2019albert}
Zhenzhong Lan, Mingda Chen, Sebastian Goodman, Kevin Gimpel, Piyush Sharma, and
  Radu Soricut. 2020.
\newblock \href {https://openreview.net/forum?id=H1eA7AEtvS} {{ALBERT:} {A}
  lite {BERT} for self-supervised learning of language representations}.
\newblock In \emph{8th International Conference on Learning Representations,
  {ICLR} 2020, Addis Ababa, Ethiopia, April 26-30, 2020}.

\bibitem[{Litvinova and Ryzhkova(2018)}]{litvinova2018rusneuropsych}
Tatiana Litvinova and Ekarerina Ryzhkova. 2018.
\newblock \href {http://injoit.ru/index.php/j1/article/view/542}
  {Rusneuropsych: open corpus for study relations between author demographic,
  personality traits, lateral preferences and affect in text}.
\newblock \emph{International Journal of Open Information Technologies},
  6(3):32--36.

\bibitem[{Lynn et~al.(2018)Lynn, Goodman, Niederhoffer, Loveys, Resnik, and
  Schwartz}]{lynn2018clpsych}
Veronica Lynn, Alissa Goodman, Kate Niederhoffer, Kate Loveys, Philip Resnik,
  and H.~Andrew Schwartz. 2018.
\newblock \href {https://doi.org/10.18653/v1/W18-0604} {{CLP}sych 2018 shared
  task: Predicting current and future psychological health from childhood
  essays}.
\newblock In \emph{Proceedings of the Fifth Workshop on Computational
  Linguistics and Clinical Psychology: From Keyboard to Clinic}, pages 37--46,
  New Orleans, LA. Association for Computational Linguistics.

\bibitem[{Mayer et~al.(2024)Mayer, Warikoo, Eliassaf, Atzil-Slonim, and
  Gurevych}]{mayer2024predicting}
Tobias Mayer, Neha Warikoo, Amir Eliassaf, Dana Atzil-Slonim, and Iryna
  Gurevych. 2024.
\newblock \href {https://doi.org/10.18653/v1/2024.eacl-long.88} {Predicting
  client emotions and therapist interventions in psychotherapy dialogues}.
\newblock In \emph{Proceedings of the 18th Conference of the European Chapter
  of the Association for Computational Linguistics (Volume 1: Long Papers)},
  pages 1463--1477, St. Julian{'}s, Malta. Association for Computational
  Linguistics.

\bibitem[{Medvedeva et~al.(2021)Medvedeva, Enikolopov, Boyko, Vorontsova, and
  Stankevich}]{medvedeva2021lexical}
Tatyana~I. Medvedeva, Sergei~N. Enikolopov, Olga~M. Boyko, Oksana~Yu.
  Vorontsova, and Maxim~A. Stankevich. 2021.
\newblock \href {https://doi.org/10.11621/vsp.2021.03.03} {Lexical analysis of
  statements about covid-19 of people with a high level of somatization}.
\newblock \emph{Lomonosov Psychology Journal}, 14(3):39--64.

\bibitem[{Morales and Levitan(2016)}]{morales2016speech}
Michelle~Renee Morales and Rivka Levitan. 2016.
\newblock \href {https://doi.org/10.1109/SLT.2016.7846256} {Speech vs. text:
  {A} comparative analysis of features for depression detection systems}.
\newblock In \emph{2016 {IEEE} Spoken Language Technology Workshop, {SLT} 2016,
  San Diego, CA, USA, December 13-16, 2016}, pages 136--143. {IEEE}.

\bibitem[{Nikolich et~al.(2024)Nikolich, Korolev, Bratchikov, Kiselev, and
  Shelmanov}]{nikolich2024vikhr}
Aleksandr Nikolich, Konstantin Korolev, Sergei Bratchikov, Igor Kiselev, and
  Artem Shelmanov. 2024.
\newblock \href {https://doi.org/10.18653/v1/2024.mrl-1.15} {Vikhr:
  Constructing a state-of-the-art bilingual open-source instruction-following
  large language model for {R}ussian}.
\newblock In \emph{Proceedings of the Fourth Workshop on Multilingual
  Representation Learning (MRL 2024)}, pages 189--199, Miami, Florida, USA.
  Association for Computational Linguistics.

\bibitem[{Omar and Levkovich(2025)}]{OMAR2025234}
Mahmud Omar and Inbar Levkovich. 2025.
\newblock \href {https://doi.org/10.1016/j.jad.2024.11.052} {Exploring the
  efficacy and potential of large language models for depression: A systematic
  review}.
\newblock \emph{Journal of Affective Disorders}, 371:234--244.

\bibitem[{Owen et~al.(2020)Owen, Camacho-Collados, and
  Espinosa~Anke}]{owen-etal-2020-towards}
David Owen, Jose Camacho-Collados, and Luis Espinosa~Anke. 2020.
\newblock \href {https://aclanthology.org/2020.smm4h-1.12} {Towards preemptive
  detection of depression and anxiety in {T}witter}.
\newblock In \emph{Proceedings of the Fifth Social Media Mining for Health
  Applications Workshop {\&} Shared Task}, pages 82--89, Barcelona, Spain
  (Online). Association for Computational Linguistics.

\bibitem[{Qin et~al.(2023)Qin, Chen, Wang, Lan, Ren, and Hong}]{qin2023read}
Wei Qin, Zetong Chen, Lei Wang, Yunshi Lan, Weijieying Ren, and Richang Hong.
  2023.
\newblock \href {https://arxiv.org/abs/2305.05138} {Read, diagnose and chat:
  Towards explainable and interactive llms-augmented depression detection in
  social media}.
\newblock \emph{arXiv preprint arXiv:2305.05138}.

\bibitem[{Ringeval et~al.(2017)Ringeval, Schuller, Valstar, Gratch, Cowie,
  Scherer, Mozgai, Cummins, Schmitt, and Pantic}]{ringeval2017avec}
Fabien Ringeval, Bj{\"{o}}rn~W. Schuller, Michel~F. Valstar, Jonathan Gratch,
  Roddy Cowie, Stefan Scherer, Sharon Mozgai, Nicholas Cummins, Maximilian
  Schmitt, and Maja Pantic. 2017.
\newblock \href {https://doi.org/10.1145/3133944.3133953} {{AVEC} 2017:
  Real-life depression, and affect recognition workshop and challenge}.
\newblock In \emph{Proceedings of the 7th Annual Workshop on Audio/Visual
  Emotion Challenge, Mountain View, CA, USA, October 23 - 27, 2017}, pages
  3--9. {ACM}.

\bibitem[{Shah et~al.(2020)Shah, Ahmed, Saha~Joy, Ahmed, Sadek, Shil, and
  Kabir}]{shan2020depression}
Faisal~Muhammad Shah, Farzad Ahmed, Sajib~Kumar Saha~Joy, Sifat Ahmed, Samir
  Sadek, Rimon Shil, and Md.~Hasanul Kabir. 2020.
\newblock \href {https://doi.org/10.1109/TENSYMP50017.2020.9231008} {Early
  depression detection from social network using deep learning techniques}.
\newblock In \emph{2020 IEEE Region 10 Symposium (TENSYMP)}, pages 823--826.

\bibitem[{Smirnov et~al.(2021)Smirnov, Stankevich, Kuznetsova, Suvorova,
  Larionov, Nikitina, Savelov, and Grigoriev}]{smirnov2021titanis}
Ivan~V. Smirnov, Maksim Stankevich, Yulia Kuznetsova, Margarita Suvorova,
  Daniil Larionov, Elena Nikitina, Mikhail Savelov, and Oleg~G. Grigoriev.
  2021.
\newblock \href {https://doi.org/10.1007/978-3-030-86855-0\_16} {{TITANIS:} {A}
  tool for intelligent text analysis in social media}.
\newblock In \emph{Artificial Intelligence - 19th Russian Conference, {RCAI}
  2021, Taganrog, Russia, October 11-16, 2021, Proceedings}, pages 232--247.
  Springer.

\bibitem[{Stankevich et~al.(2019)Stankevich, Kuznetsova, Smirnov, Kiselnikova,
  and Enikolopov}]{stankevich2019predicting}
Maksim Stankevich, Yulia Kuznetsova, Ivan Smirnov, Natalia Kiselnikova, and
  Sergey Enikolopov. 2019.
\newblock Predicting depression from essays in russian.
\newblock In \emph{Komp'juternaja Lingvistika i Intellektual'nye Tehnologii},
  pages 647--657.

\bibitem[{Tadesse et~al.(2019)Tadesse, Lin, Xu, and
  Yang}]{tadesse2019depression}
Michael~M. Tadesse, Hongfei Lin, Bo~Xu, and Liang Yang. 2019.
\newblock \href {https://doi.org/10.1109/ACCESS.2019.2909180} {Detection of
  depression-related posts in reddit social media forum}.
\newblock \emph{IEEE Access}, 7:44883--44893.

\bibitem[{Team et~al.(2024)Team, Riviere, Pathak, Sessa, Hardin, Bhupatiraju,
  Hussenot, Mesnard, Shahriari, Ramé, Ferret, Liu, Tafti, Friesen, Casbon,
  Ramos, Kumar, Lan, Jerome, Tsitsulin, Vieillard, Stanczyk, Girgin, Momchev,
  Hoffman, Thakoor, Grill, Neyshabur, Bachem, Walton, Severyn, Parrish, Ahmad,
  Hutchison, Abdagic, Carl, Shen, Brock, Coenen, Laforge, Paterson, Bastian,
  Piot, Wu, Royal, Chen, Kumar, Perry, Welty, Choquette-Choo, Sinopalnikov,
  Weinberger, Vijaykumar, Rogozińska, Herbison, Bandy, Wang, Noland, Moreira,
  Senter, Eltyshev, Visin, Rasskin, Wei, Cameron, Martins, Hashemi,
  Klimczak-Plucińska, Batra, Dhand, Nardini, Mein, Zhou, Svensson, Stanway,
  Chan, Zhou, Carrasqueira, Iljazi, Becker, Fernandez, van Amersfoort, Gordon,
  Lipschultz, Newlan, yeong Ji, Mohamed, Badola, Black, Millican, McDonell,
  Nguyen, Sodhia, Greene, Sjoesund, Usui, Sifre, Heuermann, Lago, McNealus,
  Soares, Kilpatrick, Dixon, Martins, Reid, Singh, Iverson, Görner, Velloso,
  Wirth, Davidow, Miller, Rahtz, Watson, Risdal, Kazemi, Moynihan, Zhang,
  Kahng, Park, Rahman, Khatwani, Dao, Bardoliwalla, Devanathan, Dumai, Chauhan,
  Wahltinez, Botarda, Barnes, Barham, Michel, Jin, Georgiev, Culliton, Kuppala,
  Comanescu, Merhej, Jana, Rokni, Agarwal, Mullins, Saadat, Carthy, Cogan,
  Perrin, Arnold, Krause, Dai, Garg, Sheth, Ronstrom, Chan, Jordan, Yu, Eccles,
  Hennigan, Kocisky, Doshi, Jain, Yadav, Meshram, Dharmadhikari, Barkley, Wei,
  Ye, Han, Kwon, Xu, Shen, Gong, Wei, Cotruta, Kirk, Rao, Giang, Peran,
  Warkentin, Collins, Barral, Ghahramani, Hadsell, Sculley, Banks, Dragan,
  Petrov, Vinyals, Dean, Hassabis, Kavukcuoglu, Farabet, Buchatskaya, Borgeaud,
  Fiedel, Joulin, Kenealy, Dadashi, and Andreev}]{team2024gemma}
Gemma Team, Morgane Riviere, Shreya Pathak, Pier~Giuseppe Sessa, Cassidy
  Hardin, Surya Bhupatiraju, Léonard Hussenot, Thomas Mesnard, Bobak
  Shahriari, Alexandre Ramé, Johan Ferret, Peter Liu, Pouya Tafti, Abe
  Friesen, Michelle Casbon, Sabela Ramos, Ravin Kumar, Charline~Le Lan, Sammy
  Jerome, and 179 others. 2024.
\newblock \href {https://arxiv.org/abs/2408.00118} {Gemma 2: Improving open
  language models at a practical size}.
\newblock \emph{arXiv preprint arXiv:2408.00118}.

\bibitem[{Tejaswini et~al.(2024)Tejaswini, Babu, and
  Sahoo}]{Tejaswini2022DepressionDF}
Vankayala Tejaswini, Korra~Sathya Babu, and Bibhudatta Sahoo. 2024.
\newblock \href {https://doi.org/10.1145/3569580} {Depression detection from
  social media text analysis using natural language processing techniques and
  hybrid deep learning model}.
\newblock \emph{{ACM} Trans. Asian Low Resour. Lang. Inf. Process.},
  23(1):4:1--4:20.

\bibitem[{Wang et~al.(2024)Wang, Inkpen, and
  Kirinde~Gamaarachchige}]{wang-etal-2024-explainable}
Yuxi Wang, Diana Inkpen, and Prasadith Kirinde~Gamaarachchige. 2024.
\newblock \href {https://aclanthology.org/2024.clpsych-1.8} {Explainable
  depression detection using large language models on social media data}.
\newblock In \emph{Proceedings of the 9th Workshop on Computational Linguistics
  and Clinical Psychology (CLPsych 2024)}, pages 108--126, St. Julians, Malta.
  Association for Computational Linguistics.

\bibitem[{Wolf et~al.(2020)Wolf, Debut, Sanh, Chaumond, Delangue, Moi, Cistac,
  Rault, Louf, Funtowicz, Davison, Shleifer, von Platen, Ma, Jernite, Plu, Xu,
  Le~Scao, Gugger, Drame, Lhoest, and Rush}]{wolf-etal-2020-transformers}
Thomas Wolf, Lysandre Debut, Victor Sanh, Julien Chaumond, Clement Delangue,
  Anthony Moi, Pierric Cistac, Tim Rault, Remi Louf, Morgan Funtowicz, Joe
  Davison, Sam Shleifer, Patrick von Platen, Clara Ma, Yacine Jernite, Julien
  Plu, Canwen Xu, Teven Le~Scao, Sylvain Gugger, and 3 others. 2020.
\newblock \href {https://doi.org/10.18653/v1/2020.emnlp-demos.6} {Transformers:
  State-of-the-art natural language processing}.
\newblock In \emph{Proceedings of the 2020 Conference on Empirical Methods in
  Natural Language Processing: System Demonstrations}, pages 38--45, Online.
  Association for Computational Linguistics.

\bibitem[{Yalunin et~al.(2022)Yalunin, Nesterov, and
  Umerenkov}]{yalunin2022rubioroberta}
Alexander Yalunin, Alexander Nesterov, and Dmitriy Umerenkov. 2022.
\newblock \href {https://arxiv.org/abs/2204.03951} {Rubioroberta: a pre-trained
  biomedical language model for russian language biomedical text mining}.
\newblock \emph{arXiv preprint arXiv:2204.03951}.

\bibitem[{Yang et~al.(2024)Yang, Yang, Hui, Zheng, Yu, Zhou, Li, Li, Liu,
  Huang, Dong, Wei, Lin, Tang, Wang, Yang, Tu, Zhang, Ma, Yang, Xu, Zhou, Bai,
  He, Lin, Dang, Lu, Chen, Yang, Li, Xue, Ni, Zhang, Wang, Peng, Men, Gao, Lin,
  Wang, Bai, Tan, Zhu, Li, Liu, Ge, Deng, Zhou, Ren, Zhang, Wei, Ren, Liu, Fan,
  Yao, Zhang, Wan, Chu, Liu, Cui, Zhang, Guo, and Fan}]{yang2024qwen2}
An~Yang, Baosong Yang, Binyuan Hui, Bo~Zheng, Bowen Yu, Chang Zhou, Chengpeng
  Li, Chengyuan Li, Dayiheng Liu, Fei Huang, Guanting Dong, Haoran Wei, Huan
  Lin, Jialong Tang, Jialin Wang, Jian Yang, Jianhong Tu, Jianwei Zhang,
  Jianxin Ma, and 43 others. 2024.
\newblock \href {https://arxiv.org/abs/2407.10671} {Qwen2 technical report}.
\newblock \emph{arXiv preprint arXiv:2407.10671}.

\bibitem[{Zhang et~al.(2022)Zhang, Schoene, Ji, and
  Ananiadou}]{zhang2022natural}
Tianlin Zhang, Annika~Marie Schoene, Shaoxiong Ji, and Sophia Ananiadou. 2022.
\newblock \href {https://doi.org/10.1038/S41746-022-00589-7} {Natural language
  processing applied to mental illness detection: a narrative review}.
\newblock \emph{NPJ digital medicine}, 5(1):46.

\end{thebibliography}
\clearpage
\onecolumn

\appendix

\section{Hyperparameters}
\label{app:hyperpars}
The used net and optimal parameters, alongside the used checkpoint of encoder models, are presented in \Cref{tab:hp_transformers}. For the models trained on the psycholinguistic features and TF-IDF, the best model was selected with the AutoML pipeline, with the most models chosen as a Random Forest ensemble of up to four models.

The hyperparameters for LLMs, tuned with LoRA, are presented in \Cref{tab:hp_llms}. The LoRA was applied to the following target layers: \verb|q_proj|, \verb|up_proj|, \verb|o_proj|, \verb|k_proj|, \verb|down_proj|, \verb|gate_proj|, \verb|v_proj|. For LLMs 0-shot and 5-shot evaluation, we used the following generation parameters: \verb|do_sample=False|, \verb|temperature=1.0|, \verb|top_p=1.0|, \verb|top_k=50|, \verb|repetition_penalty=1.0|. The predicted classes were extracted using the string matching algorithm. For MMLU-style evaluation, the generation length was limited to one token, while for regular evaluation, it was set to 64 tokens. Prompt examples are presented in \Cref{tab:prompts}. For the 0-shot and 5-shot evaluation, we averaged the results on several prompts with various system parts to take into account the sensitivity of the generation with respect to the prompt. These prompt variations are presented in \Cref{tab:prompt_variations}. In all 0-shot and 5-shot experiments, we did not conduct best-of-N aggregation of answers and used one extracted answer per generation, which was averaged over several prompts to ensure generalizability of results.

The used hardware, as well as GPU hours, carbon footprint, and memory requirements are presented in \Cref{tab:gpu_hours}.

\begin{table*}
\centering
\resizebox{0.7\linewidth}{!}{
\begin{tabular}{llp{10cm}}
\toprule
\textbf{Task} &   \textbf{Mode}       & \textbf{Prompt example (system \& user)} \\
\midrule
 Depression & 0-shot & \textit{System:} You play the role of a psychologist's assistant who helps diagnose the presence or absence of a depressive disorder. You will be given a text written by a person. Determine the author's depression level from the text, where 0 is no depression, 1 is depression, and then write why you chose this answer.\newline \textit{User:} Text: \{input\_text\} \newline Answer (0 or 1):
\\

 Depression & 5-shot & \textit{System:} You play the role of a psychologist's assistant who helps diagnose the presence or absence of a depressive disorder. You will be given a text written by a person. Determine the author's depression level from the text, where 0 is no depression, 1 is depression, and then write why you chose this answer.\newline \textit{User:} Text: \{example\_1\} \newline Answer (0 or 1): 0 \newline Text: \{example\_2\} \newline Answer (0 or 1): 0 \newline Text: \{example\_3\} \newline Answer (0 or 1): 0 \newline Text: \{example\_4\} \newline Answer (0 or 1): 1 \newline Text: \{example\_5\} \newline Answer (0 or 1): 1 \newline Text: \{input\_text\} \newline Answer (0 or 1):
\\
\hline
 Anxiety & 0-shot & \textit{System:} You play the role of a psychologist's assistant who helps diagnose the presence or absence of an anxiety disorder. You will be given a text written by a person. Determine the level of anxiety of the author of the text, where 0 is no anxiety, 1 is anxiety, and then write why you chose this answer.\newline \textit{User:} Text: \{input\_text\} \newline Answer (0 or 1): \\

 Anxiety & 5-shot & \textit{System:} You play the role of a psychologist's assistant who helps diagnose the presence or absence of an anxiety disorder. You will be given a text written by a person. Determine the level of anxiety of the author of the text, where 0 is no anxiety, 1 is anxiety, and then write why you chose this answer.\newline \textit{User:} Text: \{example\_1\} \newline Answer (0 or 1): 0 \newline Text: \{example\_2\} \newline Answer (0 or 1): 0 \newline Text: \{example\_3\} \newline Answer (0 or 1): 0 \newline Text: \{example\_4\} \newline Answer (0 or 1): 1 \newline Text: \{example\_5\} \newline Answer (0 or 1): 1 \newline Text: \{input\_text\} \newline Answer (0 or 1):\\
\bottomrule
\end{tabular}
}\caption{Prompts for LLMs evaluation. For a better understanding, we present a translated English version (originally we used the same prompts in Russian). For models without system prompts (e. g. Gemma2) we concatenated the system prompt to the user prompt. Text in italics was used to denote system and user roles and was replaced by model-specific templates during evaluation.}
\label{tab:prompts}
\end{table*}
\begin{table*}
\centering
\resizebox{0.7\linewidth}{!}{
\begin{tabular}{llp{10cm}}
\toprule
\textbf{Task} &   \textbf{Prompt number}       & \textbf{Prompt example (only system)} \\
\midrule
 Depression & 1 & \textit{System:} Read the provided text and determine whether the author has signs of depression. Use the scale: 0 - no depression, 1 - depression. Then explain why you chose this option. 
\\

 Depression & 2 & \textit{System:} Evaluate the text for the author's depressive state. Scale: 0 - no depression, 1 - depression is present. Justify your choice after indicating the answer.
\\
Depression & 3 & \textit{System:} Analyze this text and identify the presence of depressive manifestations in its author. Use a binary assessment: 0 - no depression, 1 - presence of depression. Provide a rationale for your decision. 
\\
Depression & 4 & \textit{System:} Assess the psychological state of the author of the text in the context of depression. Use the gradation: 0 - no signs of depression, 1 - there are signs of depression. Detail the reasons for making your decision. 
\\
\hline
 Anxiety & 1 & \textit{System:} Read the provided text and determine whether the author has signs of anxiety. Use the scale: 0 - no anxiety, 1 - anxiety. Then explain why you chose this option. \\

 Anxiety & 2 & \textit{System:} Evaluate the text for the author's anxiety. Scale: 0 - no anxiety, 1 - anxiety is present. Justify your choice after indicating the answer. \\

 Anxiety & 3 & \textit{System:} Analyze this text and identify the presence of anxiety in its author. Use a binary assessment: 0 - no anxiety, 1 - anxiety present. Justify your decision. \\

 Anxiety & 4 & \textit{System:} Assess the psychological state of the author of the text in the context of anxiety. Use the gradation: 0 - no signs of anxiety, 1 - signs of anxiety present. Detail the reasons for making your decision. \\
\bottomrule
\end{tabular}
}\caption{Additional prompts used to average results for LLMs evaluation. Here we present only the examples of system prompts, because the other prompt parts remained unchanged as in \Cref{tab:prompts}.}
\label{tab:prompt_variations}
\end{table*}
\begin{table*}
\centering
\resizebox{0.5\linewidth}{!}{
\begin{tabular}{llllll}
\toprule
\textbf{Model name} &   \textbf{Corpus}       &  \begin{tabular}[c]{@{}l@{}} \textbf{Num. of} \\ \textbf{epochs}\end{tabular} & \begin{tabular}[c]{@{}l@{}} \textbf{Learning} \\ \textbf{rate}\end{tabular} & \begin{tabular}[c]{@{}l@{}} \textbf{Batch} \\ \textbf{size}\end{tabular} & \begin{tabular}[c]{@{}l@{}} \textbf{Weight} \\ \textbf{decay}\end{tabular} \\
\midrule
    \multirow{7}{*}{RuBERT} & A-all  &                15 &         5e-05 &         16 &          0.10 \\
    ~ & D-all  &                15 &         5e-05 &         16 &          0.10 \\
   ~ & DE  &                10 &         7e-05 &         32 &          0.10 \\
  ~ & AD  &                 2 &         6e-06 &          4 &          0.10 \\
  ~ & AL  &                 2 &         9e-06 &          4 &          0.10 \\
   ~ & AC  &                 2 &         6e-06 &          4 &          0.10 \\
   ~ & DSM  &                13 &         1e-04 &          4 &          0.00 \\
%   \hline
%    K-all &             RuBERT &    F1-macro &          11 &         1e-04 &          4 &          0.0 \\
%    D-all &             RuBERT &    F1-macro &          15 &         5e-05 &         16 &          0.1 \\
%   DE &             RuBERT &    F1-macro &           8 &         7e-05 &         32 &         0.01 \\
%  AD &             RuBERT &    F1-macro &          12 &        0.0001 &          4 &          0.0 \\
%  AL &             RuBERT &    F1-macro &           7 &         9e-06 &         32 &         0.01 \\
%   AC &             RuBERT &    F1-macro &           4 &         1e-05 &          4 &          0.0 \\
%   DSM &             RuBERT &    F1-macro &           8 &         1e-05 &          4 &          0.0 \\
   \hline
   \multirow{7}{*}{RuRoBERTa} & A-all  &                13 &         9e-06 &         16 &          0.01 \\
    ~ & D-all  &                11 &         6e-06 &         8 &          0.00 \\
   ~ & DE  &                9 &         9e-06 &         4 &          0.00 \\
  ~ & AD  &                 7 &         2e-05 &          4 &          0.00 \\
  ~ & AL  &                 11 &         6e-06 &          8 &          0.00 \\
   ~ & AC  &                 6 &         1e-04 &          32 &          0.01 \\
   ~ & DSM  &                13 &         5e-06 &          16 &          0.10 \\
   \hline
   \multirow{7}{*}{RuBioRoBERTa} & A-all  &           15 &         3e-05 &         16 &         0.01 \\
    ~ & D-all  &             11 &         7e-06 &         32 &          0.00 \\
   ~ & DE  &           8 &         1e-05 &          4 &          0.00 \\
  ~ & AD  &               2 &         7e-06 &          8 &          0.10 \\
 ~ &  AL  &                 9 &         2e-05 &         16 &         0.01 \\
   ~ & AC  &                 8 &         1e-05 &          4 &          0.00 \\
   ~ & DSM  &                15 &         9e-06 &         16 &         0.01 \\
   \hline
%    A-all & RuBioRoBERTa &    F1-macro &          10 &         7e-05 &         32 &          0.1 \\
%    D-all & RuBioRoBERTa &    F1-macro &           9 &         7e-06 &         32 &          0.0 \\
%   DE & RuBioRoBERTa &    F1-macro &           8 &         1e-05 &         16 &         0.01 \\
%  AD & RuBioRoBERTa &    F1-macro &           2 &         3e-05 &          8 &          0.1 \\
%  AL & RuBioRoBERTa &    F1-macro &          11 &         6e-06 &          8 &          0.0 \\
%   AC & RuBioRoBERTa &    F1-macro &          10 &         7e-05 &         32 &          0.1 \\
%   DSM & RuBioRoBERTa &    F1-macro &           4 &         2e-05 &         32 &         0.01 \\
%   \hline
    \multirow{7}{*}{BERT} & A-all  &                14 &         7e-06 &         16 &         0.01 \\
    ~ & D-all  &                15 &         5e-05 &         16 &          0.10 \\
   ~ & DE  &                15 &         5e-05 &         16 &          0.10 \\
  ~ & AD  &                 2 &         6e-06 &          4 &          0.10 \\
  ~ & AL  &                 2 &         6e-06 &          4 &          0.10 \\
   ~ & AC  &                 4 &         1e-04 &         32 &         0.01 \\
   ~ & DSM  &                 5 &         1e-04 &         32 &         0.01 \\
%   \hline
%    A-all &  BERT &    F1-macro &           2 &         7e-06 &          8 &          0.1 \\
%    D-all &  BERT &    F1-macro &          10 &         7e-05 &         32 &          0.1 \\
%   DE &  BERT &    F1-macro &          15 &         5e-05 &         16 &          0.1 \\
%  AD &  BERT &    F1-macro &          12 &         5e-06 &          4 &          0.1 \\
%  AL &  BERT &    F1-macro &           5 &         7e-05 &         32 &          0.1 \\
%   AC &  BERT &    F1-macro &           2 &         9e-06 &          4 &          0.1 \\
%   DSM &  BERT &    F1-macro &          15 &         6e-06 &         16 &          0.1 \\

\bottomrule
\end{tabular}
}\caption{Optimal hyperparameters for transformer models. We used the following net for hyperparameters tuning: learning rate - [5e-6, 6e-6, 7e-6, 9e-6, 1e-5, 2e-5, 3e-5, 5e-5, 7e-5, 1e-4], num. of epochs - from 2 to 15, batch size - [4, 8, 16, 32], weight decay - [0, 0.01, 0.1].}
\label{tab:hp_transformers}
\end{table*}
\begin{table*}
\centering
\resizebox{0.7\linewidth}{!}{
\begin{tabular}{lllllllll}
\toprule
\textbf{Model name} &   \textbf{Corpus}       &  \begin{tabular}[c]{@{}l@{}} \textbf{Num. of} \\ \textbf{epochs}\end{tabular} & \begin{tabular}[c]{@{}l@{}} \textbf{Learning} \\ \textbf{rate}\end{tabular} & \begin{tabular}[c]{@{}l@{}} \textbf{Batch} \\ \textbf{size}\end{tabular} & \begin{tabular}[c]{@{}l@{}} \textbf{Weight} \\ \textbf{decay} \end{tabular} & \textbf{LoRA $\alpha$} &  \begin{tabular}[c]{@{}l@{}} \textbf{LoRA} \\ \textbf{dropout} \end{tabular} &  \begin{tabular}[c]{@{}l@{}} \textbf{LoRA} \\ \textbf{rank} \end{tabular} \\
\midrule
 \multirow{7}{*}{SaigaLlama3 8B} & DE & 13 & 1e-04 & 4 & 1e-01 & 16 & 1e-01 & 8 \\
~ & DSM & 3 & 1e-05 & 16 & 1e-02 & 32 & 5e-02 & 16 \\
~ & AL & 15 & 1e-04 & 4 & 1e-01 & 16 & 1e-01 & 16 \\
~ & AD & 15 & 1e-04 & 4 & 1e-01 & 16 & 1e-01 & 16 \\
~ & AC & 11 & 7e-05 & 16 & 1e-01 & 32 & 5e-02 & 8 \\
~ & K-all & 14 & 3e-05 & 16 & 0e+00 & 16 & 5e-02 & 16 \\
~ & D-all & 11 & 7e-05 & 16 & 1e-01 & 32 & 5e-02 & 8 \\
\hline
\multirow{7}{*}{Vikhr 7B IT 0.4} & DE & 2 & 5e-05 & 4 & 1e-01 & 32 & 5e-02 & 8 \\
~ & DSM & 11 & 7e-05 & 16 & 1e-01 & 32 & 5e-02 & 8 \\
~ & AL & 4 & 9e-06 & 8 & 1e-02 & 16 & 1e-01 & 16 \\
~ & AD & 11 & 1e-04 & 4 & 0e+00 & 32 & 1e-01 & 16 \\
~ & AC & 3 & 1e-04 & 16 & 0e+00 & 16 & 5e-02 & 16 \\
~ & K-all & 11 & 2e-05 & 16 & 1e-02 & 16 & 1e-01 & 16 \\
~ & D-all & 11 & 7e-05 & 16 & 1e-01 & 32 & 5e-02 & 8 \\
\hline
\multirow{7}{*}{Vikhr 7B IT 5.4} & DE & 15 & 1e-04 & 4 & 1e-01 & 16 & 1e-01 & 16 \\
~ & DSM & 4 & 9e-06 & 8 & 1e-02 & 16 & 1e-01 & 16 \\
~ & AL & 7 & 6e-06 & 8 & 0e+00 & 16 & 1e-01 & 8 \\
~ & AD & 8 & 5e-06 & 4 & 1e-02 & 16 & 1e-01 & 16 \\
~ & AC & 15 & 1e-04 & 4 & 0e+00 & 16 & 1e-01 & 8 \\
~ & K-all & 12 & 3e-05 & 16 & 0e+00 & 16 & 1e-01 & 8 \\
~ & D-all & 8 & 5e-06 & 4 & 1e-02 & 16 & 1e-01 & 16 \\
\hline
\multirow{7}{*}{VikhrGemma 2B IT} & DE & 15 & 1e-04 & 4 & 0e+00 & 16 & 1e-01 & 8 \\
~ & DSM & 2 & 6e-06 & 8 & 0e+00 & 16 & 1e-01 & 16 \\
~ & AL & 5 & 5e-05 & 4 & 1e-02 & 16 & 5e-02 & 16 \\
~ & AD & 8 & 2e-05 & 4 & 0e+00 & 32 & 5e-02 & 16 \\
~ & AC & 2 & 5e-05 & 4 & 1e-01 & 32 & 5e-02 & 8 \\
~ & K-all & 11 & 2e-05 & 16 & 1e-02 & 16 & 1e-01 & 16 \\
~ & D-all & 15 & 1e-04 & 4 & 1e-01 & 16 & 1e-01 & 16 \\
\hline
\multirow{7}{*}{Gemma2 2B IT} & DE & 15 & 1e-04 & 4 & 1e-01 & 16 & 1e-01 & 16 \\
~ & DSM & 8 & 7e-05 & 16 & 0e+00 & 32 & 5e-02 & 8 \\
~ & AL & 13 & 1e-04 & 4 & 0e+00 & 16 & 1e-01 & 8 \\
~ & AD & 12 & 7e-05 & 16 & 0e+00 & 32 & 5e-02 & 8 \\
~ & AC & 14 & 1e-04 & 4 & 0e+00 & 32 & 1e-01 & 16 \\
~ & K-all & 11 & 7e-05 & 16 & 1e-01 & 32 & 5e-02 & 8 \\
~ & D-all & 15 & 1e-04 & 4 & 1e-01 & 16 & 1e-01 & 16 \\
\hline
\multirow{7}{*}{Gemma2 9B IT} & DE & 9 & 1e-04 & 8 & 0e+00 & 16 & 5e-02 & 8 \\
~ & DSM & 13 & 5e-06 & 8 & 1e-02 & 16 & 1e-01 & 16 \\
~ & AL & 6 & 1e-04 & 8 & 0e+00 & 16 & 1e-01 & 16 \\
~ & AD & 11 & 2e-05 & 16 & 1e-02 & 16 & 1e-01 & 16 \\
~ & AC & 15 & 1e-04 & 4 & 0e+00 & 16 & 1e-01 & 8 \\
~ & K-all & 11 & 7e-05 & 16 & 1e-01 & 32 & 5e-02 & 8 \\
~ & D-all & 15 & 1e-04 & 4 & 0e+00 & 16 & 1e-01 & 8 \\
\hline
\multirow{7}{*}{Qwen2 7B IT} & DE & 5 & 5e-05 & 4 & 1e-02 & 16 & 5e-02 & 16 \\
~ & DSM & 11 & 7e-05 & 16 & 1e-01 & 32 & 5e-02 & 8 \\
~ & AL & 4 & 5e-05 & 8 & 0e+00 & 16 & 5e-02 & 8 \\
~ & AD & 11 & 7e-05 & 16 & 1e-01 & 32 & 5e-02 & 8 \\
~ & AC & 9 & 3e-05 & 4 & 0e+00 & 32 & 1e-01 & 8 \\
~ & K-all & 11 & 7e-05 & 16 & 1e-01 & 32 & 5e-02 & 8 \\
~ & D-all & 13 & 1e-04 & 8 & 1e-01 & 32 & 5e-02 & 8 \\
\bottomrule
\end{tabular}
}\caption{Optimal hyperparameters for LLMs with LoRA. We used the following net for hyperparameters tuning: learning rate - [5e-6, 6e-6, 7e-6, 9e-6, 1e-5, 2e-5, 3e-5, 5e-5, 7e-5, 1e-4], num. of epochs - from 2 to 15, batch size - [4, 8, 16, 32], weight decay - [0, 0.01, 0.1], LoRA $\alpha$ - [16, 32], LoRA dropout - [0.05, 0.1], LoRA rank - [8,16].}
\label{tab:hp_llms}
\end{table*}
\begin{table*}
\centering
\resizebox{1.0\linewidth}{!}{
\begin{tabular}{lllll}
\toprule
\textbf{GPU type} &   \textbf{NVIDIA V100 32GB}      &  \textbf{NVIDIA A100 80 GB} & \textbf{NVIDIA H100 80 GB} & \textbf{Total} \\
\midrule
GPU Hours & 119  &  371 & 43 & 533 \\
Carbon footprint, kg CO$_{2}$ & 11.50  &  29.87 & 3.46   &    44.83 \\ 
\bottomrule
\end{tabular}
}\caption{The approximate number of GPU hours and carbon footprint for all experiments.}
\label{tab:gpu_hours}
\end{table*}

\section{Full Results for Models}
\label{app:full_res}
\Cref{tab:results_main_exp} shows results for TF-IDF, models on linguistic features, and encoder-based transformers. \Cref{tab:results_llm_peft,tab:results_llm_reg,tab:results_llm_mmlu} contain results for LLMs with PEFT, 0-shot and 5-shot evaluation, and 0-shot and 5-shot MMLU-style evaluation correspondingly.

\begin{table*}[t]
\centering
\resizebox{\linewidth}{!}{%
    \begin{tabular}{lllllllll}
    \hline
    \textbf{Corpus} & \textbf{Model} & \begin{tabular}[c]{@{}l@{}} \textbf{Precision} \\ \textbf{healthy}\end{tabular} & \begin{tabular}[c]{@{}l@{}} \textbf{Recall} \\ \textbf{healthy}\end{tabular} & \textbf{F1-healthy} & \begin{tabular}[c]{@{}l@{}} \textbf{Precision} \\ \textbf{pathology}\end{tabular} & \begin{tabular}[c]{@{}l@{}} \textbf{Recall} \\ \textbf{pathology}\end{tabular} & \textbf{F1-pathology} & \textbf{F1-macro}  \\
    \hline
    \multirow{7}{*}{DE} & RuRoBERTa & 90.44\tiny{$\pm$ 4.62} & 95.93\tiny{$\pm$ 2.28} & 92.98\tiny{$\pm$ 1.92} & 64.70\tiny{$\pm$ 29.29} & 56.82\tiny{$\pm$ 25.81} & 60.37\tiny{$\pm$ 27.21} & 76.67\tiny{$\pm$ 14.52} \\
    ~ & RuBioRoBERTa & 91.92\tiny{$\pm$ 5.21} & 95.19\tiny{$\pm$ 3.05} & 93.36\tiny{$\pm$ 2.13} & 64.25\tiny{$\pm$ 29.33} & 63.64\tiny{$\pm$ 28.63} & 63.73\tiny{$\pm$ 28.62} & 78.54\tiny{$\pm$ 15.30} \\
    ~ & RuBERT & 94.45\tiny{$\pm$ 1.16} & 91.11\tiny{$\pm$ 1.70} & 92.74\tiny{$\pm$ 1.04} & 68.41\tiny{$\pm$ 4.04} & 78.03\tiny{$\pm$ 4.85} & 72.80\tiny{$\pm$ 3.43} & 82.77\tiny{$\pm$ 2.21} \\
    ~ & BERT & 92.97\tiny{$\pm$ 0.80} & 92.96\tiny{$\pm$ 1.23} & 92.96\tiny{$\pm$ 0.85} & 71.34\tiny{$\pm$ 3.96} & 71.21\tiny{$\pm$ 3.39} & 71.23\tiny{$\pm$ 3.14} & 82.09\tiny{$\pm$ 1.98} \\
    ~ & Linguistic features & 94.00\tiny{$\pm$ 0.80} & 95.20\tiny{$\pm$ 1.00} & 94.60\tiny{$\pm$ 0.80} & 79.30\tiny{$\pm$ 4.00} & 75.00\tiny{$\pm$ 3.50} & \textbf{77.00\tiny{$\pm$ 3.30}} & \textbf{85.80\tiny{$\pm$ 2.10}}  \\ 
    ~ & TF-IDF & 91.60\tiny{$\pm$ 2.10} & 94.40\tiny{$\pm$ 2.20} & 93.00\tiny{$\pm$ 1.20} & 74.80\tiny{$\pm$ 7.00} & 64.40\tiny{$\pm$ 10.30} & 68.50\tiny{$\pm$ 6.60} & 80.70\tiny{$\pm$ 3.80}  \\ \hline
    \multirow{7}{*}{DSM} & RuRoBERTa & 61.31\tiny{$\pm$ 4.38} & 73.46\tiny{$\pm$ 12.73} & 66.58\tiny{$\pm$ 7.66} & 46.76\tiny{$\pm$ 10.29} & 31.48\tiny{$\pm$ 6.93} & 36.68\tiny{$\pm$ 6.20} & 51.63\tiny{$\pm$ 5.80} \\
    ~ & RuBioRoBERTa & 62.10\tiny{$\pm$ 2.98} & 62.96\tiny{$\pm$ 6.05} & 62.46\tiny{$\pm$ 4.30} & 43.66\tiny{$\pm$ 4.43} & 42.59\tiny{$\pm$ 4.14} & 43.01\tiny{$\pm$ 3.70} & 52.74\tiny{$\pm$ 3.67} \\
    ~ & RuBERT & 60.78\tiny{$\pm$ 1.75} & 96.91\tiny{$\pm$ 6.90} & 74.52\tiny{$\pm$ 1.07} & 9.09\tiny{$\pm$ 20.33} & 5.56\tiny{$\pm$ 12.42} & 6.90\tiny{$\pm$ 15.42} & 40.71\tiny{$\pm$ 7.18} \\
    ~ & BERT & 61.09\tiny{$\pm$ 1.90} & 69.14\tiny{$\pm$ 14.61} & 64.13\tiny{$\pm$ 5.24} & 34.59\tiny{$\pm$ 15.71} & 33.33\tiny{$\pm$ 16.97} & 33.71\tiny{$\pm$ 16.01} & 48.92\tiny{$\pm$ 5.62} \\
    ~ & Linguistic features & 62.80\tiny{$\pm$ 1.80} & 59.30\tiny{$\pm$ 7.40} & 60.70\tiny{$\pm$ 4.00} & 43.70\tiny{$\pm$ 2.70} & 47.20\tiny{$\pm$ 7.70} & 45.10\tiny{$\pm$ 3.80} & \textbf{52.90\tiny{$\pm$ 2.00}}  \\ 
    ~ & TF-IDF & 62.20\tiny{$\pm$ 2.60} & 51.20\tiny{$\pm$ 13.90} & 55.30\tiny{$\pm$ 8.50} & 42.90\tiny{$\pm$ 3.60} & 53.70\tiny{$\pm$ 11.40} & \textbf{46.90\tiny{$\pm$ 4.70}} & 51.10\tiny{$\pm$ 3.50}  \\ \hline
    \multirow{7}{*}{AL} & RuRoBERTa & 56.82\tiny{$\pm$ 5.65} & 75.00\tiny{$\pm$ 7.76} & 64.49\tiny{$\pm$ 5.59} & 52.54\tiny{$\pm$ 10.76} & 33.33\tiny{$\pm$ 12.77} & 40.17\tiny{$\pm$ 12.48} & 52.33\tiny{$\pm$ 8.31} \\
    ~ & RuBioRoBERTa & 53.62\tiny{$\pm$ 1.09} & 78.79\tiny{$\pm$ 15.67} & 63.22\tiny{$\pm$ 5.04} & 30.66\tiny{$\pm$ 21.79} & 21.05\tiny{$\pm$ 15.79} & 24.76\tiny{$\pm$ 18.06} & 43.99\tiny{$\pm$ 6.71} \\
    ~ & RuBERT & 53.63\tiny{$\pm$ 3.20} & 77.27\tiny{$\pm$ 19.99} & 62.08\tiny{$\pm$ 7.52} & 45.83\tiny{$\pm$ 29.56} & 21.93\tiny{$\pm$ 20.48} & 24.65\tiny{$\pm$ 17.57} & 43.36\tiny{$\pm$ 5.79} \\
    ~ & BERT & 52.71\tiny{$\pm$ 1.02} & 93.94\tiny{$\pm$ 8.16} & 67.44\tiny{$\pm$ 2.96} & 6.25\tiny{$\pm$ 13.98} & 2.63\tiny{$\pm$ 5.88} & 3.70\tiny{$\pm$ 8.28} & 35.57\tiny{$\pm$ 2.99} \\
    ~ & Linguistic features & 47.40\tiny{$\pm$ 21.40} & 36.40\tiny{$\pm$ 17.80} & 40.90\tiny{$\pm$ 19.20} & 48.50\tiny{$\pm$ 2.70} & 68.40\tiny{$\pm$ 14.60} & 56.10\tiny{$\pm$ 3.70} & 48.50\tiny{$\pm$ 8.20}  \\ 
    ~ & TF-IDF & 61.50\tiny{$\pm$ 5.00} & 46.20\tiny{$\pm$ 3.10} & 52.60\tiny{$\pm$ 2.30} & 51.20\tiny{$\pm$ 2.50} & 65.80\tiny{$\pm$ 7.90} & \textbf{57.50\tiny{$\pm$ 4.50}} & \textbf{55.00\tiny{$\pm$ 2.90}}  \\ \hline 
    \multirow{7}{*}{AD} & RuRoBERTa & 57.63\tiny{$\pm$ 2.52} & 56.67\tiny{$\pm$ 4.71} & 57.07\tiny{$\pm$ 3.19} & 52.80\tiny{$\pm$ 2.82} & 53.70\tiny{$\pm$ 4.14} & \textbf{53.17\tiny{$\pm$ 2.84}} & \textbf{55.12\tiny{$\pm$ 2.58}} \\
    ~ & RuBioRoBERTa & 45.06\tiny{$\pm$ 5.52} & 43.33\tiny{$\pm$ 19.72} & 41.49\tiny{$\pm$ 13.80} & 38.29\tiny{$\pm$ 8.58} & 40.74\tiny{$\pm$ 23.28} & 37.72\tiny{$\pm$ 12.46} & 39.60\tiny{$\pm$ 5.38} \\
    ~ & RuBERT & 48.63\tiny{$\pm$ 7.83} & 80.00\tiny{$\pm$ 29.58} & 59.34\tiny{$\pm$ 15.36} & 12.59\tiny{$\pm$ 18.14} & 12.96\tiny{$\pm$ 18.33} & 12.70\tiny{$\pm$ 18.04} & 36.02\tiny{$\pm$ 4.94} \\
    ~ & BERT & 52.03\tiny{$\pm$ 1.00} & 93.33\tiny{$\pm$ 7.99} & 66.74\tiny{$\pm$ 2.88} & 20.56\tiny{$\pm$ 21.12} & 4.63\tiny{$\pm$ 4.99} & 7.34\tiny{$\pm$ 7.71} & 37.04\tiny{$\pm$ 2.63} \\
    ~ & Linguistic features & 50.80\tiny{$\pm$ 3.30} & 43.30\tiny{$\pm$ 12.50} & 45.30\tiny{$\pm$ 8.10} & 44.70\tiny{$\pm$ 2.60} & 51.90\tiny{$\pm$ 15.90} & 47.30\tiny{$\pm$ 7.40} & 46.30\tiny{$\pm$ 1.80}  \\ 
    ~ & TF-IDF & 54.90\tiny{$\pm$ 15.10} & 37.50\tiny{$\pm$ 4.80} & 43.20\tiny{$\pm$ 2.50} & 44.60\tiny{$\pm$ 7.80} & 59.30\tiny{$\pm$ 21.00} & 50.40\tiny{$\pm$ 12.70} & 46.80\tiny{$\pm$ 7.10}  \\ \hline 
    \multirow{7}{*}{AC} & RuRoBERTa & 45.70\tiny{$\pm$ 20.48} & 77.04\tiny{$\pm$ 35.48} & 57.19\tiny{$\pm$ 25.66} & 24.57\tiny{$\pm$ 24.96} & 24.56\tiny{$\pm$ 35.49} & 21.22\tiny{$\pm$ 23.82} & 39.20\tiny{$\pm$ 7.55} \\
    ~ & RuBioRoBERTa & 52.92\tiny{$\pm$ 1.41} & 76.30\tiny{$\pm$ 14.15} & 62.09\tiny{$\pm$ 5.54} & 34.88\tiny{$\pm$ 15.72} & 20.18\tiny{$\pm$ 11.64} & 24.84\tiny{$\pm$ 12.88} & 43.47\tiny{$\pm$ 4.21} \\
    ~ & RuBERT & 54.29\tiny{$\pm$ 1.57} & 87.04\tiny{$\pm$ 19.35} & 66.00\tiny{$\pm$ 8.24} & 43.59\tiny{$\pm$ 22.45} & 14.04\tiny{$\pm$ 17.10} & 17.15\tiny{$\pm$ 15.70} & 41.58\tiny{$\pm$ 4.88} \\
    ~ & BERT & 57.01\tiny{$\pm$ 5.15} & 62.96\tiny{$\pm$ 16.81} & 58.44\tiny{$\pm$ 6.53} & 46.07\tiny{$\pm$ 8.77} & 41.67\tiny{$\pm$ 22.14} & 41.28\tiny{$\pm$ 17.15} & 49.86\tiny{$\pm$ 6.84} \\
    ~ & Linguistic features & 64.80\tiny{$\pm$ 16.00} & 45.60\tiny{$\pm$ 21.00} & 47.00\tiny{$\pm$ 19.60} & 48.70\tiny{$\pm$ 3.60} & 60.50\tiny{$\pm$ 20.10} & \textbf{52.60\tiny{$\pm$ 7.30}} & 49.80\tiny{$\pm$ 7.80}  \\ 
    ~ & TF-IDF & 56.30\tiny{$\pm$ 2.10} & 63.30\tiny{$\pm$ 4.20} & 59.50\tiny{$\pm$ 2.50} & 48.90\tiny{$\pm$ 3.30} & 41.70\tiny{$\pm$ 5.60} & 44.90\tiny{$\pm$ 4.30} & \textbf{52.20\tiny{$\pm$ 2.70}}  \\
    \hline
    \end{tabular}
    }
    
\caption{The results for encoder models and AutoML models on the five main datasets.}
\label{tab:results_main_exp}
\end{table*}
\begin{table*}[t]
\centering
\resizebox{\linewidth}{!}{%
    \begin{tabular}{lllllllll}
    \hline
    %Corpus & Model & \begin{tabular}[c]{@{}l@{}} Precision \\ healthy\end{tabular} & \begin{tabular}[c]{@{}l@{}} Recall \\ healthy\end{tabular} & F1-healthy & \begin{tabular}[c]{@{}l@{}} Precision \\ pathology\end{tabular} & \begin{tabular}[c]{@{}l@{}} Recall \\ pathology\end{tabular} & F1-pathology & F1-macro  \\
    \textbf{Corpus} & \textbf{Model} & \begin{tabular}[c]{@{}l@{}} \textbf{Precision} \\ \textbf{healthy}\end{tabular} & \begin{tabular}[c]{@{}l@{}} \textbf{Recall} \\ \textbf{healthy}\end{tabular} & \textbf{F1-healthy} & \begin{tabular}[c]{@{}l@{}} \textbf{Precision} \\ \textbf{pathology}\end{tabular} & \begin{tabular}[c]{@{}l@{}} \textbf{Recall} \\ \textbf{pathology}\end{tabular} & \textbf{F1-pathology} & \textbf{F1-macro}  \\
    \hline
    \multirow{7}{*}{DE} & Gemma2 2B IT & 92.96\tiny{$\pm$ 1.39} & 97.22\tiny{$\pm$ 1.40} & 95.02\tiny{$\pm$ 0.48} & 86.63\tiny{$\pm$ 4.79} & 69.70\tiny{$\pm$ 6.78} & 76.84\tiny{$\pm$ 3.12} & 85.93\tiny{$\pm$ 1.75} \\
~ & Gemma2 9B IT & 91.90\tiny{$\pm$ 0.90} & 96.48\tiny{$\pm$ 1.00} & 94.13\tiny{$\pm$ 0.58} & 82.11\tiny{$\pm$ 3.76} & 65.15\tiny{$\pm$ 4.29} & 72.53\tiny{$\pm$ 2.91} & 83.33\tiny{$\pm$ 1.72} \\
~ & Qwen2 7B IT & 90.18\tiny{$\pm$ 0.34} & 95.19\tiny{$\pm$ 2.10} & 92.60\tiny{$\pm$ 0.96} & 75.46\tiny{$\pm$ 7.33} & 57.58\tiny{$\pm$ 2.14} & 65.05\tiny{$\pm$ 2.19} & 78.83\tiny{$\pm$ 1.56} \\
~ & SaigaLlama3 8B & 91.59\tiny{$\pm$ 0.79} & 98.70\tiny{$\pm$ 1.19} & 95.01\tiny{$\pm$ 0.41} & 92.88\tiny{$\pm$ 5.97} & 62.88\tiny{$\pm$ 4.08} & 74.74\tiny{$\pm$ 2.12} & 84.87\tiny{$\pm$ 1.22} \\
~ & Vikhr 7B IT 0.4 & 90.26\tiny{$\pm$ 1.15} & 94.07\tiny{$\pm$ 2.77} & 92.10\tiny{$\pm$ 1.11} & 72.82\tiny{$\pm$ 12.43} & 58.33\tiny{$\pm$ 6.11} & 63.92\tiny{$\pm$ 3.46} & 78.01\tiny{$\pm$ 2.17} \\
~ & Vikhr 7B IT 5.4 & 90.80\tiny{$\pm$ 2.36} & 95.74\tiny{$\pm$ 2.52} & 93.15\tiny{$\pm$ 1.15} & 79.07\tiny{$\pm$ 9.93} & 59.85\tiny{$\pm$ 12.14} & 66.91\tiny{$\pm$ 8.35} & 80.03\tiny{$\pm$ 4.65} \\
~ & VikhrGemma 2B IT & 94.74\tiny{$\pm$ 0.88} & 96.48\tiny{$\pm$ 1.62} & 95.59\tiny{$\pm$ 0.66} & 84.96\tiny{$\pm$ 5.72} & 78.03\tiny{$\pm$ 4.08} & \textbf{81.13\tiny{$\pm$ 2.42}} & \textbf{88.36\tiny{$\pm$ 1.52}} \\ \hline
    \multirow{7}{*}{DSM} & Gemma2 2B IT & 63.29\tiny{$\pm$ 4.98} & 58.64\tiny{$\pm$ 9.18} & 60.62\tiny{$\pm$ 6.33} & 44.42\tiny{$\pm$ 5.41} & 49.07\tiny{$\pm$ 9.31} & 46.24\tiny{$\pm$ 6.49} & 53.43\tiny{$\pm$ 5.33} \\
~ & Gemma2 9B IT & 64.20\tiny{$\pm$ 3.12} & 56.79\tiny{$\pm$ 6.65} & 60.16\tiny{$\pm$ 4.97} & 45.12\tiny{$\pm$ 4.14} & 52.78\tiny{$\pm$ 4.24} & 48.54\tiny{$\pm$ 3.57} & 54.35\tiny{$\pm$ 3.90} \\
~ & Qwen2 7B IT & 68.87\tiny{$\pm$ 6.39} & 72.22\tiny{$\pm$ 8.21} & 70.40\tiny{$\pm$ 6.75} & 55.33\tiny{$\pm$ 11.08} & 50.93\tiny{$\pm$ 10.84} & \textbf{52.81\tiny{$\pm$ 10.26}} & \textbf{61.61\tiny{$\pm$ 8.32}} \\
~ & SaigaLlama3 8B & 57.46\tiny{$\pm$ 7.16} & 56.17\tiny{$\pm$ 15.48} & 56.13\tiny{$\pm$ 10.59} & 37.43\tiny{$\pm$ 9.05} & 38.89\tiny{$\pm$ 13.98} & 37.21\tiny{$\pm$ 10.11} & 46.67\tiny{$\pm$ 7.70} \\
~ & Vikhr 7B IT 0.4 & 64.12\tiny{$\pm$ 2.52} & 70.99\tiny{$\pm$ 11.40} & 67.06\tiny{$\pm$ 5.90} & 50.75\tiny{$\pm$ 11.53} & 40.74\tiny{$\pm$ 7.64} & 44.07\tiny{$\pm$ 4.72} & 55.57\tiny{$\pm$ 4.06} \\
~ & Vikhr 7B IT 5.4 & 55.50\tiny{$\pm$ 5.73} & 54.32\tiny{$\pm$ 8.73} & 54.79\tiny{$\pm$ 6.93} & 34.30\tiny{$\pm$ 7.65} & 35.19\tiny{$\pm$ 7.64} & 34.58\tiny{$\pm$ 7.06} & 44.69\tiny{$\pm$ 6.59} \\
~ & VikhrGemma 2B IT & 61.41\tiny{$\pm$ 6.24} & 58.64\tiny{$\pm$ 25.16} & 57.67\tiny{$\pm$ 18.04} & 48.66\tiny{$\pm$ 12.71} & 49.07\tiny{$\pm$ 16.79} & 45.68\tiny{$\pm$ 6.56} & 51.68\tiny{$\pm$ 8.61} \\ \hline
    \multirow{7}{*}{AL} & Gemma2 2B IT & 60.18\tiny{$\pm$ 4.51} & 60.61\tiny{$\pm$ 13.55} & 59.66\tiny{$\pm$ 8.17} & 54.87\tiny{$\pm$ 7.30} & 53.51\tiny{$\pm$ 10.71} & \textbf{53.45\tiny{$\pm$ 6.06}} & \textbf{56.56\tiny{$\pm$ 5.40}} \\
~ & Gemma2 9B IT & 47.52\tiny{$\pm$ 5.48} & 46.21\tiny{$\pm$ 4.08} & 46.79\tiny{$\pm$ 4.42} & 39.00\tiny{$\pm$ 6.04} & 40.35\tiny{$\pm$ 8.95} & 39.60\tiny{$\pm$ 7.34} & 43.20\tiny{$\pm$ 5.65} \\
~ & Qwen2 7B IT & 53.25\tiny{$\pm$ 10.65} & 50.00\tiny{$\pm$ 13.38} & 51.40\tiny{$\pm$ 11.90} & 46.74\tiny{$\pm$ 11.99} & 50.00\tiny{$\pm$ 11.67} & 48.16\tiny{$\pm$ 11.50} & 49.78\tiny{$\pm$ 11.33} \\
~ & SaigaLlama3 8B & 56.45\tiny{$\pm$ 3.35} & 53.03\tiny{$\pm$ 6.25} & 54.32\tiny{$\pm$ 3.10} & 48.33\tiny{$\pm$ 4.02} & 51.75\tiny{$\pm$ 11.13} & 49.67\tiny{$\pm$ 7.13} & 51.99\tiny{$\pm$ 3.55} \\
~ & Vikhr 7B IT 0.4 & 54.87\tiny{$\pm$ 5.68} & 50.00\tiny{$\pm$ 6.94} & 52.04\tiny{$\pm$ 4.93} & 47.01\tiny{$\pm$ 4.37} & 51.75\tiny{$\pm$ 10.27} & 49.01\tiny{$\pm$ 6.60} & 50.53\tiny{$\pm$ 4.71} \\
~ & Vikhr 7B IT 5.4 & 57.33\tiny{$\pm$ 3.30} & 56.06\tiny{$\pm$ 8.16} & 56.47\tiny{$\pm$ 4.96} & 50.76\tiny{$\pm$ 4.71} & 51.75\tiny{$\pm$ 6.39} & 50.99\tiny{$\pm$ 3.89} & 53.73\tiny{$\pm$ 3.56} \\
~ & VikhrGemma 2B IT & 59.30\tiny{$\pm$ 2.53} & 56.82\tiny{$\pm$ 8.60} & 57.59\tiny{$\pm$ 4.70} & 52.10\tiny{$\pm$ 3.11} & 54.39\tiny{$\pm$ 9.92} & 52.81\tiny{$\pm$ 5.51} & 55.20\tiny{$\pm$ 2.95} \\ \hline 
    \multirow{7}{*}{AD} & Gemma2 2B IT & 59.21\tiny{$\pm$ 5.93} & 55.83\tiny{$\pm$ 10.17} & 56.62\tiny{$\pm$ 6.27} & 52.58\tiny{$\pm$ 5.69} & 55.56\tiny{$\pm$ 15.38} & 53.36\tiny{$\pm$ 9.71} & 54.99\tiny{$\pm$ 5.86} \\
~ & Gemma2 9B IT & 52.03\tiny{$\pm$ 4.74} & 46.67\tiny{$\pm$ 5.53} & 49.05\tiny{$\pm$ 4.48} & 46.50\tiny{$\pm$ 4.47} & 51.85\tiny{$\pm$ 8.28} & 48.91\tiny{$\pm$ 5.93} & 48.98\tiny{$\pm$ 4.50} \\
~ & Qwen2 7B IT & 48.10\tiny{$\pm$ 2.19} & 49.17\tiny{$\pm$ 12.72} & 47.91\tiny{$\pm$ 7.44} & 41.98\tiny{$\pm$ 4.13} & 41.67\tiny{$\pm$ 13.13} & 41.08\tiny{$\pm$ 7.91} & 44.50\tiny{$\pm$ 2.76} \\
~ & SaigaLlama3 8B & 57.84\tiny{$\pm$ 6.15} & 53.33\tiny{$\pm$ 4.71} & 55.22\tiny{$\pm$ 3.94} & 51.13\tiny{$\pm$ 5.98} & 55.56\tiny{$\pm$ 12.42} & 53.01\tiny{$\pm$ 8.89} & 54.12\tiny{$\pm$ 5.88} \\
~ & Vikhr 7B IT 0.4 & 59.81\tiny{$\pm$ 4.42} & 51.67\tiny{$\pm$ 12.13} & 54.66\tiny{$\pm$ 7.59} & 53.47\tiny{$\pm$ 4.23} & 61.11\tiny{$\pm$ 11.56} & \textbf{56.43\tiny{$\pm$ 6.19}} & \textbf{55.55\tiny{$\pm$ 4.55}} \\
~ & Vikhr 7B IT 5.4 & 51.91\tiny{$\pm$ 8.89} & 55.00\tiny{$\pm$ 12.58} & 53.32\tiny{$\pm$ 10.58} & 47.66\tiny{$\pm$ 10.86} & 44.44\tiny{$\pm$ 8.49} & 45.88\tiny{$\pm$ 9.46} & 49.60\tiny{$\pm$ 9.75} \\
~ & VikhrGemma 2B IT & 45.93\tiny{$\pm$ 4.26} & 39.17\tiny{$\pm$ 6.72} & 42.13\tiny{$\pm$ 5.50} & 42.09\tiny{$\pm$ 3.57} & 49.07\tiny{$\pm$ 5.93} & 45.22\tiny{$\pm$ 4.01} & 43.68\tiny{$\pm$ 3.91} \\ \hline 
    \multirow{7}{*}{AC} & Gemma2 2B IT & 58.73\tiny{$\pm$ 3.32} & 63.70\tiny{$\pm$ 4.19} & 61.08\tiny{$\pm$ 3.43} & 52.17\tiny{$\pm$ 4.36} & 46.93\tiny{$\pm$ 5.35} & 49.35\tiny{$\pm$ 4.71} & 55.22\tiny{$\pm$ 3.84} \\
~ & Gemma2 9B IT & 56.84\tiny{$\pm$ 1.85} & 61.48\tiny{$\pm$ 5.97} & 58.85\tiny{$\pm$ 2.17} & 49.07\tiny{$\pm$ 2.00} & 44.30\tiny{$\pm$ 8.78} & 46.21\tiny{$\pm$ 5.61} & 52.53\tiny{$\pm$ 2.19} \\
~ & Qwen2 7B IT & 57.36\tiny{$\pm$ 3.77} & 57.04\tiny{$\pm$ 5.97} & 57.15\tiny{$\pm$ 4.76} & 49.73\tiny{$\pm$ 4.74} & 50.00\tiny{$\pm$ 4.02} & 49.81\tiny{$\pm$ 4.08} & 53.48\tiny{$\pm$ 4.25} \\
~ & SaigaLlama3 8B & 59.02\tiny{$\pm$ 3.26} & 65.56\tiny{$\pm$ 5.70} & 62.03\tiny{$\pm$ 3.81} & 53.16\tiny{$\pm$ 4.52} & 46.05\tiny{$\pm$ 5.84} & 49.19\tiny{$\pm$ 4.61} & 55.61\tiny{$\pm$ 3.74} \\
~ & Vikhr 7B IT 0.4 & 59.62\tiny{$\pm$ 3.49} & 60.37\tiny{$\pm$ 8.36} & 59.78\tiny{$\pm$ 5.60} & 52.72\tiny{$\pm$ 4.82} & 51.75\tiny{$\pm$ 6.56} & 52.00\tiny{$\pm$ 4.61} & 55.89\tiny{$\pm$ 4.23} \\
~ & Vikhr 7B IT 5.4 & 61.87\tiny{$\pm$ 2.07} & 62.96\tiny{$\pm$ 8.18} & 62.13\tiny{$\pm$ 4.30} & 55.50\tiny{$\pm$ 3.03} & 53.95\tiny{$\pm$ 7.25} & \textbf{54.33\tiny{$\pm$ 3.58}} & \textbf{58.23\tiny{$\pm$ 2.38}} \\
~ & VikhrGemma 2B IT & 57.77\tiny{$\pm$ 2.90} & 64.81\tiny{$\pm$ 6.21} & 60.98\tiny{$\pm$ 3.77} & 51.40\tiny{$\pm$ 4.47} & 43.86\tiny{$\pm$ 6.39} & 47.11\tiny{$\pm$ 4.91} & 54.05\tiny{$\pm$ 3.57} \\
    \hline
    \end{tabular}
    }
    
\caption{The results for LLMs with LoRA on the five main datasets.}
\label{tab:results_llm_peft}
\end{table*}
\begin{table*}[t]
\centering
\resizebox{0.9\linewidth}{!}{%
    \begin{tabular}{lllllllllll}
    \hline
    %Corpus & Model & Mode & \begin{tabular}[c]{@{}l@{}} Precision \\ healthy\end{tabular} & \begin{tabular}[c]{@{}l@{}} Recall \\ healthy\end{tabular} & F1-healthy & \begin{tabular}[c]{@{}l@{}} Precision \\ pathology\end{tabular} & \begin{tabular}[c]{@{}l@{}} Recall \\ pathology\end{tabular} & F1-pathology & F1-macro  \\
    \textbf{Corpus} & \textbf{Model} & \textbf{Mode} & \begin{tabular}[c]{@{}l@{}} \textbf{Precision} \\ \textbf{healthy}\end{tabular} & \begin{tabular}[c]{@{}l@{}} \textbf{Recall} \\ \textbf{healthy}\end{tabular} & \textbf{F1-healthy} & \begin{tabular}[c]{@{}l@{}} \textbf{Precision} \\ \textbf{pathology}\end{tabular} & \begin{tabular}[c]{@{}l@{}} \textbf{Recall} \\ \textbf{pathology}\end{tabular} & \textbf{F1-pathology} & \textbf{F1-macro}  \\
    \hline
    \multirow{14}{*}{DE} & Gemma2 2B IT & 5-shot MMLU & 80.00\tiny{$\pm$ 0.00} & 13.33\tiny{$\pm$ 0.00} & 22.86\tiny{$\pm$ 0.00} & 19.59\tiny{$\pm$ 0.00} & 86.36\tiny{$\pm$ 0.00} & 31.93\tiny{$\pm$ 0.00} & 27.39\tiny{$\pm$ 0.00} \\
~ & Gemma2 2B IT & 0-shot MMLU & 100.00\tiny{$\pm$ 0.00} & 13.33\tiny{$\pm$ 0.00} & 23.53\tiny{$\pm$ 0.00} & 22.00\tiny{$\pm$ 0.00} & 100.00\tiny{$\pm$ 0.00} & 36.07\tiny{$\pm$ 0.00} & 29.80\tiny{$\pm$ 0.00} \\
~ & Gemma2 9B IT & 5-shot MMLU & 93.11\tiny{$\pm$ 0.08} & 75.11\tiny{$\pm$ 0.89} & 83.15\tiny{$\pm$ 0.58} & 43.16\tiny{$\pm$ 0.85} & 77.27\tiny{$\pm$ 0.00} & \textbf{55.38\tiny{$\pm$ 0.71}} & \textbf{69.26\tiny{$\pm$ 0.64}} \\
~ & Gemma2 9B IT & 0-shot MMLU & 92.31\tiny{$\pm$ 0.00} & 53.33\tiny{$\pm$ 0.00} & 67.61\tiny{$\pm$ 0.00} & 30.00\tiny{$\pm$ 0.00} & 81.82\tiny{$\pm$ 0.00} & 43.90\tiny{$\pm$ 0.00} & 55.75\tiny{$\pm$ 0.00} \\
~ & Qwen2 7B IT & 5-shot MMLU & 88.80\tiny{$\pm$ 0.17} & 26.44\tiny{$\pm$ 0.44} & 40.75\tiny{$\pm$ 0.55} & 22.30\tiny{$\pm$ 0.10} & 86.36\tiny{$\pm$ 0.00} & 35.45\tiny{$\pm$ 0.13} & 38.10\tiny{$\pm$ 0.34} \\
~ & Qwen2 7B IT & 0-shot MMLU & 92.42\tiny{$\pm$ 0.16} & 40.67\tiny{$\pm$ 0.89} & 56.48\tiny{$\pm$ 0.89} & 26.25\tiny{$\pm$ 0.29} & 86.36\tiny{$\pm$ 0.00} & 40.26\tiny{$\pm$ 0.34} & 48.37\tiny{$\pm$ 0.62} \\
~ & SaigaLlama3 8B & 5-shot MMLU & 85.00\tiny{$\pm$ 0.00} & 18.89\tiny{$\pm$ 0.00} & 30.91\tiny{$\pm$ 0.00} & 20.65\tiny{$\pm$ 0.00} & 86.36\tiny{$\pm$ 0.00} & 33.33\tiny{$\pm$ 0.00} & 32.12\tiny{$\pm$ 0.00} \\
~ & SaigaLlama3 8B & 0-shot MMLU & 88.37\tiny{$\pm$ 0.26} & 33.78\tiny{$\pm$ 0.89} & 48.87\tiny{$\pm$ 0.96} & 23.20\tiny{$\pm$ 0.24} & 81.82\tiny{$\pm$ 0.00} & 36.15\tiny{$\pm$ 0.29} & 42.51\tiny{$\pm$ 0.63} \\
~ & Vikhr 7B IT 0.4 & 5-shot MMLU & 74.44\tiny{$\pm$ 1.11} & 16.22\tiny{$\pm$ 0.89} & 26.63\tiny{$\pm$ 1.28} & 18.40\tiny{$\pm$ 0.16} & 77.27\tiny{$\pm$ 0.00} & 29.72\tiny{$\pm$ 0.21} & 28.18\tiny{$\pm$ 0.74} \\
~ & Vikhr 7B IT 0.4 & 0-shot MMLU & 0.00\tiny{$\pm$ 0.00} & 0.00\tiny{$\pm$ 0.00} & 0.00\tiny{$\pm$ 0.00} & 19.64\tiny{$\pm$ 0.00} & 100.00\tiny{$\pm$ 0.00} & 32.84\tiny{$\pm$ 0.00} & 16.42\tiny{$\pm$ 0.00} \\
~ & Vikhr 7B IT 5.4 & 5-shot MMLU & 87.06\tiny{$\pm$ 0.00} & 82.22\tiny{$\pm$ 0.00} & 84.57\tiny{$\pm$ 0.00} & 40.74\tiny{$\pm$ 0.00} & 50.00\tiny{$\pm$ 0.00} & 44.90\tiny{$\pm$ 0.00} & 64.73\tiny{$\pm$ 0.00} \\
~ & Vikhr 7B IT 5.4 & 0-shot MMLU & 71.43\tiny{$\pm$ 0.00} & 5.56\tiny{$\pm$ 0.00} & 10.31\tiny{$\pm$ 0.00} & 19.05\tiny{$\pm$ 0.00} & 90.91\tiny{$\pm$ 0.00} & 31.50\tiny{$\pm$ 0.00} & 20.90\tiny{$\pm$ 0.00} \\
~ & VikhrGemma 2B IT & 5-shot MMLU & 85.45\tiny{$\pm$ 0.06} & 91.33\tiny{$\pm$ 0.44} & 88.29\tiny{$\pm$ 0.24} & 50.67\tiny{$\pm$ 1.33} & 36.36\tiny{$\pm$ 0.00} & 42.33\tiny{$\pm$ 0.46} & 65.31\tiny{$\pm$ 0.35} \\
~ & VikhrGemma 2B IT & 0-shot MMLU & 80.14\tiny{$\pm$ 0.07} & 98.67\tiny{$\pm$ 0.44} & 88.45\tiny{$\pm$ 0.22} & 0.00\tiny{$\pm$ 0.00} & 0.00\tiny{$\pm$ 0.00} & 0.00\tiny{$\pm$ 0.00} & 44.22\tiny{$\pm$ 0.11} \\ \hline
    \multirow{14}{*}{DSM} & Gemma2 2B IT & 5-shot MMLU & 71.14\tiny{$\pm$ 0.57} & 20.00\tiny{$\pm$ 2.96} & 31.10\tiny{$\pm$ 3.37} & 42.26\tiny{$\pm$ 0.30} & 87.78\tiny{$\pm$ 2.22} & 57.04\tiny{$\pm$ 0.22} & 44.07\tiny{$\pm$ 1.58} \\
~ & Gemma2 2B IT & 0-shot MMLU & 60.00\tiny{$\pm$ 0.00} & 44.44\tiny{$\pm$ 0.00} & 51.06\tiny{$\pm$ 0.00} & 40.00\tiny{$\pm$ 0.00} & 55.56\tiny{$\pm$ 0.00} & 46.51\tiny{$\pm$ 0.00} & 48.79\tiny{$\pm$ 0.00} \\
~ & Gemma2 9B IT & 5-shot MMLU & 64.91\tiny{$\pm$ 0.40} & 82.22\tiny{$\pm$ 1.48} & 72.54\tiny{$\pm$ 0.82} & 55.64\tiny{$\pm$ 2.18} & 33.33\tiny{$\pm$ 0.00} & 41.67\tiny{$\pm$ 0.59} & 57.11\tiny{$\pm$ 0.71} \\
~ & Gemma2 9B IT & 0-shot MMLU & 64.00\tiny{$\pm$ 0.00} & 59.26\tiny{$\pm$ 0.00} & 61.54\tiny{$\pm$ 0.00} & 45.00\tiny{$\pm$ 0.00} & 50.00\tiny{$\pm$ 0.00} & 47.37\tiny{$\pm$ 0.00} & 54.45\tiny{$\pm$ 0.00} \\
~ & Qwen2 7B IT & 5-shot MMLU & 66.67\tiny{$\pm$ 0.00} & 44.44\tiny{$\pm$ 0.00} & 53.33\tiny{$\pm$ 0.00} & 44.44\tiny{$\pm$ 0.00} & 66.67\tiny{$\pm$ 0.00} & 53.33\tiny{$\pm$ 0.00} & 53.33\tiny{$\pm$ 0.00} \\
~ & Qwen2 7B IT & 0-shot MMLU & 62.96\tiny{$\pm$ 0.00} & 62.96\tiny{$\pm$ 0.00} & 62.96\tiny{$\pm$ 0.00} & 44.44\tiny{$\pm$ 0.00} & 44.44\tiny{$\pm$ 0.00} & 44.44\tiny{$\pm$ 0.00} & 53.70\tiny{$\pm$ 0.00} \\
~ & SaigaLlama3 8B & 5-shot MMLU & 100.00\tiny{$\pm$ 0.00} & 3.70\tiny{$\pm$ 0.00} & 7.14\tiny{$\pm$ 0.00} & 40.91\tiny{$\pm$ 0.00} & 100.00\tiny{$\pm$ 0.00} & \textbf{58.06\tiny{$\pm$ 0.00}} & 32.60\tiny{$\pm$ 0.00} \\
~ & SaigaLlama3 8B & 0-shot MMLU & 63.64\tiny{$\pm$ 0.00} & 51.85\tiny{$\pm$ 0.00} & 57.14\tiny{$\pm$ 0.00} & 43.48\tiny{$\pm$ 0.00} & 55.56\tiny{$\pm$ 0.00} & 48.78\tiny{$\pm$ 0.00} & 52.96\tiny{$\pm$ 0.00} \\
~ & Vikhr 7B IT 0.4 & 5-shot MMLU & 60.00\tiny{$\pm$ 0.00} & 100.00\tiny{$\pm$ 0.00} & 75.00\tiny{$\pm$ 0.00} & 0.00\tiny{$\pm$ 0.00} & 0.00\tiny{$\pm$ 0.00} & 0.00\tiny{$\pm$ 0.00} & 37.50\tiny{$\pm$ 0.00} \\
~ & Vikhr 7B IT 0.4 & 0-shot MMLU & 0.00\tiny{$\pm$ 0.00} & 0.00\tiny{$\pm$ 0.00} & 0.00\tiny{$\pm$ 0.00} & 35.71\tiny{$\pm$ 0.00} & 83.33\tiny{$\pm$ 0.00} & 50.00\tiny{$\pm$ 0.00} & 25.00\tiny{$\pm$ 0.00} \\
~ & Vikhr 7B IT 5.4 & 5-shot MMLU & 68.75\tiny{$\pm$ 0.00} & 81.48\tiny{$\pm$ 0.00} & 74.58\tiny{$\pm$ 0.00} & 61.54\tiny{$\pm$ 0.00} & 44.44\tiny{$\pm$ 0.00} & 51.61\tiny{$\pm$ 0.00} & \textbf{63.09\tiny{$\pm$ 0.00}} \\
~ & Vikhr 7B IT 5.4 & 0-shot MMLU & 60.87\tiny{$\pm$ 0.00} & 51.85\tiny{$\pm$ 0.00} & 56.00\tiny{$\pm$ 0.00} & 40.91\tiny{$\pm$ 0.00} & 50.00\tiny{$\pm$ 0.00} & 45.00\tiny{$\pm$ 0.00} & 50.50\tiny{$\pm$ 0.00} \\
~ & VikhrGemma 2B IT & 5-shot MMLU & 62.34\tiny{$\pm$ 0.88} & 95.56\tiny{$\pm$ 1.48} & 75.44\tiny{$\pm$ 0.16} & 66.67\tiny{$\pm$ 0.00} & 13.33\tiny{$\pm$ 4.44} & 21.90\tiny{$\pm$ 5.71} & 48.67\tiny{$\pm$ 2.94} \\
~ & VikhrGemma 2B IT & 0-shot MMLU & 60.00\tiny{$\pm$ 0.00} & 100.00\tiny{$\pm$ 0.00} & 75.00\tiny{$\pm$ 0.00} & 0.00\tiny{$\pm$ 0.00} & 0.00\tiny{$\pm$ 0.00} & 0.00\tiny{$\pm$ 0.00} & 37.50\tiny{$\pm$ 0.00} \\ \hline
    \multirow{14}{*}{AL} & Gemma2 2B IT & 5-shot MMLU & 50.00\tiny{$\pm$ 0.00} & 4.55\tiny{$\pm$ 0.00} & 8.33\tiny{$\pm$ 0.00} & 46.15\tiny{$\pm$ 0.00} & 94.74\tiny{$\pm$ 0.00} & 62.07\tiny{$\pm$ 0.00} & 35.20\tiny{$\pm$ 0.00} \\
~ & Gemma2 2B IT & 0-shot MMLU & 0.00\tiny{$\pm$ 0.00} & 0.00\tiny{$\pm$ 0.00} & 0.00\tiny{$\pm$ 0.00} & 46.34\tiny{$\pm$ 0.00} & 100.00\tiny{$\pm$ 0.00} & \textbf{63.33\tiny{$\pm$ 0.00}} & 31.67\tiny{$\pm$ 0.00} \\
~ & Gemma2 9B IT & 5-shot MMLU & 66.67\tiny{$\pm$ 0.00} & 45.45\tiny{$\pm$ 0.00} & 54.05\tiny{$\pm$ 0.00} & 53.85\tiny{$\pm$ 0.00} & 73.68\tiny{$\pm$ 0.00} & 62.22\tiny{$\pm$ 0.00} & 58.14\tiny{$\pm$ 0.00} \\
~ & Gemma2 9B IT & 0-shot MMLU & 64.71\tiny{$\pm$ 0.00} & 50.00\tiny{$\pm$ 0.00} & 56.41\tiny{$\pm$ 0.00} & 54.17\tiny{$\pm$ 0.00} & 68.42\tiny{$\pm$ 0.00} & 60.47\tiny{$\pm$ 0.00} & 58.44\tiny{$\pm$ 0.00} \\
~ & Qwen2 7B IT & 5-shot MMLU & 60.00\tiny{$\pm$ 0.00} & 27.27\tiny{$\pm$ 0.00} & 37.50\tiny{$\pm$ 0.00} & 48.39\tiny{$\pm$ 0.00} & 78.95\tiny{$\pm$ 0.00} & 60.00\tiny{$\pm$ 0.00} & 48.75\tiny{$\pm$ 0.00} \\
~ & Qwen2 7B IT & 0-shot MMLU & 61.03\tiny{$\pm$ 0.52} & 85.45\tiny{$\pm$ 1.82} & 71.20\tiny{$\pm$ 0.99} & 68.73\tiny{$\pm$ 2.55} & 36.84\tiny{$\pm$ 0.00} & 47.95\tiny{$\pm$ 0.64} & \textbf{59.58\tiny{$\pm$ 0.82}} \\
~ & SaigaLlama3 8B & 5-shot MMLU & 0.00\tiny{$\pm$ 0.00} & 0.00\tiny{$\pm$ 0.00} & 0.00\tiny{$\pm$ 0.00} & 46.34\tiny{$\pm$ 0.00} & 100.00\tiny{$\pm$ 0.00} & \textbf{63.33\tiny{$\pm$ 0.00}} & 31.67\tiny{$\pm$ 0.00} \\
~ & SaigaLlama3 8B & 0-shot MMLU & 63.64\tiny{$\pm$ 0.00} & 31.82\tiny{$\pm$ 0.00} & 42.42\tiny{$\pm$ 0.00} & 50.00\tiny{$\pm$ 0.00} & 78.95\tiny{$\pm$ 0.00} & 61.22\tiny{$\pm$ 0.00} & 51.82\tiny{$\pm$ 0.00} \\
~ & Vikhr 7B IT 0.4 & 5-shot MMLU & 55.88\tiny{$\pm$ 0.00} & 86.36\tiny{$\pm$ 0.00} & 67.86\tiny{$\pm$ 0.00} & 57.14\tiny{$\pm$ 0.00} & 21.05\tiny{$\pm$ 0.00} & 30.77\tiny{$\pm$ 0.00} & 49.31\tiny{$\pm$ 0.00} \\
~ & Vikhr 7B IT 0.4 & 0-shot MMLU & 0.00\tiny{$\pm$ 0.00} & 0.00\tiny{$\pm$ 0.00} & 0.00\tiny{$\pm$ 0.00} & 45.00\tiny{$\pm$ 0.00} & 94.74\tiny{$\pm$ 0.00} & 61.02\tiny{$\pm$ 0.00} & 30.51\tiny{$\pm$ 0.00} \\
~ & Vikhr 7B IT 5.4 & 5-shot MMLU & 60.00\tiny{$\pm$ 0.00} & 81.82\tiny{$\pm$ 0.00} & 69.23\tiny{$\pm$ 0.00} & 63.64\tiny{$\pm$ 0.00} & 36.84\tiny{$\pm$ 0.00} & 46.67\tiny{$\pm$ 0.00} & 57.95\tiny{$\pm$ 0.00} \\
~ & Vikhr 7B IT 5.4 & 0-shot MMLU & 33.33\tiny{$\pm$ 0.00} & 4.55\tiny{$\pm$ 0.00} & 8.00\tiny{$\pm$ 0.00} & 44.74\tiny{$\pm$ 0.00} & 89.47\tiny{$\pm$ 0.00} & 59.65\tiny{$\pm$ 0.00} & 33.82\tiny{$\pm$ 0.00} \\
~ & VikhrGemma 2B IT & 5-shot MMLU & 51.92\tiny{$\pm$ 0.60} & 73.64\tiny{$\pm$ 1.82} & 60.89\tiny{$\pm$ 1.03} & 40.89\tiny{$\pm$ 1.78} & 21.05\tiny{$\pm$ 0.00} & 27.78\tiny{$\pm$ 0.39} & 44.34\tiny{$\pm$ 0.71} \\
~ & VikhrGemma 2B IT & 0-shot MMLU & 53.66\tiny{$\pm$ 0.00} & 100.00\tiny{$\pm$ 0.00} & 69.84\tiny{$\pm$ 0.00} & 0.00\tiny{$\pm$ 0.00} & 0.00\tiny{$\pm$ 0.00} & 0.00\tiny{$\pm$ 0.00} & 34.92\tiny{$\pm$ 0.00} \\ \hline 
    \multirow{14}{*}{AD} & Gemma2 2B IT & 5-shot MMLU & 0.00\tiny{$\pm$ 0.00} & 0.00\tiny{$\pm$ 0.00} & 0.00\tiny{$\pm$ 0.00} & 47.37\tiny{$\pm$ 0.00} & 100.00\tiny{$\pm$ 0.00} & 64.29\tiny{$\pm$ 0.00} & 32.14\tiny{$\pm$ 0.00} \\
~ & Gemma2 2B IT & 0-shot MMLU & 50.00\tiny{$\pm$ 0.00} & 10.00\tiny{$\pm$ 0.00} & 16.67\tiny{$\pm$ 0.00} & 47.06\tiny{$\pm$ 0.00} & 88.89\tiny{$\pm$ 0.00} & 61.54\tiny{$\pm$ 0.00} & 39.10\tiny{$\pm$ 0.00} \\
~ & Gemma2 9B IT & 5-shot MMLU & 55.71\tiny{$\pm$ 2.86} & 19.00\tiny{$\pm$ 2.00} & 28.32\tiny{$\pm$ 2.62} & 48.08\tiny{$\pm$ 0.60} & 83.33\tiny{$\pm$ 0.00} & 60.98\tiny{$\pm$ 0.49} & 44.65\tiny{$\pm$ 1.56} \\
~ & Gemma2 9B IT & 0-shot MMLU & 77.78\tiny{$\pm$ 0.00} & 35.00\tiny{$\pm$ 0.00} & 48.28\tiny{$\pm$ 0.00} & 55.17\tiny{$\pm$ 0.00} & 88.89\tiny{$\pm$ 0.00} & 68.09\tiny{$\pm$ 0.00} & 58.18\tiny{$\pm$ 0.00} \\
~ & Qwen2 7B IT & 5-shot MMLU & 100.00\tiny{$\pm$ 0.00} & 5.00\tiny{$\pm$ 0.00} & 9.52\tiny{$\pm$ 0.00} & 48.65\tiny{$\pm$ 0.00} & 100.00\tiny{$\pm$ 0.00} & 65.45\tiny{$\pm$ 0.00} & 37.49\tiny{$\pm$ 0.00} \\
~ & Qwen2 7B IT & 0-shot MMLU & 70.74\tiny{$\pm$ 1.47} & 70.00\tiny{$\pm$ 0.00} & 70.36\tiny{$\pm$ 0.72} & 67.02\tiny{$\pm$ 0.70} & 67.78\tiny{$\pm$ 2.22} & 67.39\tiny{$\pm$ 1.44} & \textbf{68.87\tiny{$\pm$ 1.08}} \\
~ & SaigaLlama3 8B & 5-shot MMLU & 0.00\tiny{$\pm$ 0.00} & 0.00\tiny{$\pm$ 0.00} & 0.00\tiny{$\pm$ 0.00} & 47.37\tiny{$\pm$ 0.00} & 100.00\tiny{$\pm$ 0.00} & 64.29\tiny{$\pm$ 0.00} & 32.14\tiny{$\pm$ 0.00} \\
~ & SaigaLlama3 8B & 0-shot MMLU & 80.00\tiny{$\pm$ 0.00} & 20.00\tiny{$\pm$ 0.00} & 32.00\tiny{$\pm$ 0.00} & 51.52\tiny{$\pm$ 0.00} & 94.44\tiny{$\pm$ 0.00} & 66.67\tiny{$\pm$ 0.00} & 49.33\tiny{$\pm$ 0.00} \\
~ & Vikhr 7B IT 0.4 & 5-shot MMLU & 55.62\tiny{$\pm$ 1.25} & 89.00\tiny{$\pm$ 2.00} & 68.46\tiny{$\pm$ 1.54} & 63.33\tiny{$\pm$ 6.67} & 21.11\tiny{$\pm$ 2.22} & 31.67\tiny{$\pm$ 3.33} & 50.06\tiny{$\pm$ 2.44} \\
~ & Vikhr 7B IT 0.4 & 0-shot MMLU & 13.33\tiny{$\pm$ 26.67} & 2.00\tiny{$\pm$ 4.00} & 3.48\tiny{$\pm$ 6.96} & 46.47\tiny{$\pm$ 1.05} & 94.44\tiny{$\pm$ 0.00} & 62.28\tiny{$\pm$ 0.93} & 32.88\tiny{$\pm$ 3.94} \\
~ & Vikhr 7B IT 5.4 & 5-shot MMLU & 66.67\tiny{$\pm$ 0.00} & 10.00\tiny{$\pm$ 0.00} & 17.39\tiny{$\pm$ 0.00} & 48.57\tiny{$\pm$ 0.00} & 94.44\tiny{$\pm$ 0.00} & 64.15\tiny{$\pm$ 0.00} & 40.77\tiny{$\pm$ 0.00} \\
~ & Vikhr 7B IT 5.4 & 0-shot MMLU & 83.33\tiny{$\pm$ 0.00} & 25.00\tiny{$\pm$ 0.00} & 38.46\tiny{$\pm$ 0.00} & 53.12\tiny{$\pm$ 0.00} & 94.44\tiny{$\pm$ 0.00} & \textbf{68.00\tiny{$\pm$ 0.00}} & 53.23\tiny{$\pm$ 0.00} \\
~ & VikhrGemma 2B IT & 5-shot MMLU & 38.67\tiny{$\pm$ 2.67} & 10.00\tiny{$\pm$ 0.00} & 15.88\tiny{$\pm$ 0.25} & 45.11\tiny{$\pm$ 0.68} & 82.22\tiny{$\pm$ 2.22} & 58.26\tiny{$\pm$ 1.13} & 37.07\tiny{$\pm$ 0.69} \\
~ & VikhrGemma 2B IT & 0-shot MMLU & 52.63\tiny{$\pm$ 0.00} & 100.00\tiny{$\pm$ 0.00} & 68.97\tiny{$\pm$ 0.00} & 0.00\tiny{$\pm$ 0.00} & 0.00\tiny{$\pm$ 0.00} & 0.00\tiny{$\pm$ 0.00} & 34.48\tiny{$\pm$ 0.00} \\ \hline 
    \multirow{14}{*}{AC} & Gemma2 2B IT & 5-shot MMLU & 74.55\tiny{$\pm$ 0.91} & 19.56\tiny{$\pm$ 0.89} & 30.98\tiny{$\pm$ 1.20} & 49.16\tiny{$\pm$ 0.27} & 92.11\tiny{$\pm$ 0.00} & 64.10\tiny{$\pm$ 0.23} & 47.54\tiny{$\pm$ 0.72} \\
~ & Gemma2 2B IT & 0-shot MMLU & 72.73\tiny{$\pm$ 0.00} & 17.78\tiny{$\pm$ 0.00} & 28.57\tiny{$\pm$ 0.00} & 48.61\tiny{$\pm$ 0.00} & 92.11\tiny{$\pm$ 0.00} & 63.64\tiny{$\pm$ 0.00} & 46.10\tiny{$\pm$ 0.00} \\
~ & Gemma2 9B IT & 5-shot MMLU & 60.46\tiny{$\pm$ 1.66} & 51.56\tiny{$\pm$ 18.67} & 53.91\tiny{$\pm$ 7.83} & 52.31\tiny{$\pm$ 4.62} & 58.95\tiny{$\pm$ 18.95} & 52.50\tiny{$\pm$ 10.56} & 53.20\tiny{$\pm$ 1.37} \\
~ & Gemma2 9B IT & 0-shot MMLU & 58.46\tiny{$\pm$ 1.54} & 33.78\tiny{$\pm$ 0.89} & 42.82\tiny{$\pm$ 1.13} & 47.72\tiny{$\pm$ 0.70} & 71.58\tiny{$\pm$ 1.05} & 57.26\tiny{$\pm$ 0.84} & 50.04\tiny{$\pm$ 0.98} \\
~ & Qwen2 7B IT & 5-shot MMLU & 71.52\tiny{$\pm$ 2.42} & 34.67\tiny{$\pm$ 1.78} & 46.69\tiny{$\pm$ 2.13} & 51.97\tiny{$\pm$ 0.98} & 83.68\tiny{$\pm$ 1.05} & 64.12\tiny{$\pm$ 1.06} & 55.41\tiny{$\pm$ 1.60} \\
~ & Qwen2 7B IT & 0-shot MMLU & 59.89\tiny{$\pm$ 0.22} & 52.44\tiny{$\pm$ 1.78} & 55.91\tiny{$\pm$ 1.12} & 50.93\tiny{$\pm$ 0.47} & 58.42\tiny{$\pm$ 1.05} & 54.41\tiny{$\pm$ 0.18} & 55.16\tiny{$\pm$ 0.47} \\
~ & SaigaLlama3 8B & 5-shot MMLU & 100.00\tiny{$\pm$ 0.00} & 13.33\tiny{$\pm$ 0.00} & 23.53\tiny{$\pm$ 0.00} & 49.35\tiny{$\pm$ 0.00} & 100.00\tiny{$\pm$ 0.00} & \textbf{66.09\tiny{$\pm$ 0.00}} & 44.81\tiny{$\pm$ 0.00} \\
~ & SaigaLlama3 8B & 0-shot MMLU & 67.41\tiny{$\pm$ 1.48} & 40.44\tiny{$\pm$ 0.89} & 50.56\tiny{$\pm$ 1.11} & 52.14\tiny{$\pm$ 0.71} & 76.84\tiny{$\pm$ 1.05} & 62.13\tiny{$\pm$ 0.85} & 56.34\tiny{$\pm$ 0.98} \\
~ & Vikhr 7B IT 0.4 & 5-shot MMLU & 54.22\tiny{$\pm$ 0.00} & 100.00\tiny{$\pm$ 0.00} & 70.31\tiny{$\pm$ 0.00} & 0.00\tiny{$\pm$ 0.00} & 0.00\tiny{$\pm$ 0.00} & 0.00\tiny{$\pm$ 0.00} & 35.16\tiny{$\pm$ 0.00} \\
~ & Vikhr 7B IT 0.4 & 0-shot MMLU & 69.15\tiny{$\pm$ 2.43} & 41.78\tiny{$\pm$ 0.89} & 52.09\tiny{$\pm$ 1.39} & 53.04\tiny{$\pm$ 1.06} & 77.89\tiny{$\pm$ 2.11} & 63.11\tiny{$\pm$ 1.45} & \textbf{57.60\tiny{$\pm$ 1.42}} \\
~ & Vikhr 7B IT 5.4 & 5-shot MMLU & 90.67\tiny{$\pm$ 18.67} & 8.89\tiny{$\pm$ 4.44} & 15.33\tiny{$\pm$ 5.67} & 47.12\tiny{$\pm$ 0.76} & 96.32\tiny{$\pm$ 7.37} & 63.22\tiny{$\pm$ 2.37} & 39.28\tiny{$\pm$ 1.65} \\
~ & Vikhr 7B IT 5.4 & 0-shot MMLU & 57.50\tiny{$\pm$ 0.00} & 51.11\tiny{$\pm$ 0.00} & 54.12\tiny{$\pm$ 0.00} & 48.84\tiny{$\pm$ 0.00} & 55.26\tiny{$\pm$ 0.00} & 51.85\tiny{$\pm$ 0.00} & 52.98\tiny{$\pm$ 0.00} \\
~ & VikhrGemma 2B IT & 5-shot MMLU & 55.29\tiny{$\pm$ 0.82} & 97.78\tiny{$\pm$ 0.00} & 70.63\tiny{$\pm$ 0.67} & 60.00\tiny{$\pm$ 30.00} & 6.32\tiny{$\pm$ 3.16} & 11.43\tiny{$\pm$ 5.71} & 41.03\tiny{$\pm$ 3.19} \\
~ & VikhrGemma 2B IT & 0-shot MMLU & 53.66\tiny{$\pm$ 0.00} & 97.78\tiny{$\pm$ 0.00} & 69.29\tiny{$\pm$ 0.00} & 0.00\tiny{$\pm$ 0.00} & 0.00\tiny{$\pm$ 0.00} & 0.00\tiny{$\pm$ 0.00} & 34.65\tiny{$\pm$ 0.00} \\
    \hline
    \end{tabular}
    }
    
\caption{The results for LLMs MMLU-style evaluation on the five main datasets.}
\label{tab:results_llm_mmlu}
\end{table*}
\begin{table*}[t]
\centering
\resizebox{0.9\linewidth}{!}{%
    \begin{tabular}{llllllllllll}
    \hline
    %Corpus & Model & Mode & \begin{tabular}[c]{@{}l@{}} Precision \\ healthy\end{tabular} & \begin{tabular}[c]{@{}l@{}} Recall \\ healthy\end{tabular} & F1-healthy & \begin{tabular}[c]{@{}l@{}} Precision \\ pathology\end{tabular} & \begin{tabular}[c]{@{}l@{}} Recall \\ pathology\end{tabular} & F1-pathology & F1-macro & \begin{tabular}[c]{@{}l@{}} Matched \\ percentage\end{tabular} \\
    \textbf{Corpus} & \textbf{Model} & \textbf{Mode} & \begin{tabular}[c]{@{}l@{}} \textbf{Precision} \\ \textbf{healthy}\end{tabular} & \begin{tabular}[c]{@{}l@{}} \textbf{Recall} \\ \textbf{healthy}\end{tabular} & \textbf{F1-healthy} & \begin{tabular}[c]{@{}l@{}} \textbf{Precision} \\ \textbf{pathology}\end{tabular} & \begin{tabular}[c]{@{}l@{}} \textbf{Recall} \\ \textbf{pathology}\end{tabular} & \textbf{F1-pathology} & \textbf{F1-macro} & \begin{tabular}[c]{@{}l@{}} \textbf{Matched} \\ \textbf{percentage}\end{tabular} \\
    \hline
    \multirow{14}{*}{DE} & Gemma2 2B IT & 5-shot & 82.38\tiny{$\pm$ 1.12} & 10.44\tiny{$\pm$ 0.89} & 18.53\tiny{$\pm$ 1.41} & 20.08\tiny{$\pm$ 0.16} & 90.91\tiny{$\pm$ 0.00} & 32.90\tiny{$\pm$ 0.22} & 17.14\tiny{$\pm$ 0.54} & 99.11\tiny{$\pm$ 0.00} \\
~ & Gemma2 2B IT & 0-shot & 95.83\tiny{$\pm$ 0.00} & 25.56\tiny{$\pm$ 0.00} & 40.35\tiny{$\pm$ 0.00} & 23.86\tiny{$\pm$ 0.00} & 95.45\tiny{$\pm$ 0.00} & 38.18\tiny{$\pm$ 0.00} & 39.27\tiny{$\pm$ 0.00} & 100.00\tiny{$\pm$ 0.00} \\
~ & Gemma2 9B IT & 5-shot & 92.80\tiny{$\pm$ 0.11} & 57.33\tiny{$\pm$ 0.89} & 70.88\tiny{$\pm$ 0.72} & 31.92\tiny{$\pm$ 0.44} & 81.82\tiny{$\pm$ 0.00} & 45.92\tiny{$\pm$ 0.46} & 58.40\tiny{$\pm$ 0.59} & 100.00\tiny{$\pm$ 0.00} \\
~ & Gemma2 9B IT & 0-shot & 93.33\tiny{$\pm$ 0.00} & 62.22\tiny{$\pm$ 0.00} & 74.67\tiny{$\pm$ 0.00} & 34.62\tiny{$\pm$ 0.00} & 81.82\tiny{$\pm$ 0.00} & \textbf{48.65\tiny{$\pm$ 0.00}} & 61.66\tiny{$\pm$ 0.00} & 100.00\tiny{$\pm$ 0.00} \\
~ & Qwen2 7B IT & 5-shot & 88.80\tiny{$\pm$ 0.17} & 26.44\tiny{$\pm$ 0.44} & 40.75\tiny{$\pm$ 0.55} & 22.30\tiny{$\pm$ 0.10} & 86.36\tiny{$\pm$ 0.00} & 35.45\tiny{$\pm$ 0.13} & 38.10\tiny{$\pm$ 0.34} & 100.00\tiny{$\pm$ 0.00} \\
~ & Qwen2 7B IT & 0-shot & 92.33\tiny{$\pm$ 1.05} & 42.89\tiny{$\pm$ 0.89} & 58.57\tiny{$\pm$ 1.04} & 26.78\tiny{$\pm$ 0.72} & 85.45\tiny{$\pm$ 1.82} & 40.79\tiny{$\pm$ 1.04} & 49.68\tiny{$\pm$ 1.04} & 100.00\tiny{$\pm$ 0.00} \\
~ & SaigaLlama3 8B & 5-shot & 85.00\tiny{$\pm$ 0.00} & 18.89\tiny{$\pm$ 0.00} & 30.91\tiny{$\pm$ 0.00} & 20.65\tiny{$\pm$ 0.00} & 86.36\tiny{$\pm$ 0.00} & 33.33\tiny{$\pm$ 0.00} & 32.12\tiny{$\pm$ 0.00} & 100.00\tiny{$\pm$ 0.00} \\
~ & SaigaLlama3 8B & 0-shot & 100.00\tiny{$\pm$ 0.00} & 8.00\tiny{$\pm$ 0.44} & 14.81\tiny{$\pm$ 0.76} & 20.99\tiny{$\pm$ 0.08} & 100.00\tiny{$\pm$ 0.00} & 34.70\tiny{$\pm$ 0.11} & 24.76\tiny{$\pm$ 0.43} & 100.00\tiny{$\pm$ 0.00} \\
~ & Vikhr 7B IT 0.4 & 5-shot & 73.07\tiny{$\pm$ 1.24} & 15.11\tiny{$\pm$ 0.89} & 25.04\tiny{$\pm$ 1.30} & 18.20\tiny{$\pm$ 0.15} & 77.27\tiny{$\pm$ 0.00} & 29.46\tiny{$\pm$ 0.20} & 27.25\tiny{$\pm$ 0.75} & 100.00\tiny{$\pm$ 0.00} \\
~ & Vikhr 7B IT 0.4 & 0-shot & 0.00\tiny{$\pm$ 0.00} & 0.00\tiny{$\pm$ 0.00} & 0.00\tiny{$\pm$ 0.00} & 19.64\tiny{$\pm$ 0.00} & 100.00\tiny{$\pm$ 0.00} & 32.84\tiny{$\pm$ 0.00} & 16.42\tiny{$\pm$ 0.00} & 100.00\tiny{$\pm$ 0.00} \\
~ & Vikhr 7B IT 5.4 & 5-shot & 87.06\tiny{$\pm$ 0.00} & 82.22\tiny{$\pm$ 0.00} & 84.57\tiny{$\pm$ 0.00} & 40.74\tiny{$\pm$ 0.00} & 50.00\tiny{$\pm$ 0.00} & 44.90\tiny{$\pm$ 0.00} & \textbf{64.73\tiny{$\pm$ 0.00}} & 100.00\tiny{$\pm$ 0.00} \\
~ & Vikhr 7B IT 5.4 & 0-shot & 71.43\tiny{$\pm$ 0.00} & 5.56\tiny{$\pm$ 0.00} & 10.31\tiny{$\pm$ 0.00} & 19.05\tiny{$\pm$ 0.00} & 90.91\tiny{$\pm$ 0.00} & 31.50\tiny{$\pm$ 0.00} & 20.90\tiny{$\pm$ 0.00} & 100.00\tiny{$\pm$ 0.00} \\
~ & VikhrGemma 2B IT & 5-shot & 88.54\tiny{$\pm$ 0.07} & 68.67\tiny{$\pm$ 0.44} & 77.35\tiny{$\pm$ 0.31} & 33.18\tiny{$\pm$ 0.31} & 63.64\tiny{$\pm$ 0.00} & 43.62\tiny{$\pm$ 0.27} & 60.48\tiny{$\pm$ 0.29} & 100.00\tiny{$\pm$ 0.00} \\
~ & VikhrGemma 2B IT & 0-shot & 85.24\tiny{$\pm$ 0.95} & 6.44\tiny{$\pm$ 0.44} & 11.98\tiny{$\pm$ 0.78} & 20.59\tiny{$\pm$ 0.40} & 94.55\tiny{$\pm$ 1.82} & 33.82\tiny{$\pm$ 0.65} & 15.27\tiny{$\pm$ 0.48} & 96.25\tiny{$\pm$ 0.36} \\ \hline
    \multirow{14}{*}{DSM} & Gemma2 2B IT & 5-shot & 54.00\tiny{$\pm$ 8.00} & 11.11\tiny{$\pm$ 7.41} & 17.89\tiny{$\pm$ 9.97} & 39.45\tiny{$\pm$ 0.86} & 86.67\tiny{$\pm$ 4.44} & 54.16\tiny{$\pm$ 0.16} & 32.97\tiny{$\pm$ 1.20} & 99.56\tiny{$\pm$ 0.89} \\
~ & Gemma2 2B IT & 0-shot & 64.71\tiny{$\pm$ 0.00} & 40.74\tiny{$\pm$ 0.00} & 50.00\tiny{$\pm$ 0.00} & 42.86\tiny{$\pm$ 0.00} & 66.67\tiny{$\pm$ 0.00} & 52.17\tiny{$\pm$ 0.00} & 51.09\tiny{$\pm$ 0.00} & 100.00\tiny{$\pm$ 0.00} \\
~ & Gemma2 9B IT & 5-shot & 64.24\tiny{$\pm$ 1.21} & 29.63\tiny{$\pm$ 7.41} & 40.14\tiny{$\pm$ 6.60} & 41.83\tiny{$\pm$ 1.31} & 75.56\tiny{$\pm$ 4.44} & 53.74\tiny{$\pm$ 0.21} & 46.94\tiny{$\pm$ 3.20} & 100.00\tiny{$\pm$ 0.00} \\
~ & Gemma2 9B IT & 0-shot & 62.64\tiny{$\pm$ 0.44} & 80.74\tiny{$\pm$ 1.48} & 70.54\tiny{$\pm$ 0.85} & 49.09\tiny{$\pm$ 1.82} & 27.78\tiny{$\pm$ 0.00} & 35.47\tiny{$\pm$ 0.49} & 53.01\tiny{$\pm$ 0.67} & 100.00\tiny{$\pm$ 0.00} \\
~ & Qwen2 7B IT & 5-shot & 70.59\tiny{$\pm$ 0.00} & 44.44\tiny{$\pm$ 0.00} & 54.55\tiny{$\pm$ 0.00} & 46.43\tiny{$\pm$ 0.00} & 72.22\tiny{$\pm$ 0.00} & 56.52\tiny{$\pm$ 0.00} & 55.53\tiny{$\pm$ 0.00} & 100.00\tiny{$\pm$ 0.00} \\
~ & Qwen2 7B IT & 0-shot & 65.52\tiny{$\pm$ 0.00} & 70.37\tiny{$\pm$ 0.00} & 67.86\tiny{$\pm$ 0.00} & 50.00\tiny{$\pm$ 0.00} & 44.44\tiny{$\pm$ 0.00} & 47.06\tiny{$\pm$ 0.00} & 57.46\tiny{$\pm$ 0.00} & 100.00\tiny{$\pm$ 0.00} \\
~ & SaigaLlama3 8B & 5-shot & 100.00\tiny{$\pm$ 0.00} & 3.70\tiny{$\pm$ 0.00} & 7.14\tiny{$\pm$ 0.00} & 40.91\tiny{$\pm$ 0.00} & 100.00\tiny{$\pm$ 0.00} & \textbf{58.06\tiny{$\pm$ 0.00}} & 32.60\tiny{$\pm$ 0.00} & 100.00\tiny{$\pm$ 0.00} \\
~ & SaigaLlama3 8B & 0-shot & 55.30\tiny{$\pm$ 1.52} & 22.96\tiny{$\pm$ 1.48} & 32.44\tiny{$\pm$ 1.73} & 38.47\tiny{$\pm$ 0.46} & 72.22\tiny{$\pm$ 0.00} & 50.20\tiny{$\pm$ 0.39} & 41.32\tiny{$\pm$ 1.06} & 100.00\tiny{$\pm$ 0.00} \\
~ & Vikhr 7B IT 0.4 & 5-shot & 69.19\tiny{$\pm$ 0.89} & 81.48\tiny{$\pm$ 0.00} & 74.83\tiny{$\pm$ 0.51} & 62.09\tiny{$\pm$ 1.10} & 45.56\tiny{$\pm$ 2.22} & 52.54\tiny{$\pm$ 1.85} & \textbf{63.69\tiny{$\pm$ 1.18}} & 100.00\tiny{$\pm$ 0.00} \\
~ & Vikhr 7B IT 0.4 & 0-shot & 0.00\tiny{$\pm$ 0.00} & 0.00\tiny{$\pm$ 0.00} & 0.00\tiny{$\pm$ 0.00} & 35.71\tiny{$\pm$ 0.00} & 83.33\tiny{$\pm$ 0.00} & 50.00\tiny{$\pm$ 0.00} & 25.00\tiny{$\pm$ 0.00} & 100.00\tiny{$\pm$ 0.00} \\
~ & Vikhr 7B IT 5.4 & 5-shot & 68.75\tiny{$\pm$ 0.00} & 81.48\tiny{$\pm$ 0.00} & 74.58\tiny{$\pm$ 0.00} & 61.54\tiny{$\pm$ 0.00} & 44.44\tiny{$\pm$ 0.00} & 51.61\tiny{$\pm$ 0.00} & 63.09\tiny{$\pm$ 0.00} & 100.00\tiny{$\pm$ 0.00} \\
~ & Vikhr 7B IT 5.4 & 0-shot & 59.09\tiny{$\pm$ 0.00} & 48.15\tiny{$\pm$ 0.00} & 53.06\tiny{$\pm$ 0.00} & 39.13\tiny{$\pm$ 0.00} & 50.00\tiny{$\pm$ 0.00} & 43.90\tiny{$\pm$ 0.00} & 48.48\tiny{$\pm$ 0.00} & 100.00\tiny{$\pm$ 0.00} \\
~ & VikhrGemma 2B IT & 5-shot & 61.34\tiny{$\pm$ 5.04} & 37.04\tiny{$\pm$ 0.00} & 46.12\tiny{$\pm$ 1.33} & 40.46\tiny{$\pm$ 2.35} & 64.44\tiny{$\pm$ 6.67} & 49.69\tiny{$\pm$ 3.73} & 47.90\tiny{$\pm$ 2.53} & 100.00\tiny{$\pm$ 0.00} \\
~ & VikhrGemma 2B IT & 0-shot & 57.14\tiny{$\pm$ 0.00} & 14.81\tiny{$\pm$ 0.00} & 23.53\tiny{$\pm$ 0.00} & 41.94\tiny{$\pm$ 0.00} & 72.22\tiny{$\pm$ 0.00} & 53.06\tiny{$\pm$ 0.00} & 25.53\tiny{$\pm$ 0.00} & 84.44\tiny{$\pm$ 0.00} \\ \hline
    \multirow{14}{*}{AL} & Gemma2 2B IT & 5-shot & 57.14\tiny{$\pm$ 0.00} & 18.18\tiny{$\pm$ 0.00} & 27.59\tiny{$\pm$ 0.00} & 47.06\tiny{$\pm$ 0.00} & 84.21\tiny{$\pm$ 0.00} & 60.38\tiny{$\pm$ 0.00} & 43.98\tiny{$\pm$ 0.00} & 100.00\tiny{$\pm$ 0.00} \\
~ & Gemma2 2B IT & 0-shot & 63.64\tiny{$\pm$ 0.00} & 31.82\tiny{$\pm$ 0.00} & 42.42\tiny{$\pm$ 0.00} & 50.00\tiny{$\pm$ 0.00} & 78.95\tiny{$\pm$ 0.00} & 61.22\tiny{$\pm$ 0.00} & 51.82\tiny{$\pm$ 0.00} & 100.00\tiny{$\pm$ 0.00} \\
~ & Gemma2 9B IT & 5-shot & 75.00\tiny{$\pm$ 0.00} & 27.27\tiny{$\pm$ 0.00} & 40.00\tiny{$\pm$ 0.00} & 51.52\tiny{$\pm$ 0.00} & 89.47\tiny{$\pm$ 0.00} & \textbf{65.38\tiny{$\pm$ 0.00}} & 52.69\tiny{$\pm$ 0.00} & 100.00\tiny{$\pm$ 0.00} \\
~ & Gemma2 9B IT & 0-shot & 66.67\tiny{$\pm$ 0.00} & 63.64\tiny{$\pm$ 0.00} & 65.12\tiny{$\pm$ 0.00} & 60.00\tiny{$\pm$ 0.00} & 63.16\tiny{$\pm$ 0.00} & 61.54\tiny{$\pm$ 0.00} & \textbf{63.33\tiny{$\pm$ 0.00}} & 100.00\tiny{$\pm$ 0.00} \\
~ & Qwen2 7B IT & 5-shot & 61.33\tiny{$\pm$ 2.67} & 27.27\tiny{$\pm$ 0.00} & 37.74\tiny{$\pm$ 0.48} & 48.71\tiny{$\pm$ 0.65} & 80.00\tiny{$\pm$ 2.11} & 60.55\tiny{$\pm$ 1.10} & 49.15\tiny{$\pm$ 0.79} & 100.00\tiny{$\pm$ 0.00} \\
~ & Qwen2 7B IT & 0-shot & 60.00\tiny{$\pm$ 0.00} & 95.45\tiny{$\pm$ 0.00} & 73.68\tiny{$\pm$ 0.00} & 83.33\tiny{$\pm$ 0.00} & 26.32\tiny{$\pm$ 0.00} & 40.00\tiny{$\pm$ 0.00} & 56.84\tiny{$\pm$ 0.00} & 100.00\tiny{$\pm$ 0.00} \\
~ & SaigaLlama3 8B & 5-shot & 100.00\tiny{$\pm$ 0.00} & 4.55\tiny{$\pm$ 0.00} & 8.70\tiny{$\pm$ 0.00} & 47.50\tiny{$\pm$ 0.00} & 100.00\tiny{$\pm$ 0.00} & 64.41\tiny{$\pm$ 0.00} & 36.55\tiny{$\pm$ 0.00} & 100.00\tiny{$\pm$ 0.00} \\
~ & SaigaLlama3 8B & 0-shot & 36.67\tiny{$\pm$ 6.67} & 5.45\tiny{$\pm$ 1.82} & 9.48\tiny{$\pm$ 2.95} & 44.98\tiny{$\pm$ 0.48} & 89.47\tiny{$\pm$ 0.00} & 59.86\tiny{$\pm$ 0.43} & 34.67\tiny{$\pm$ 1.69} & 100.00\tiny{$\pm$ 0.00} \\
~ & Vikhr 7B IT 0.4 & 5-shot & 50.00\tiny{$\pm$ 0.00} & 45.45\tiny{$\pm$ 0.00} & 47.62\tiny{$\pm$ 0.00} & 42.86\tiny{$\pm$ 0.00} & 47.37\tiny{$\pm$ 0.00} & 45.00\tiny{$\pm$ 0.00} & 46.31\tiny{$\pm$ 0.00} & 100.00\tiny{$\pm$ 0.00} \\
~ & Vikhr 7B IT 0.4 & 0-shot & 0.00\tiny{$\pm$ 0.00} & 0.00\tiny{$\pm$ 0.00} & 0.00\tiny{$\pm$ 0.00} & 45.00\tiny{$\pm$ 0.00} & 94.74\tiny{$\pm$ 0.00} & 61.02\tiny{$\pm$ 0.00} & 30.51\tiny{$\pm$ 0.00} & 100.00\tiny{$\pm$ 0.00} \\
~ & Vikhr 7B IT 5.4 & 5-shot & 60.00\tiny{$\pm$ 0.00} & 81.82\tiny{$\pm$ 0.00} & 69.23\tiny{$\pm$ 0.00} & 63.64\tiny{$\pm$ 0.00} & 36.84\tiny{$\pm$ 0.00} & 46.67\tiny{$\pm$ 0.00} & 57.95\tiny{$\pm$ 0.00} & 100.00\tiny{$\pm$ 0.00} \\
~ & Vikhr 7B IT 5.4 & 0-shot & 33.33\tiny{$\pm$ 0.00} & 4.55\tiny{$\pm$ 0.00} & 8.00\tiny{$\pm$ 0.00} & 44.74\tiny{$\pm$ 0.00} & 89.47\tiny{$\pm$ 0.00} & 59.65\tiny{$\pm$ 0.00} & 33.82\tiny{$\pm$ 0.00} & 100.00\tiny{$\pm$ 0.00} \\
~ & VikhrGemma 2B IT & 5-shot & 66.67\tiny{$\pm$ 0.00} & 27.27\tiny{$\pm$ 0.00} & 38.71\tiny{$\pm$ 0.00} & 50.00\tiny{$\pm$ 0.00} & 84.21\tiny{$\pm$ 0.00} & 62.75\tiny{$\pm$ 0.00} & 50.73\tiny{$\pm$ 0.00} & 100.00\tiny{$\pm$ 0.00} \\
~ & VikhrGemma 2B IT & 0-shot & 0.00\tiny{$\pm$ 0.00} & 0.00\tiny{$\pm$ 0.00} & 0.00\tiny{$\pm$ 0.00} & 43.87\tiny{$\pm$ 0.56} & 90.53\tiny{$\pm$ 2.11} & 59.10\tiny{$\pm$ 0.96} & 29.55\tiny{$\pm$ 0.48} & 100.00\tiny{$\pm$ 0.00} \\ \hline 
    \multirow{14}{*}{AD} & Gemma2 2B IT & 5-shot & 0.00\tiny{$\pm$ 0.00} & 0.00\tiny{$\pm$ 0.00} & 0.00\tiny{$\pm$ 0.00} & 47.37\tiny{$\pm$ 0.00} & 100.00\tiny{$\pm$ 0.00} & 64.29\tiny{$\pm$ 0.00} & 32.14\tiny{$\pm$ 0.00} & 100.00\tiny{$\pm$ 0.00} \\
~ & Gemma2 2B IT & 0-shot & 64.29\tiny{$\pm$ 0.00} & 45.00\tiny{$\pm$ 0.00} & 52.94\tiny{$\pm$ 0.00} & 54.17\tiny{$\pm$ 0.00} & 72.22\tiny{$\pm$ 0.00} & 61.90\tiny{$\pm$ 0.00} & 57.42\tiny{$\pm$ 0.00} & 100.00\tiny{$\pm$ 0.00} \\
~ & Gemma2 9B IT & 5-shot & 100.00\tiny{$\pm$ 0.00} & 5.00\tiny{$\pm$ 0.00} & 9.52\tiny{$\pm$ 0.00} & 48.65\tiny{$\pm$ 0.00} & 100.00\tiny{$\pm$ 0.00} & 65.45\tiny{$\pm$ 0.00} & 37.49\tiny{$\pm$ 0.00} & 100.00\tiny{$\pm$ 0.00} \\
~ & Gemma2 9B IT & 0-shot & 60.87\tiny{$\pm$ 0.00} & 70.00\tiny{$\pm$ 0.00} & 65.12\tiny{$\pm$ 0.00} & 60.00\tiny{$\pm$ 0.00} & 50.00\tiny{$\pm$ 0.00} & 54.55\tiny{$\pm$ 0.00} & 59.83\tiny{$\pm$ 0.00} & 100.00\tiny{$\pm$ 0.00} \\
~ & Qwen2 7B IT & 5-shot & 100.00\tiny{$\pm$ 0.00} & 5.00\tiny{$\pm$ 0.00} & 9.52\tiny{$\pm$ 0.00} & 48.65\tiny{$\pm$ 0.00} & 100.00\tiny{$\pm$ 0.00} & 65.45\tiny{$\pm$ 0.00} & 37.49\tiny{$\pm$ 0.00} & 100.00\tiny{$\pm$ 0.00} \\
~ & Qwen2 7B IT & 0-shot & 67.25\tiny{$\pm$ 1.16} & 80.00\tiny{$\pm$ 0.00} & 73.07\tiny{$\pm$ 0.68} & 71.81\tiny{$\pm$ 0.76} & 56.67\tiny{$\pm$ 2.22} & 63.33\tiny{$\pm$ 1.67} & \textbf{68.20\tiny{$\pm$ 1.17}} & 100.00\tiny{$\pm$ 0.00} \\
~ & SaigaLlama3 8B & 5-shot & 100.00\tiny{$\pm$ 0.00} & 5.00\tiny{$\pm$ 0.00} & 9.52\tiny{$\pm$ 0.00} & 48.65\tiny{$\pm$ 0.00} & 100.00\tiny{$\pm$ 0.00} & 65.45\tiny{$\pm$ 0.00} & 37.49\tiny{$\pm$ 0.00} & 100.00\tiny{$\pm$ 0.00} \\
~ & SaigaLlama3 8B & 0-shot & 54.55\tiny{$\pm$ 0.00} & 30.00\tiny{$\pm$ 0.00} & 38.71\tiny{$\pm$ 0.00} & 48.15\tiny{$\pm$ 0.00} & 72.22\tiny{$\pm$ 0.00} & 57.78\tiny{$\pm$ 0.00} & 48.24\tiny{$\pm$ 0.00} & 100.00\tiny{$\pm$ 0.00} \\
~ & Vikhr 7B IT 0.4 & 5-shot & 50.91\tiny{$\pm$ 1.82} & 30.00\tiny{$\pm$ 0.00} & 37.74\tiny{$\pm$ 0.48} & 46.55\tiny{$\pm$ 0.80} & 67.78\tiny{$\pm$ 2.22} & 55.19\tiny{$\pm$ 1.29} & 46.47\tiny{$\pm$ 0.89} & 100.00\tiny{$\pm$ 0.00} \\
~ & Vikhr 7B IT 0.4 & 0-shot & 13.33\tiny{$\pm$ 26.67} & 2.00\tiny{$\pm$ 4.00} & 3.48\tiny{$\pm$ 6.96} & 46.47\tiny{$\pm$ 1.05} & 94.44\tiny{$\pm$ 0.00} & 62.28\tiny{$\pm$ 0.93} & 32.88\tiny{$\pm$ 3.94} & 100.00\tiny{$\pm$ 0.00} \\
~ & Vikhr 7B IT 5.4 & 5-shot & 66.67\tiny{$\pm$ 0.00} & 10.00\tiny{$\pm$ 0.00} & 17.39\tiny{$\pm$ 0.00} & 48.57\tiny{$\pm$ 0.00} & 94.44\tiny{$\pm$ 0.00} & 64.15\tiny{$\pm$ 0.00} & 40.77\tiny{$\pm$ 0.00} & 100.00\tiny{$\pm$ 0.00} \\
~ & Vikhr 7B IT 5.4 & 0-shot & 83.33\tiny{$\pm$ 0.00} & 25.00\tiny{$\pm$ 0.00} & 38.46\tiny{$\pm$ 0.00} & 53.12\tiny{$\pm$ 0.00} & 94.44\tiny{$\pm$ 0.00} & \textbf{68.00\tiny{$\pm$ 0.00}} & 53.23\tiny{$\pm$ 0.00} & 100.00\tiny{$\pm$ 0.00} \\
~ & VikhrGemma 2B IT & 5-shot & 100.00\tiny{$\pm$ 0.00} & 5.00\tiny{$\pm$ 0.00} & 9.52\tiny{$\pm$ 0.00} & 48.65\tiny{$\pm$ 0.00} & 100.00\tiny{$\pm$ 0.00} & 65.45\tiny{$\pm$ 0.00} & 37.49\tiny{$\pm$ 0.00} & 100.00\tiny{$\pm$ 0.00} \\
~ & VikhrGemma 2B IT & 0-shot & 66.67\tiny{$\pm$ 0.00} & 20.00\tiny{$\pm$ 0.00} & 30.77\tiny{$\pm$ 0.00} & 50.00\tiny{$\pm$ 0.00} & 88.89\tiny{$\pm$ 0.00} & 64.00\tiny{$\pm$ 0.00} & 47.38\tiny{$\pm$ 0.00} & 100.00\tiny{$\pm$ 0.00} \\ \hline 
    \multirow{14}{*}{AC} & Gemma2 2B IT & 5-shot & 72.00\tiny{$\pm$ 2.67} & 24.00\tiny{$\pm$ 0.89} & 36.00\tiny{$\pm$ 1.33} & 50.45\tiny{$\pm$ 0.60} & 88.95\tiny{$\pm$ 1.05} & 64.38\tiny{$\pm$ 0.76} & 33.46\tiny{$\pm$ 0.70} & 98.80\tiny{$\pm$ 0.00} \\
~ & Gemma2 2B IT & 0-shot & 97.50\tiny{$\pm$ 5.00} & 15.56\tiny{$\pm$ 0.00} & 26.82\tiny{$\pm$ 0.20} & 49.87\tiny{$\pm$ 0.27} & 99.47\tiny{$\pm$ 1.05} & 66.43\tiny{$\pm$ 0.47} & 46.63\tiny{$\pm$ 0.34} & 100.00\tiny{$\pm$ 0.00} \\
~ & Gemma2 9B IT & 5-shot & 68.00\tiny{$\pm$ 16.00} & 13.78\tiny{$\pm$ 0.89} & 22.84\tiny{$\pm$ 2.04} & 47.26\tiny{$\pm$ 1.37} & 91.58\tiny{$\pm$ 4.21} & 62.34\tiny{$\pm$ 2.16} & 42.59\tiny{$\pm$ 2.10} & 100.00\tiny{$\pm$ 0.00} \\
~ & Gemma2 9B IT & 0-shot & 54.55\tiny{$\pm$ 0.00} & 26.67\tiny{$\pm$ 0.00} & 35.82\tiny{$\pm$ 0.00} & 47.46\tiny{$\pm$ 0.00} & 73.68\tiny{$\pm$ 0.00} & 57.73\tiny{$\pm$ 0.00} & 31.18\tiny{$\pm$ 0.00} & 97.59\tiny{$\pm$ 0.00} \\
~ & Qwen2 7B IT & 5-shot & 84.89\tiny{$\pm$ 3.56} & 26.22\tiny{$\pm$ 5.33} & 39.85\tiny{$\pm$ 6.96} & 52.08\tiny{$\pm$ 1.72} & 94.74\tiny{$\pm$ 0.00} & \textbf{67.20\tiny{$\pm$ 1.46}} & 53.52\tiny{$\pm$ 4.21} & 100.00\tiny{$\pm$ 0.00} \\
~ & Qwen2 7B IT & 0-shot & 55.93\tiny{$\pm$ 0.24} & 52.44\tiny{$\pm$ 1.78} & 54.11\tiny{$\pm$ 0.87} & 47.55\tiny{$\pm$ 0.09} & 51.05\tiny{$\pm$ 2.11} & 49.22\tiny{$\pm$ 1.00} & 51.67\tiny{$\pm$ 0.07} & 100.00\tiny{$\pm$ 0.00} \\
~ & SaigaLlama3 8B & 5-shot & 100.00\tiny{$\pm$ 0.00} & 13.33\tiny{$\pm$ 0.00} & 23.53\tiny{$\pm$ 0.00} & 49.35\tiny{$\pm$ 0.00} & 100.00\tiny{$\pm$ 0.00} & 66.09\tiny{$\pm$ 0.00} & 44.81\tiny{$\pm$ 0.00} & 100.00\tiny{$\pm$ 0.00} \\
~ & SaigaLlama3 8B & 0-shot & 66.05\tiny{$\pm$ 0.85} & 46.67\tiny{$\pm$ 0.00} & 54.69\tiny{$\pm$ 0.29} & 53.12\tiny{$\pm$ 0.36} & 71.58\tiny{$\pm$ 1.05} & 60.98\tiny{$\pm$ 0.62} & \textbf{57.84\tiny{$\pm$ 0.45}} & 100.00\tiny{$\pm$ 0.00} \\
~ & Vikhr 7B IT 0.4 & 5-shot & 54.20\tiny{$\pm$ 0.30} & 88.89\tiny{$\pm$ 0.00} & 67.34\tiny{$\pm$ 0.23} & 45.56\tiny{$\pm$ 2.22} & 11.05\tiny{$\pm$ 1.05} & 17.78\tiny{$\pm$ 1.52} & 42.56\tiny{$\pm$ 0.88} & 100.00\tiny{$\pm$ 0.00} \\
~ & Vikhr 7B IT 0.4 & 0-shot & 67.48\tiny{$\pm$ 1.05} & 37.78\tiny{$\pm$ 0.00} & 48.43\tiny{$\pm$ 0.27} & 51.55\tiny{$\pm$ 0.34} & 78.42\tiny{$\pm$ 1.05} & 62.21\tiny{$\pm$ 0.58} & 55.32\tiny{$\pm$ 0.43} & 100.00\tiny{$\pm$ 0.00} \\
~ & Vikhr 7B IT 5.4 & 5-shot & 90.67\tiny{$\pm$ 18.67} & 8.89\tiny{$\pm$ 4.44} & 15.33\tiny{$\pm$ 5.67} & 47.12\tiny{$\pm$ 0.76} & 96.32\tiny{$\pm$ 7.37} & 63.22\tiny{$\pm$ 2.37} & 39.28\tiny{$\pm$ 1.65} & 100.00\tiny{$\pm$ 0.00} \\
~ & Vikhr 7B IT 5.4 & 0-shot & 57.50\tiny{$\pm$ 0.00} & 51.11\tiny{$\pm$ 0.00} & 54.12\tiny{$\pm$ 0.00} & 48.84\tiny{$\pm$ 0.00} & 55.26\tiny{$\pm$ 0.00} & 51.85\tiny{$\pm$ 0.00} & 52.98\tiny{$\pm$ 0.00} & 100.00\tiny{$\pm$ 0.00} \\
~ & VikhrGemma 2B IT & 5-shot & 52.75\tiny{$\pm$ 1.02} & 80.00\tiny{$\pm$ 4.44} & 63.56\tiny{$\pm$ 2.12} & 40.00\tiny{$\pm$ 5.00} & 15.26\tiny{$\pm$ 1.05} & 21.94\tiny{$\pm$ 0.56} & 42.75\tiny{$\pm$ 0.78} & 100.00\tiny{$\pm$ 0.00} \\
~ & VikhrGemma 2B IT & 0-shot & 64.26\tiny{$\pm$ 0.88} & 24.00\tiny{$\pm$ 0.89} & 34.94\tiny{$\pm$ 1.08} & 50.00\tiny{$\pm$ 0.00} & 73.68\tiny{$\pm$ 0.00} & 59.57\tiny{$\pm$ 0.00} & 31.51\tiny{$\pm$ 0.36} & 87.71\tiny{$\pm$ 0.48} \\
    \hline
    \end{tabular}
    }
    
\caption{The results for LLMs evaluation on the five main datasets.}
\label{tab:results_llm_reg}
\end{table*}

\section{Full Results on D-all and A-all Datasets}
\label{app:results_kd_ka}
\Cref{tab:results_combined_datasets,tab:results_llm_mmlu_kd_ka,tab:results_llm_reg_kd_ka} show results for all models on D-all and A-all datasets. Among models with SFT Gemma2 2B IT is the best on D-all, while on A-all the best are encoder-based models -- RuRoBERTa and BERT. While in general prompting performs worse than SFT on D-all, the best model (Gemma 2 9B IT) shows comparable performance with AutoML models. On the contrary, prompting on the A-all shows better results than SFT. The best models in various prompting settings are Qwen2 7B IT, SaigaLlama3 8B, and Gemma2 2B IT.

\begin{table*}[t]
\centering
\resizebox{\linewidth}{!}{%
    \begin{tabular}{lllllllll}
    \hline
    %Corpus & Model & \begin{tabular}[c]{@{}l@{}} Precision \\ healthy\end{tabular} & \begin{tabular}[c]{@{}l@{}} Recall \\ healthy\end{tabular} & F1-healthy & \begin{tabular}[c]{@{}l@{}} Precision \\ pathology\end{tabular} & \begin{tabular}[c]{@{}l@{}} Recall \\ pathology\end{tabular} & F1-pathology & F1-macro  \\
    \textbf{Corpus} & \textbf{Model} & \begin{tabular}[c]{@{}l@{}} \textbf{Precision} \\ \textbf{healthy}\end{tabular} & \begin{tabular}[c]{@{}l@{}} \textbf{Recall} \\ \textbf{healthy}\end{tabular} & \textbf{F1-healthy} & \begin{tabular}[c]{@{}l@{}} \textbf{Precision} \\ \textbf{pathology}\end{tabular} & \begin{tabular}[c]{@{}l@{}} \textbf{Recall} \\ \textbf{pathology}\end{tabular} & \textbf{F1-pathology} & \textbf{F1-macro} \\
    \hline
    \multirow{13}{*}{D-all} & Gemma2 2B IT & 92.33\tiny{$\pm$ 0.94} & 97.14\tiny{$\pm$ 1.04} & 94.67\tiny{$\pm$ 0.72} & 76.44\tiny{$\pm$ 6.36} & 52.98\tiny{$\pm$ 6.33} & \textbf{62.33\tiny{$\pm$ 5.60}} & \textbf{78.50\tiny{$\pm$ 3.13}} \\
~ & Gemma2 9B IT & 91.58\tiny{$\pm$ 1.24} & 97.65\tiny{$\pm$ 0.82} & 94.51\tiny{$\pm$ 0.68} & 77.86\tiny{$\pm$ 6.23} & 47.62\tiny{$\pm$ 8.42} & 58.69\tiny{$\pm$ 6.91} & 76.60\tiny{$\pm$ 3.77} \\
~ & Qwen2 7B IT & 91.09\tiny{$\pm$ 0.64} & 97.03\tiny{$\pm$ 0.96} & 93.96\tiny{$\pm$ 0.17} & 72.97\tiny{$\pm$ 5.26} & 44.64\tiny{$\pm$ 4.94} & 54.98\tiny{$\pm$ 2.77} & 74.47\tiny{$\pm$ 1.36} \\
~ & SaigaLlama3 8B & 90.88\tiny{$\pm$ 1.00} & 96.73\tiny{$\pm$ 1.49} & 93.71\tiny{$\pm$ 1.01} & 70.16\tiny{$\pm$ 10.04} & 43.45\tiny{$\pm$ 6.65} & 53.41\tiny{$\pm$ 7.35} & 73.56\tiny{$\pm$ 4.13} \\
~ & Vikhr 7B IT 0.4 & 90.46\tiny{$\pm$ 1.23} & 96.83\tiny{$\pm$ 1.20} & 93.53\tiny{$\pm$ 1.01} & 68.86\tiny{$\pm$ 11.28} & 40.48\tiny{$\pm$ 8.16} & 50.73\tiny{$\pm$ 8.61} & 72.13\tiny{$\pm$ 4.79} \\
~ & Vikhr 7B IT 5.4 & 90.18\tiny{$\pm$ 1.16} & 92.02\tiny{$\pm$ 1.28} & 91.09\tiny{$\pm$ 1.16} & 47.32\tiny{$\pm$ 7.90} & 41.67\tiny{$\pm$ 7.04} & 44.28\tiny{$\pm$ 7.35} & 67.69\tiny{$\pm$ 4.24} \\
~ & VikhrGemma 2B IT & 91.83\tiny{$\pm$ 0.80} & 97.44\tiny{$\pm$ 1.68} & 94.54\tiny{$\pm$ 0.66} & 79.25\tiny{$\pm$ 11.48} & 49.40\tiny{$\pm$ 5.98} & 60.04\tiny{$\pm$ 3.80} & 77.29\tiny{$\pm$ 2.09} \\
~ & RuRoBERTa & 92.48\tiny{$\pm$ 0.63} & 95.50\tiny{$\pm$ 1.16} & 93.96\tiny{$\pm$ 0.42} & 68.19\tiny{$\pm$ 5.05} & 54.76\tiny{$\pm$ 4.45} & 60.48\tiny{$\pm$ 2.31} & 77.22\tiny{$\pm$ 1.27} \\
    ~ & RuBioRoBERTa & 90.74\tiny{$\pm$ 2.52} & 96.93\tiny{$\pm$ 1.66} & 93.69\tiny{$\pm$ 1.04} & 58.50\tiny{$\pm$ 26.99} & 41.67\tiny{$\pm$ 19.20} & 48.61\tiny{$\pm$ 22.33} & 71.15\tiny{$\pm$ 11.60} \\
    ~ & RuBERT & 92.26\tiny{$\pm$ 0.52} & 96.22\tiny{$\pm$ 1.56} & 94.19\tiny{$\pm$ 0.68} & 72.12\tiny{$\pm$ 10.10} & 52.98\tiny{$\pm$ 3.81} & 60.60\tiny{$\pm$ 2.87} & 77.39\tiny{$\pm$ 1.72} \\
    ~ & BERT & 90.16\tiny{$\pm$ 3.31} & 96.63\tiny{$\pm$ 1.69} & 93.22\tiny{$\pm$ 1.43} & 50.52\tiny{$\pm$ 28.15} & 37.50\tiny{$\pm$ 24.38} & 42.50\tiny{$\pm$ 26.53} & 67.86\tiny{$\pm$ 13.90} \\
    ~ & Linguistic features & 88.10\tiny{$\pm$ 0.60} & 88.10\tiny{$\pm$ 3.40} & 88.00\tiny{$\pm$ 1.60} & 51.60\tiny{$\pm$ 5.80} & 50.70\tiny{$\pm$ 4.10} & 50.80\tiny{$\pm$ 2.60} & 69.40\tiny{$\pm$ 1.90}  \\ 
    ~ & TF-IDF & 89.50\tiny{$\pm$ 1.00} & 84.20\tiny{$\pm$ 1.00} & 86.80\tiny{$\pm$ 0.60} & 47.50\tiny{$\pm$ 1.80} & 59.10\tiny{$\pm$ 4.20} & 52.60\tiny{$\pm$ 2.50} & 69.70\tiny{$\pm$ 1.50}  \\ \hline
    \multirow{13}{*}{A-all} & Gemma2 2B IT & 57.61\tiny{$\pm$ 3.21} & 57.09\tiny{$\pm$ 4.28} & 57.27\tiny{$\pm$ 3.15} & 50.61\tiny{$\pm$ 3.62} & 51.11\tiny{$\pm$ 6.00} & 50.77\tiny{$\pm$ 4.48} & 54.02\tiny{$\pm$ 3.39} \\
~ & Gemma2 9B IT & 55.44\tiny{$\pm$ 1.46} & 56.32\tiny{$\pm$ 4.04} & 55.83\tiny{$\pm$ 2.54} & 48.47\tiny{$\pm$ 1.77} & 47.56\tiny{$\pm$ 3.14} & 47.95\tiny{$\pm$ 1.96} & 51.89\tiny{$\pm$ 1.63} \\
~ & Qwen2 7B IT & 56.43\tiny{$\pm$ 1.57} & 56.13\tiny{$\pm$ 3.95} & 56.23\tiny{$\pm$ 2.59} & 49.51\tiny{$\pm$ 2.04} & 49.78\tiny{$\pm$ 2.85} & 49.59\tiny{$\pm$ 1.86} & 52.91\tiny{$\pm$ 1.79} \\
~ & SaigaLlama3 8B & 58.56\tiny{$\pm$ 1.32} & 56.32\tiny{$\pm$ 4.83} & 57.33\tiny{$\pm$ 2.79} & 51.59\tiny{$\pm$ 2.03} & 53.78\tiny{$\pm$ 3.74} & 52.56\tiny{$\pm$ 1.83} & 54.95\tiny{$\pm$ 1.57} \\
~ & Vikhr 7B IT 0.4 & 57.08\tiny{$\pm$ 3.28} & 56.13\tiny{$\pm$ 2.43} & 56.58\tiny{$\pm$ 2.68} & 49.92\tiny{$\pm$ 3.40} & 50.89\tiny{$\pm$ 4.95} & 50.38\tiny{$\pm$ 4.09} & 53.48\tiny{$\pm$ 3.32} \\
~ & Vikhr 7B IT 5.4 & 56.89\tiny{$\pm$ 2.40} & 55.36\tiny{$\pm$ 7.22} & 55.95\tiny{$\pm$ 4.71} & 50.10\tiny{$\pm$ 3.07} & 51.56\tiny{$\pm$ 4.79} & 50.65\tiny{$\pm$ 2.68} & 53.30\tiny{$\pm$ 2.82} \\
~ & VikhrGemma 2B IT & 54.79\tiny{$\pm$ 2.65} & 53.26\tiny{$\pm$ 2.94} & 53.97\tiny{$\pm$ 2.34} & 47.34\tiny{$\pm$ 2.63} & 48.89\tiny{$\pm$ 4.91} & 48.06\tiny{$\pm$ 3.59} & 51.01\tiny{$\pm$ 2.62} \\
    ~ & RuRoBERTa & 61.53\tiny{$\pm$ 3.69} & 60.15\tiny{$\pm$ 13.80} & 59.68\tiny{$\pm$ 4.47} & 54.39\tiny{$\pm$ 0.97} & 54.44\tiny{$\pm$ 18.08} & 52.27\tiny{$\pm$ 11.87} & \textbf{55.97\tiny{$\pm$ 3.88}} \\
    ~ & RuBioRoBERTa & 55.16\tiny{$\pm$ 1.73} & 73.18\tiny{$\pm$ 19.96} & 61.66\tiny{$\pm$ 6.68} & 32.94\tiny{$\pm$ 23.36} & 30.22\tiny{$\pm$ 22.11} & 31.31\tiny{$\pm$ 22.37} & 46.48\tiny{$\pm$ 8.32} \\
    ~ & RuBERT & 57.18\tiny{$\pm$ 1.93} & 51.53\tiny{$\pm$ 5.61} & 54.09\tiny{$\pm$ 3.67} & 49.69\tiny{$\pm$ 1.98} & 55.33\tiny{$\pm$ 4.34} & 52.25\tiny{$\pm$ 2.28} & 53.17\tiny{$\pm$ 2.10} \\
    ~ & BERT & 59.44\tiny{$\pm$ 1.74} & 46.93\tiny{$\pm$ 3.66} & 52.33\tiny{$\pm$ 2.10} & 50.43\tiny{$\pm$ 0.98} & 62.67\tiny{$\pm$ 4.81} & \textbf{55.82\tiny{$\pm$ 2.21}} & 54.08\tiny{$\pm$ 1.08} \\
    ~ & Linguistic features & 53.40\tiny{$\pm$ 2.90} & 39.70\tiny{$\pm$ 19.30} & 42.00\tiny{$\pm$ 17.90} & 46.10\tiny{$\pm$ 2.40} & 60.20\tiny{$\pm$ 20.40} & 51.00\tiny{$\pm$ 7.80} & 46.50\tiny{$\pm$ 6.20}  \\ 
    ~ & TF-IDF & 51.80\tiny{$\pm$ 0.70} & 40.00\tiny{$\pm$ 15.60} & 43.30\tiny{$\pm$ 10.80} & 44.30\tiny{$\pm$ 2.10} & 56.40\tiny{$\pm$ 17.80} & 48.60\tiny{$\pm$ 8.50} & 46.00\tiny{$\pm$ 2.40}  \\
    \hline  
    \end{tabular}
    }
\caption{Results of classification on D-all and A-all datasets for encoder models, AutoML models and LLMs with LoRA.}
\label{tab:results_combined_datasets}
\end{table*}
\begin{table*}[t]
\centering
\resizebox{0.9\linewidth}{!}{%
    \begin{tabular}{lllllllllll}
    \hline
    %Corpus & Model & Mode & \begin{tabular}[c]{@{}l@{}} Precision \\ healthy\end{tabular} & \begin{tabular}[c]{@{}l@{}} Recall \\ healthy\end{tabular} & F1-healthy & \begin{tabular}[c]{@{}l@{}} Precision \\ pathology\end{tabular} & \begin{tabular}[c]{@{}l@{}} Recall \\ pathology\end{tabular} & F1-pathology & F1-macro  \\
    \textbf{Corpus} & \textbf{Model} & \textbf{Mode} & \begin{tabular}[c]{@{}l@{}} \textbf{Precision} \\ \textbf{healthy}\end{tabular} & \begin{tabular}[c]{@{}l@{}} \textbf{Recall} \\ \textbf{healthy}\end{tabular} & \textbf{F1-healthy} & \begin{tabular}[c]{@{}l@{}} \textbf{Precision} \\ \textbf{pathology}\end{tabular} & \begin{tabular}[c]{@{}l@{}} \textbf{Recall} \\ \textbf{pathology}\end{tabular} & \textbf{F1-pathology} & \textbf{F1-macro} \\
    \hline
    \multirow{14}{*}{D-all} & Gemma2 2B IT & 5-shot MMLU & 100.00\tiny{$\pm$ 0.00} & 7.36\tiny{$\pm$ 0.00} & 13.71\tiny{$\pm$ 0.00} & 15.64\tiny{$\pm$ 0.00} & 100.00\tiny{$\pm$ 0.00} & 27.05\tiny{$\pm$ 0.00} & 20.38\tiny{$\pm$ 0.00} \\
~ & Gemma2 2B IT & 0-shot MMLU & 100.00\tiny{$\pm$ 0.00} & 30.55\tiny{$\pm$ 0.25} & 46.80\tiny{$\pm$ 0.29} & 19.83\tiny{$\pm$ 0.06} & 100.00\tiny{$\pm$ 0.00} & 33.10\tiny{$\pm$ 0.08} & 39.95\tiny{$\pm$ 0.18} \\
~ & Gemma2 9B IT & 5-shot MMLU & 88.83\tiny{$\pm$ 0.25} & 96.56\tiny{$\pm$ 0.49} & 92.53\tiny{$\pm$ 0.36} & 59.56\tiny{$\pm$ 4.84} & 29.29\tiny{$\pm$ 1.43} & 39.26\tiny{$\pm$ 2.32} & \textbf{65.90\tiny{$\pm$ 1.34}} \\
~ & Gemma2 9B IT & 0-shot MMLU & 94.87\tiny{$\pm$ 0.00} & 68.10\tiny{$\pm$ 0.00} & 79.29\tiny{$\pm$ 0.00} & 29.73\tiny{$\pm$ 0.00} & 78.57\tiny{$\pm$ 0.00} & \textbf{43.14\tiny{$\pm$ 0.00}} & 61.21\tiny{$\pm$ 0.00} \\
~ & Qwen2 7B IT & 5-shot MMLU & 94.74\tiny{$\pm$ 0.00} & 22.09\tiny{$\pm$ 0.00} & 35.82\tiny{$\pm$ 0.00} & 16.99\tiny{$\pm$ 0.00} & 92.86\tiny{$\pm$ 0.00} & 28.73\tiny{$\pm$ 0.00} & 32.28\tiny{$\pm$ 0.00} \\
~ & Qwen2 7B IT & 0-shot MMLU & 95.66\tiny{$\pm$ 0.08} & 54.11\tiny{$\pm$ 0.98} & 69.12\tiny{$\pm$ 0.83} & 24.30\tiny{$\pm$ 0.38} & 85.71\tiny{$\pm$ 0.00} & 37.86\tiny{$\pm$ 0.47} & 53.49\tiny{$\pm$ 0.65} \\
~ & SaigaLlama3 8B & 5-shot MMLU & 100.00\tiny{$\pm$ 0.00} & 1.10\tiny{$\pm$ 0.25} & 2.18\tiny{$\pm$ 0.48} & 14.80\tiny{$\pm$ 0.03} & 100.00\tiny{$\pm$ 0.00} & 25.78\tiny{$\pm$ 0.05} & 13.98\tiny{$\pm$ 0.26} \\
~ & SaigaLlama3 8B & 0-shot MMLU & 92.98\tiny{$\pm$ 0.04} & 40.61\tiny{$\pm$ 0.25} & 56.53\tiny{$\pm$ 0.24} & 19.20\tiny{$\pm$ 0.06} & 82.14\tiny{$\pm$ 0.00} & 31.12\tiny{$\pm$ 0.08} & 43.83\tiny{$\pm$ 0.16} \\
~ & Vikhr 7B IT 0.4 & 5-shot MMLU & 88.97\tiny{$\pm$ 0.24} & 90.06\tiny{$\pm$ 0.25} & 89.51\tiny{$\pm$ 0.24} & 37.69\tiny{$\pm$ 1.54} & 35.00\tiny{$\pm$ 1.43} & 36.30\tiny{$\pm$ 1.48} & 62.90\tiny{$\pm$ 0.86} \\
~ & Vikhr 7B IT 0.4 & 0-shot MMLU & 100.00\tiny{$\pm$ 0.00} & 2.45\tiny{$\pm$ 0.00} & 4.79\tiny{$\pm$ 0.00} & 14.97\tiny{$\pm$ 0.00} & 100.00\tiny{$\pm$ 0.00} & 26.05\tiny{$\pm$ 0.00} & 15.42\tiny{$\pm$ 0.00} \\
~ & Vikhr 7B IT 5.4 & 5-shot MMLU & 96.88\tiny{$\pm$ 0.00} & 57.06\tiny{$\pm$ 0.00} & 71.81\tiny{$\pm$ 0.00} & 26.32\tiny{$\pm$ 0.00} & 89.29\tiny{$\pm$ 0.00} & 40.65\tiny{$\pm$ 0.00} & 56.23\tiny{$\pm$ 0.00} \\
~ & Vikhr 7B IT 5.4 & 0-shot MMLU & 90.94\tiny{$\pm$ 0.06} & 30.80\tiny{$\pm$ 0.25} & 46.01\tiny{$\pm$ 0.28} & 16.94\tiny{$\pm$ 0.05} & 82.14\tiny{$\pm$ 0.00} & 28.08\tiny{$\pm$ 0.07} & 37.05\tiny{$\pm$ 0.18} \\
~ & VikhrGemma 2B IT & 5-shot MMLU & 84.93\tiny{$\pm$ 0.00} & 38.04\tiny{$\pm$ 0.00} & 52.54\tiny{$\pm$ 0.00} & 14.41\tiny{$\pm$ 0.00} & 60.71\tiny{$\pm$ 0.00} & 23.29\tiny{$\pm$ 0.00} & 37.92\tiny{$\pm$ 0.00} \\
~ & VikhrGemma 2B IT & 0-shot MMLU & 85.25\tiny{$\pm$ 0.03} & 99.26\tiny{$\pm$ 0.25} & 91.72\tiny{$\pm$ 0.12} & 0.00\tiny{$\pm$ 0.00} & 0.00\tiny{$\pm$ 0.00} & 0.00\tiny{$\pm$ 0.00} & 45.86\tiny{$\pm$ 0.06} \\ \hline
    \multirow{14}{*}{A-all} & Gemma2 2B IT & 5-shot MMLU & 62.86\tiny{$\pm$ 0.71} & 6.67\tiny{$\pm$ 1.84} & 11.99\tiny{$\pm$ 2.92} & 46.86\tiny{$\pm$ 0.22} & 95.47\tiny{$\pm$ 1.07} & 62.86\tiny{$\pm$ 0.04} & 37.42\tiny{$\pm$ 1.44} \\
~ & Gemma2 2B IT & 0-shot MMLU & 66.67\tiny{$\pm$ 0.00} & 11.49\tiny{$\pm$ 0.00} & 19.61\tiny{$\pm$ 0.00} & 47.62\tiny{$\pm$ 0.00} & 93.33\tiny{$\pm$ 0.00} & 63.06\tiny{$\pm$ 0.00} & 41.34\tiny{$\pm$ 0.00} \\
~ & Gemma2 9B IT & 5-shot MMLU & 57.52\tiny{$\pm$ 0.19} & 63.45\tiny{$\pm$ 5.06} & 60.24\tiny{$\pm$ 2.04} & 51.94\tiny{$\pm$ 1.03} & 45.60\tiny{$\pm$ 4.80} & 48.36\tiny{$\pm$ 2.58} & 54.30\tiny{$\pm$ 0.27} \\
~ & Gemma2 9B IT & 0-shot MMLU & 63.85\tiny{$\pm$ 0.77} & 38.16\tiny{$\pm$ 0.46} & 47.77\tiny{$\pm$ 0.58} & 51.09\tiny{$\pm$ 0.36} & 74.93\tiny{$\pm$ 0.53} & 60.76\tiny{$\pm$ 0.43} & 54.26\tiny{$\pm$ 0.50} \\
~ & Qwen2 7B IT & 5-shot MMLU & 65.42\tiny{$\pm$ 1.30} & 45.29\tiny{$\pm$ 5.52} & 53.28\tiny{$\pm$ 3.06} & 53.22\tiny{$\pm$ 0.77} & 72.00\tiny{$\pm$ 5.33} & 61.08\tiny{$\pm$ 1.61} & 57.18\tiny{$\pm$ 0.73} \\
~ & Qwen2 7B IT & 0-shot MMLU & 62.67\tiny{$\pm$ 0.06} & 64.83\tiny{$\pm$ 1.38} & 63.72\tiny{$\pm$ 0.65} & 57.51\tiny{$\pm$ 0.47} & 55.20\tiny{$\pm$ 1.07} & 56.32\tiny{$\pm$ 0.32} & \textbf{60.02\tiny{$\pm$ 0.17}} \\
~ & SaigaLlama3 8B & 5-shot MMLU & 80.51\tiny{$\pm$ 5.64} & 11.26\tiny{$\pm$ 0.46} & 19.76\tiny{$\pm$ 0.88} & 48.46\tiny{$\pm$ 0.41} & 96.80\tiny{$\pm$ 1.07} & \textbf{64.59\tiny{$\pm$ 0.60}} & 42.18\tiny{$\pm$ 0.74} \\
~ & SaigaLlama3 8B & 0-shot MMLU & 67.91\tiny{$\pm$ 0.93} & 33.56\tiny{$\pm$ 0.46} & 44.92\tiny{$\pm$ 0.62} & 51.43\tiny{$\pm$ 0.34} & 81.60\tiny{$\pm$ 0.53} & 63.09\tiny{$\pm$ 0.41} & 54.01\tiny{$\pm$ 0.51} \\
~ & Vikhr 7B IT 0.4 & 5-shot MMLU & 56.68\tiny{$\pm$ 1.42} & 80.00\tiny{$\pm$ 8.28} & 66.10\tiny{$\pm$ 1.52} & 54.26\tiny{$\pm$ 2.13} & 28.53\tiny{$\pm$ 12.27} & 35.58\tiny{$\pm$ 14.09} & 50.84\tiny{$\pm$ 6.29} \\
~ & Vikhr 7B IT 0.4 & 0-shot MMLU & 64.91\tiny{$\pm$ 1.21} & 22.07\tiny{$\pm$ 0.46} & 32.93\tiny{$\pm$ 0.34} & 48.79\tiny{$\pm$ 0.16} & 86.13\tiny{$\pm$ 1.07} & 62.29\tiny{$\pm$ 0.41} & 47.61\tiny{$\pm$ 0.04} \\
~ & Vikhr 7B IT 5.4 & 5-shot MMLU & 62.12\tiny{$\pm$ 4.24} & 22.07\tiny{$\pm$ 4.14} & 32.16\tiny{$\pm$ 4.54} & 48.08\tiny{$\pm$ 0.10} & 83.73\tiny{$\pm$ 4.80} & 61.04\tiny{$\pm$ 1.30} & 46.60\tiny{$\pm$ 1.62} \\
~ & Vikhr 7B IT 5.4 & 0-shot MMLU & 59.18\tiny{$\pm$ 0.00} & 33.33\tiny{$\pm$ 0.00} & 42.65\tiny{$\pm$ 0.00} & 48.67\tiny{$\pm$ 0.00} & 73.33\tiny{$\pm$ 0.00} & 58.51\tiny{$\pm$ 0.00} & 50.58\tiny{$\pm$ 0.00} \\
~ & VikhrGemma 2B IT & 5-shot MMLU & 53.67\tiny{$\pm$ 0.50} & 97.93\tiny{$\pm$ 1.84} & 69.32\tiny{$\pm$ 0.06} & 11.67\tiny{$\pm$ 23.33} & 1.87\tiny{$\pm$ 3.73} & 3.22\tiny{$\pm$ 6.44} & 36.27\tiny{$\pm$ 3.19} \\
~ & VikhrGemma 2B IT & 0-shot MMLU & 53.42\tiny{$\pm$ 0.00} & 98.85\tiny{$\pm$ 0.00} & 69.35\tiny{$\pm$ 0.00} & 0.00\tiny{$\pm$ 0.00} & 0.00\tiny{$\pm$ 0.00} & 0.00\tiny{$\pm$ 0.00} & 34.68\tiny{$\pm$ 0.00} \\
    \hline
    \end{tabular}
    }
    
\caption{The results for LLMs MMLU-style evaluation, D-all and A-all datasets.}
\label{tab:results_llm_mmlu_kd_ka}
\end{table*}
\begin{table*}[t]
\centering
\resizebox{0.9\linewidth}{!}{%
    \begin{tabular}{llllllllllll}
    \hline
    %Corpus & Model & Mode & \begin{tabular}[c]{@{}l@{}} Precision \\ healthy\end{tabular} & \begin{tabular}[c]{@{}l@{}} Recall \\ healthy\end{tabular} & F1-healthy & \begin{tabular}[c]{@{}l@{}} Precision \\ pathology\end{tabular} & \begin{tabular}[c]{@{}l@{}} Recall \\ pathology\end{tabular} & F1-pathology & F1-macro & \begin{tabular}[c]{@{}l@{}} Matched \\ percentage\end{tabular} \\
    \textbf{Corpus} & \textbf{Model} & \textbf{Mode} & \begin{tabular}[c]{@{}l@{}} \textbf{Precision} \\ \textbf{healthy}\end{tabular} & \begin{tabular}[c]{@{}l@{}} \textbf{Recall} \\ \textbf{healthy}\end{tabular} & \textbf{F1-healthy} & \begin{tabular}[c]{@{}l@{}} \textbf{Precision} \\ \textbf{pathology}\end{tabular} & \begin{tabular}[c]{@{}l@{}} \textbf{Recall} \\ \textbf{pathology}\end{tabular} & \textbf{F1-pathology} & \textbf{F1-macro} & \begin{tabular}[c]{@{}l@{}} \textbf{Matched} \\ \textbf{percentage}\end{tabular} \\
    \hline
    \multirow{14}{*}{D-all} & Gemma2 2B IT & 5-shot & 94.25\tiny{$\pm$ 0.07} & 20.12\tiny{$\pm$ 0.25} & 33.16\tiny{$\pm$ 0.34} & 16.65\tiny{$\pm$ 0.04} & 92.86\tiny{$\pm$ 0.00} & 28.23\tiny{$\pm$ 0.06} & 30.70\tiny{$\pm$ 0.20} & 100.00\tiny{$\pm$ 0.00} \\
~ & Gemma2 2B IT & 0-shot & 96.06\tiny{$\pm$ 0.56} & 41.84\tiny{$\pm$ 0.25} & 58.29\tiny{$\pm$ 0.34} & 21.00\tiny{$\pm$ 0.33} & 90.00\tiny{$\pm$ 1.43} & 34.05\tiny{$\pm$ 0.54} & 46.17\tiny{$\pm$ 0.44} & 100.00\tiny{$\pm$ 0.00} \\
~ & Gemma2 9B IT & 5-shot & 98.17\tiny{$\pm$ 0.03} & 32.88\tiny{$\pm$ 0.49} & 49.26\tiny{$\pm$ 0.56} & 19.80\tiny{$\pm$ 0.12} & 96.43\tiny{$\pm$ 0.00} & 32.85\tiny{$\pm$ 0.16} & 41.06\tiny{$\pm$ 0.36} & 100.00\tiny{$\pm$ 0.00} \\
~ & Gemma2 9B IT & 0-shot & 93.28\tiny{$\pm$ 0.00} & 76.69\tiny{$\pm$ 0.00} & 84.18\tiny{$\pm$ 0.00} & 33.33\tiny{$\pm$ 0.00} & 67.86\tiny{$\pm$ 0.00} & \textbf{44.71\tiny{$\pm$ 0.00}} & \textbf{64.44\tiny{$\pm$ 0.00}} & 100.00\tiny{$\pm$ 0.00} \\
~ & Qwen2 7B IT & 5-shot & 94.59\tiny{$\pm$ 0.00} & 21.47\tiny{$\pm$ 0.00} & 35.00\tiny{$\pm$ 0.00} & 16.88\tiny{$\pm$ 0.00} & 92.86\tiny{$\pm$ 0.00} & 28.57\tiny{$\pm$ 0.00} & 31.79\tiny{$\pm$ 0.00} & 100.00\tiny{$\pm$ 0.00} \\
~ & Qwen2 7B IT & 0-shot & 95.64\tiny{$\pm$ 0.48} & 56.56\tiny{$\pm$ 0.98} & 71.08\tiny{$\pm$ 0.91} & 25.17\tiny{$\pm$ 0.73} & 85.00\tiny{$\pm$ 1.43} & 38.84\tiny{$\pm$ 1.02} & 54.96\tiny{$\pm$ 0.96} & 100.00\tiny{$\pm$ 0.00} \\
~ & SaigaLlama3 8B & 5-shot & 100.00\tiny{$\pm$ 0.00} & 7.48\tiny{$\pm$ 0.98} & 13.91\tiny{$\pm$ 1.72} & 15.66\tiny{$\pm$ 0.14} & 100.00\tiny{$\pm$ 0.00} & 27.08\tiny{$\pm$ 0.21} & 20.50\tiny{$\pm$ 0.97} & 100.00\tiny{$\pm$ 0.00} \\
~ & SaigaLlama3 8B & 0-shot & 96.18\tiny{$\pm$ 0.06} & 15.46\tiny{$\pm$ 0.25} & 26.64\tiny{$\pm$ 0.37} & 16.38\tiny{$\pm$ 0.04} & 96.43\tiny{$\pm$ 0.00} & 28.01\tiny{$\pm$ 0.06} & 27.32\tiny{$\pm$ 0.21} & 100.00\tiny{$\pm$ 0.00} \\
~ & Vikhr 7B IT 0.4 & 5-shot & 93.02\tiny{$\pm$ 0.08} & 40.86\tiny{$\pm$ 0.49} & 56.78\tiny{$\pm$ 0.49} & 19.26\tiny{$\pm$ 0.13} & 82.14\tiny{$\pm$ 0.00} & 31.21\tiny{$\pm$ 0.17} & 43.99\tiny{$\pm$ 0.33} & 100.00\tiny{$\pm$ 0.00} \\
~ & Vikhr 7B IT 0.4 & 0-shot & 100.00\tiny{$\pm$ 0.00} & 2.45\tiny{$\pm$ 0.00} & 4.79\tiny{$\pm$ 0.00} & 14.97\tiny{$\pm$ 0.00} & 100.00\tiny{$\pm$ 0.00} & 26.05\tiny{$\pm$ 0.00} & 15.42\tiny{$\pm$ 0.00} & 100.00\tiny{$\pm$ 0.00} \\
~ & Vikhr 7B IT 5.4 & 5-shot & 96.88\tiny{$\pm$ 0.00} & 57.06\tiny{$\pm$ 0.00} & 71.81\tiny{$\pm$ 0.00} & 26.32\tiny{$\pm$ 0.00} & 89.29\tiny{$\pm$ 0.00} & 40.65\tiny{$\pm$ 0.00} & 56.23\tiny{$\pm$ 0.00} & 100.00\tiny{$\pm$ 0.00} \\
~ & Vikhr 7B IT 5.4 & 0-shot & 90.94\tiny{$\pm$ 0.06} & 30.80\tiny{$\pm$ 0.25} & 46.01\tiny{$\pm$ 0.28} & 16.94\tiny{$\pm$ 0.05} & 82.14\tiny{$\pm$ 0.00} & 28.08\tiny{$\pm$ 0.07} & 37.05\tiny{$\pm$ 0.18} & 100.00\tiny{$\pm$ 0.00} \\
~ & VikhrGemma 2B IT & 5-shot & 93.90\tiny{$\pm$ 0.08} & 18.90\tiny{$\pm$ 0.25} & 31.46\tiny{$\pm$ 0.35} & 16.43\tiny{$\pm$ 0.04} & 92.86\tiny{$\pm$ 0.00} & 27.93\tiny{$\pm$ 0.06} & 29.69\tiny{$\pm$ 0.20} & 100.00\tiny{$\pm$ 0.00} \\
~ & VikhrGemma 2B IT & 0-shot & 96.73\tiny{$\pm$ 0.09} & 18.16\tiny{$\pm$ 0.49} & 30.58\tiny{$\pm$ 0.70} & 17.59\tiny{$\pm$ 0.35} & 95.71\tiny{$\pm$ 1.43} & 29.71\tiny{$\pm$ 0.57} & 20.10\tiny{$\pm$ 0.43} & 95.81\tiny{$\pm$ 0.00} \\ \hline
    \multirow{14}{*}{A-all} & Gemma2 2B IT & 5-shot & 69.78\tiny{$\pm$ 1.11} & 20.69\tiny{$\pm$ 0.00} & 31.92\tiny{$\pm$ 0.11} & 49.63\tiny{$\pm$ 0.00} & 89.60\tiny{$\pm$ 0.53} & 63.88\tiny{$\pm$ 0.14} & 35.14\tiny{$\pm$ 6.50} & 99.51\tiny{$\pm$ 0.25} \\
~ & Gemma2 2B IT & 0-shot & 71.44\tiny{$\pm$ 0.87} & 26.44\tiny{$\pm$ 0.00} & 38.59\tiny{$\pm$ 0.13} & 50.69\tiny{$\pm$ 0.15} & 87.73\tiny{$\pm$ 0.53} & \textbf{64.26\tiny{$\pm$ 0.27}} & 51.42\tiny{$\pm$ 0.20} & 100.00\tiny{$\pm$ 0.00} \\
~ & Gemma2 9B IT & 5-shot & 67.80\tiny{$\pm$ 0.57} & 38.62\tiny{$\pm$ 3.68} & 49.09\tiny{$\pm$ 2.66} & 52.52\tiny{$\pm$ 0.70} & 78.67\tiny{$\pm$ 2.67} & 62.96\tiny{$\pm$ 0.40} & 56.02\tiny{$\pm$ 1.13} & 100.00\tiny{$\pm$ 0.00} \\
~ & Gemma2 9B IT & 0-shot & 60.61\tiny{$\pm$ 0.00} & 45.98\tiny{$\pm$ 0.00} & 52.29\tiny{$\pm$ 0.00} & 52.13\tiny{$\pm$ 0.00} & 65.33\tiny{$\pm$ 0.00} & 57.99\tiny{$\pm$ 0.00} & 36.76\tiny{$\pm$ 0.00} & 98.77\tiny{$\pm$ 0.00} \\
~ & Qwen2 7B IT & 5-shot & 67.93\tiny{$\pm$ 2.23} & 34.25\tiny{$\pm$ 1.84} & 45.47\tiny{$\pm$ 1.01} & 51.52\tiny{$\pm$ 0.30} & 81.07\tiny{$\pm$ 3.20} & 62.98\tiny{$\pm$ 1.22} & 54.22\tiny{$\pm$ 0.11} & 100.00\tiny{$\pm$ 0.00} \\
~ & Qwen2 7B IT & 0-shot & 60.01\tiny{$\pm$ 0.41} & 69.66\tiny{$\pm$ 0.92} & 64.47\tiny{$\pm$ 0.17} & 56.72\tiny{$\pm$ 0.10} & 46.13\tiny{$\pm$ 1.60} & 50.87\tiny{$\pm$ 0.99} & \textbf{57.67\tiny{$\pm$ 0.41}} & 100.00\tiny{$\pm$ 0.00} \\
~ & SaigaLlama3 8B & 5-shot & 69.34\tiny{$\pm$ 0.46} & 17.70\tiny{$\pm$ 1.38} & 28.18\tiny{$\pm$ 1.83} & 48.79\tiny{$\pm$ 0.27} & 90.93\tiny{$\pm$ 0.53} & 63.50\tiny{$\pm$ 0.10} & 45.84\tiny{$\pm$ 0.96} & 100.00\tiny{$\pm$ 0.00} \\
~ & SaigaLlama3 8B & 0-shot & 61.30\tiny{$\pm$ 0.87} & 32.41\tiny{$\pm$ 0.46} & 42.41\tiny{$\pm$ 0.60} & 49.31\tiny{$\pm$ 0.34} & 76.27\tiny{$\pm$ 0.53} & 59.90\tiny{$\pm$ 0.42} & 51.15\tiny{$\pm$ 0.51} & 100.00\tiny{$\pm$ 0.00} \\
~ & Vikhr 7B IT 0.4 & 5-shot & 56.94\tiny{$\pm$ 1.27} & 68.05\tiny{$\pm$ 5.06} & 61.86\tiny{$\pm$ 1.14} & 51.63\tiny{$\pm$ 1.49} & 40.00\tiny{$\pm$ 8.00} & 44.69\tiny{$\pm$ 6.27} & 53.28\tiny{$\pm$ 2.56} & 100.00\tiny{$\pm$ 0.00} \\
~ & Vikhr 7B IT 0.4 & 0-shot & 63.04\tiny{$\pm$ 0.15} & 20.00\tiny{$\pm$ 0.92} & 30.36\tiny{$\pm$ 1.06} & 48.22\tiny{$\pm$ 0.13} & 86.40\tiny{$\pm$ 0.53} & 61.89\tiny{$\pm$ 0.03} & 46.12\tiny{$\pm$ 0.52} & 100.00\tiny{$\pm$ 0.00} \\
~ & Vikhr 7B IT 5.4 & 5-shot & 62.12\tiny{$\pm$ 4.24} & 22.07\tiny{$\pm$ 4.14} & 32.16\tiny{$\pm$ 4.54} & 48.08\tiny{$\pm$ 0.10} & 83.73\tiny{$\pm$ 4.80} & 61.04\tiny{$\pm$ 1.30} & 46.60\tiny{$\pm$ 1.62} & 100.00\tiny{$\pm$ 0.00} \\
~ & Vikhr 7B IT 5.4 & 0-shot & 59.18\tiny{$\pm$ 0.00} & 33.33\tiny{$\pm$ 0.00} & 42.65\tiny{$\pm$ 0.00} & 48.67\tiny{$\pm$ 0.00} & 73.33\tiny{$\pm$ 0.00} & 58.51\tiny{$\pm$ 0.00} & 50.58\tiny{$\pm$ 0.00} & 100.00\tiny{$\pm$ 0.00} \\
~ & VikhrGemma 2B IT & 5-shot & 51.38\tiny{$\pm$ 2.77} & 24.14\tiny{$\pm$ 9.20} & 32.22\tiny{$\pm$ 8.23} & 46.04\tiny{$\pm$ 1.46} & 74.40\tiny{$\pm$ 5.87} & 56.71\tiny{$\pm$ 0.87} & 41.01\tiny{$\pm$ 3.23} & 99.88\tiny{$\pm$ 0.25} \\
~ & VikhrGemma 2B IT & 0-shot & 60.17\tiny{$\pm$ 0.35} & 17.01\tiny{$\pm$ 0.46} & 26.52\tiny{$\pm$ 0.53} & 48.11\tiny{$\pm$ 0.16} & 81.60\tiny{$\pm$ 0.53} & 60.53\tiny{$\pm$ 0.28} & 29.02\tiny{$\pm$ 0.09} & 93.70\tiny{$\pm$ 0.25} \\
    \hline
    \end{tabular}
    }
    
\caption{The results for LLMs evaluation, D-all and A-all datasets.}
\label{tab:results_llm_reg_kd_ka}
\end{table*}

\section{Psychologists Evaluation Setup}
\label{app:interpretation_setup}
During the LLM generation interpretation, psychologists used the following rating scale for explanations, given by LLM: 1 - completely erroneous, 2 - mostly erroneous, 3 - partially erroneous, partially correct, 4 - mostly correct, 5 - completely correct. On this scale, the psychologists rated LLM explanations as 2.84 out of 5 with Fleiss' $\kappa=0.39$, which shows fair inter-annotator agreement. To better categorize errors in LLM explanations, psychologists also provided a list of the most common errors in each generation with detailed explanations of the error type. This list is provided in \Cref{tab:llm_error_types}.
\begin{table*}[!ht]
\centering
\footnotesize
\begin{tabularx}{\textwidth}{>{\hsize=.2\hsize\linewidth=\hsize}X
>{\hsize=.4\hsize\linewidth=\hsize}X>{\hsize=.4\hsize\linewidth=\hsize}X}
\toprule
\textbf{Error name}               & \textbf{Description}    & \textbf{Example}                                                                                                                                                                                                                                                 \\

\midrule 
1. Tautology                    &   The final part of the explanation repeats the initial part in other words without any proof.   & This text \textbf{indicates that the author has depression}, as he describes his thoughts and feelings, which are \textbf{characteristic of depressive disorders}.                                                                                                                                                         \\ 
 2. Groundless generalization                   & The patient's experience from the text is defined as a sign of depression, while this experience on its own, without the context, is not specific to depression.  &  \textbf{The desire to return to the lost state of happiness and happy life}, which is also a \textbf{sign of depression}.                                                                                                                             \\ 
3. False conclusion                    & The false inference is derived from the text statement.    &  The author also mentions that his parents, who he considers to be positive, \textbf{were unable to correct his behavior}. \textit{In the original text there is no signs of parents intention to correct authors behaviour}.                                                                                                                     \\ 
4. Confabulation                   & The explanation contains evidence, which is not mentioned in the context.      & The text describes a \textbf{deep dissatisfaction with the world, people and their actions} \textit{included in the text}, and also \textbf{expresses a desire to be happy} \textit{not in the text} and \textbf{enjoy the little things} \textit{not in the text}.                                                                                       \\ 
5. Distortion of medical understanding of depression                 & Misconception about depression.      & She also expresses a desire to ... \textbf{plan long-term plans, which also indicates the presence of a depressive disorder}. \textit{The long-term planning cannot be considered as a sign of depression}.                                                                            \\ 
6. Incompleteness of selected signs of depression            & Of the several significant signs of depression only one or two signs (mostly minor) are highlighted.       &      Thank you for this test, so that I could repeat all this to myself once again and \textbf{think about the rope}. \textit{This significant sign does not mentioned in the explanation}.   \\
%\midrule 
%1. Confabulation                   & The explanation contains evidence, which is not mentioned in the context.      & The text describes a \textbf{deep dissatisfaction with the world, people and their actions} \textit{included in the text}, and also \textbf{expresses a desire to be happy} \textit{not in the text} and \textbf{enjoy the little things} \textit{not in the text}.                                                                                       \\ 
%2. Distortion of medical understanding of depression                 & Misconception about depression.      & She also expresses a desire to ... \textbf{plan long-term plans, which also indicates the presence of a depressive disorder}. \textit{The long-term planning cannot be considered as a sign of depression}.                                                                            \\ 
%3. Incompleteness of selected signs of depression            & Of the several significant signs of depression only one or two signs (mostly minor) are highlighted.       &      Thank you for this test, so that I could repeat all this to myself once again and \textbf{think about the rope}. \textit{This significant sign does not mentioned in the explanation}.   

\bottomrule
\end{tabularx}
\caption{\label{tab:llm_error_types} Detailed description of errors in LLM explanations from the perspective of trained psychologists. The bold font indicates the significant part of the examples, illustrating the error type; the italic font highlights the psychologist's notes for the examples.}
\end{table*}

We also conducted an additional experiment with a clinically informed prompt to explore how the additional information from psychologists can affect the quality of explanations. To do so, we asked psychologists to select parts of the text that can serve as depression markers for texts that were used in a 5-shot prompt for the DE dataset. After we performed a 5-shot evaluation of the best model with a modified 5-shot prompt, which now includes depression markers along with the target label. The results are presented in~\Cref{tab:results_psy_informed_prompt}.
\begin{table*}[t]
\centering
\resizebox{\linewidth}{!}{%
    \begin{tabular}{llllllllll}
    \hline
    %Corpus & Mode & Model & \begin{tabular}[c]{@{}l@{}} Precision \\ healthy\end{tabular} & \begin{tabular}[c]{@{}l@{}} Recall \\ healthy\end{tabular} & F1-healthy & \begin{tabular}[c]{@{}l@{}} Precision \\ pathology\end{tabular} & \begin{tabular}[c]{@{}l@{}} Recall \\ pathology\end{tabular} & F1-pathology & F1-macro  \\
    \textbf{Corpus} & \textbf{Model} & \textbf{Mode} & \begin{tabular}[c]{@{}l@{}} \textbf{Precision} \\ \textbf{healthy}\end{tabular} & \begin{tabular}[c]{@{}l@{}} \textbf{Recall} \\ \textbf{healthy}\end{tabular} & \textbf{F1-healthy} & \begin{tabular}[c]{@{}l@{}} \textbf{Precision} \\ \textbf{pathology}\end{tabular} & \begin{tabular}[c]{@{}l@{}} \textbf{Recall} \\ \textbf{pathology}\end{tabular} & \textbf{F1-pathology} & \textbf{F1-macro} \\
    \hline
    \multirow{3}{*}{DE} & 0-shot & Vikhr 7B IT 5.4 & 71.43\tiny{$\pm$ 0.00} & 5.56\tiny{$\pm$ 0.00} & 10.31\tiny{$\pm$ 0.00} & 19.05\tiny{$\pm$ 0.00} & 90.91\tiny{$\pm$ 0.00} & 31.50\tiny{$\pm$ 0.00} & 20.90\tiny{$\pm$ 0.00} \\
~ & 5-shot & Vikhr 7B IT 5.4 & 87.06\tiny{$\pm$ 0.00} & 82.22\tiny{$\pm$ 0.00} & 84.57\tiny{$\pm$ 0.00} & 40.74\tiny{$\pm$ 0.00} & 50.00\tiny{$\pm$ 0.00} & 44.90\tiny{$\pm$ 0.00} & 64.73\tiny{$\pm$ 0.00} \\
~ & 5-shot clinically informed & Vikhr 7B IT 5.4 & 87.64\tiny{$\pm$ 0.00} & 86.67\tiny{$\pm$ 0.00} & 87.15\tiny{$\pm$ 0.00} & 47.83\tiny{$\pm$ 0.00} & 50.00\tiny{$\pm$ 0.00} & 48.89\tiny{$\pm$ 0.00} & 68.02\tiny{$\pm$ 0.00} \\
    \hline
    \end{tabular}
    }
    
\caption{Comparison of the results of the best generative model on DE in various settings (mean $\pm$ std).}
\label{tab:results_psy_informed_prompt}
\end{table*}
The adapted prompt leads to a slightly better F1-macro score; however, such a prompt requires qualified psychologists to label several texts for each investigated dataset. These results provide an interesting direction for future work and show that the overall quality of LLMs in mental disorders detection tasks can be improved with the help of psychologists.

To further investigate the quality of model explanations with clinically informed prompts, we asked psychologists to conduct the evaluation of results as in~\Cref{subsec:results_table_5}. During evaluation, the psychologists set the average score to 2.47 out of 5, which is lower than for prompting without a clinically informed prompt. Due to the significant difference in the explanations of LLM, the mistakes were categorized into other groups, namely: (1) confabulation, (2) undifferentiation (inability to select symptomatic parts of the text), and (3) incompleteness of explanation. With a clinically informed prompt, each of the explanations contains at least one of the mentioned mistakes. The most common types of mistakes are confabulation and undifferentiation, which are found in 47\% of the explanations. The incompleteness of the explanation occurs in 40\% of the explanations. Overall, these results show that even with the clinically informed prompt, modern LLMs are unable to generate explanations that meet the requirements of clinicians.

\section{Feature Importance Ablation}
\label{app:feature_importance}
\begin{figure*}[h]
    \centering
    \begin{subfigure}
        \centering
        \includegraphics[scale=0.3]{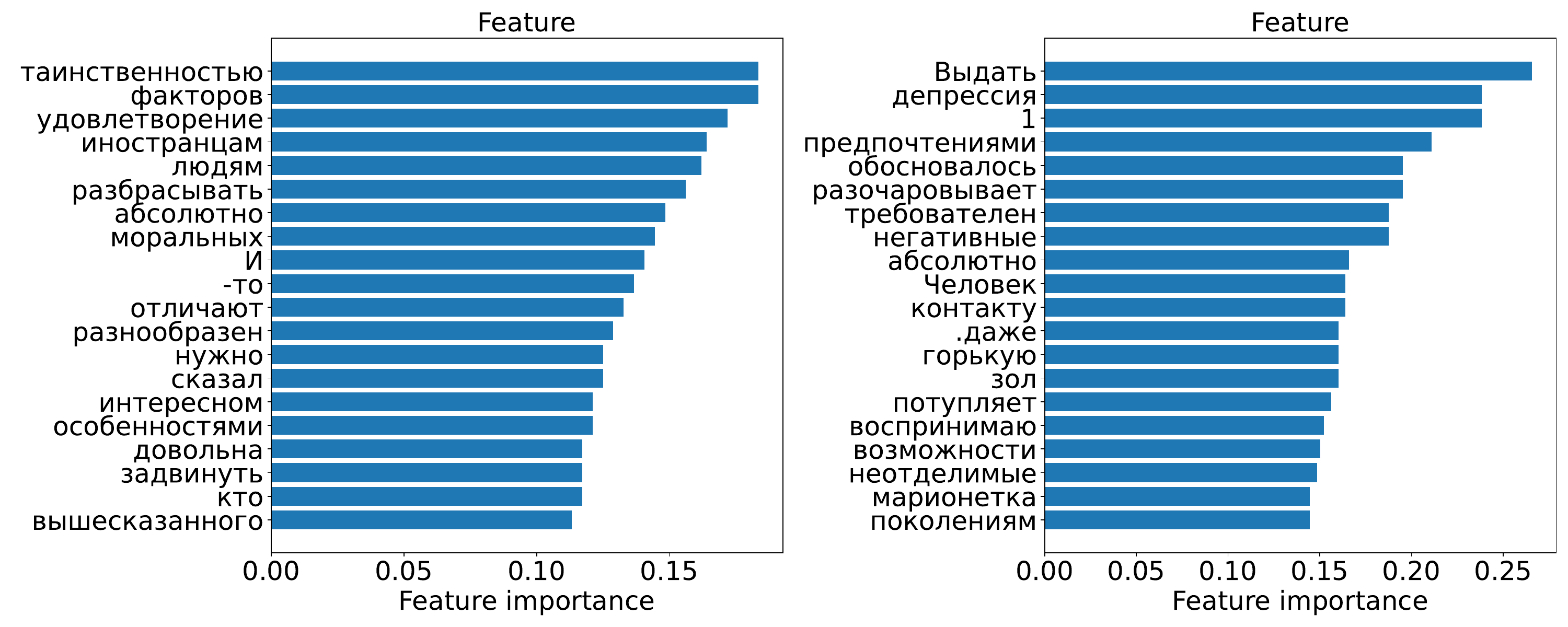}
        %\caption{The most important features in Russian.}
        \label{fig:top_n_res_tok_0_1_20_ru}
    \end{subfigure}

    \begin{subfigure}
        \centering
        \includegraphics[scale=0.3]{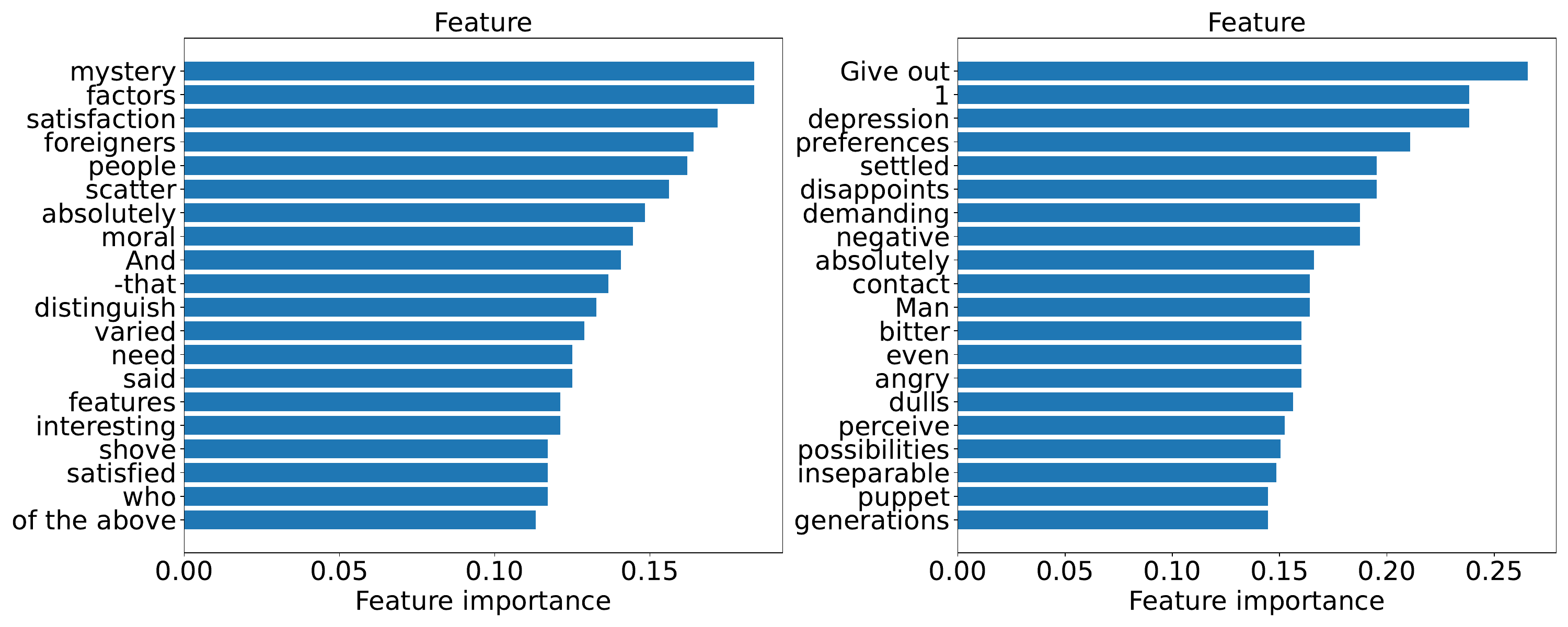}
        %\caption{The most important features in English (translated).}
        \label{fig:top_n_res_tok_0_1_20_en}
    \end{subfigure}
    \caption{The most important features for the target label generation from the test set of the DE dataset, for the predicted healthy class (left) and pathology class (right). The features are given in Russian (up) and translated into English (down).}
    \label{fig:top_n_res_tok_0_1_20}
    %\vspace{0.3cm}
\end{figure*}

\begin{figure*}[h]
    \centering
    \begin{subfigure}
        \centering
        \includegraphics[scale=0.3]{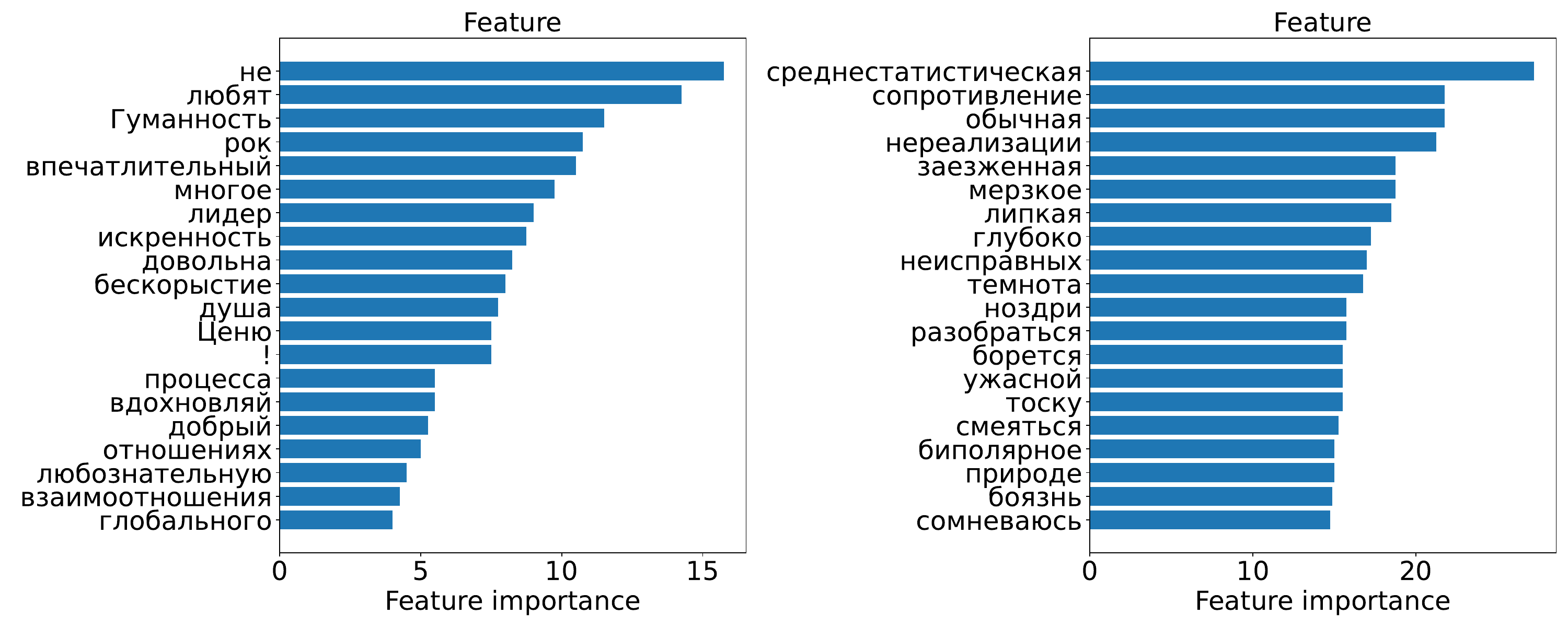}
        %\caption{The most important features in Russian.}
        \label{fig:top_n_res_seq_0_1_20_ru}
    \end{subfigure}

    \begin{subfigure}
        \centering
        \includegraphics[scale=0.3]{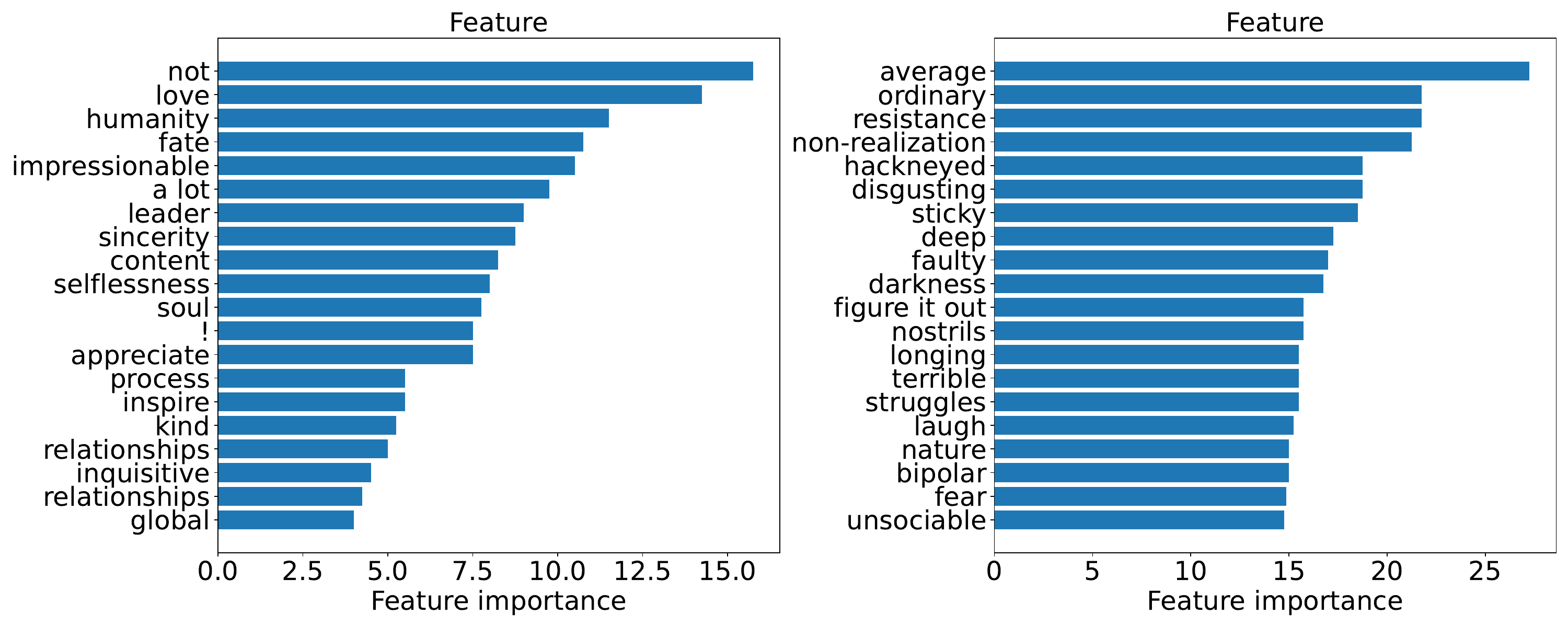}
        %\caption{The most important features in English (translated).}
        \label{fig:top_n_res_seq_0_1_20_en}
    \end{subfigure}
    \caption{The most important features for the explanation generation from the test set of the DE dataset, for the predicted healthy class (left) and pathology class (right). The features are given in Russian (up) and translated into English (down).}
    \label{fig:top_n_res_seq_0_1_20}
    %\vspace{0.3cm}
\end{figure*}

To further deepen the explanation analysis, we conducted feature ablation for the best LLM on the DE dataset. For this purpose, we extracted the top-3 words by their importance on target label generation (and on full explanation generation) for each text in the test set. The words were extracted with the Feature Ablation method from the Captum framework \cite{kokhlikyan2020captum}, which calculates feature ablation based on differences in predictions with and without features. The list of unique words with the biggest mean feature importance is shown at \Cref{fig:top_n_res_tok_0_1_20} (importance for target label generation) and \Cref{fig:top_n_res_seq_0_1_20} (importance for full explanation generation). For both target class and explanation generation, the most important features in texts with predicted pathology class contain a significant amount of words with negative meanings (such as "disappointing", "negative", "bitter", "angry", "disgusting", "darkness", etc.), as well as words with direct pathology description ("depression", "fear"). The detailed analysis from the trained clinicians reveals that most of the features with high importance for depression class are connected with fear, suffering, and unhealthy conditions. On the other hand, the most important features of a healthy class contain mostly positive semantics, such as "humanity", "unselfishness", and "kind".

\section{Disaggregated Results for DSM Dataset}
\label{app:dsm_splits}
The scores for the Depression-Social Media (DSM) dataset were aggregated based on the obtained BDI questionnaire scores. To further investigate how these results vary between subgroups with various scores within one group (inside sub-splits of healthy or pathology class), we provided \Cref{tab:results_dsm_healthy_splits,tab:results_dsm_pathology_splits} with disaggregated scores. As one can see, for the healthy class, the F1-healthy scores are mostly consistent for all models, with the exception of RuBioRoBERTa, which has a significantly lower F1-healthy score for the split with BDI Score from 3 to 6. For pathology class, F1-pathology scores vary more significantly, which can be described with a bigger variability of initial scores, especially for the split with scores from 34 to 63.
\begin{table*}[t]
\centering
\resizebox{\linewidth}{!}{%
    \begin{tabular}{lllllllllll}
    \hline
    \multirow{2}{*}{\textbf{Mode}} & \multirow{2}{*}{\textbf{Model}} & \multicolumn{3}{c}{\textbf{BDI Score 0-3}} & \multicolumn{3}{c}{\textbf{BDI Score 3-6}} & \multicolumn{3}{c}{\textbf{BDI Score 7-10}} \\
    \cline{3-11}
     &  & \begin{tabular}[c]{@{}l@{}} \textbf{Precision} \\ \textbf{healthy}\end{tabular} & \begin{tabular}[c]{@{}l@{}} \textbf{Recall} \\ \textbf{healthy}\end{tabular} & \textbf{F1-healthy} & \begin{tabular}[c]{@{}l@{}} \textbf{Precision} \\ \textbf{healthy}\end{tabular} & \begin{tabular}[c]{@{}l@{}} \textbf{Recall} \\ \textbf{healthy}\end{tabular} & \textbf{F1-healthy} & \begin{tabular}[c]{@{}l@{}} \textbf{Precision} \\ \textbf{healthy}\end{tabular} & \begin{tabular}[c]{@{}l@{}} \textbf{Recall} \\ \textbf{healthy}\end{tabular} & \textbf{F1-healthy} \\
    \hline
    5-shot & SaigaLlama3 8B & 0.00\tiny{$\pm$ 0.00} & 0.00\tiny{$\pm$ 0.00} & 0.00\tiny{$\pm$ 0.00} & 0.00\tiny{$\pm$ 0.00} & 0.00\tiny{$\pm$ 0.00} & 0.00\tiny{$\pm$ 0.00} & 100.00\tiny{$\pm$ 0.00} & 6.25\tiny{$\pm$ 0.00} & 11.76\tiny{$\pm$ 0.00} \\
5-shot & Vikhr 7B IT 0.4 & 100.00\tiny{$\pm$ 0.00} & 83.33\tiny{$\pm$ 0.00} & 90.91\tiny{$\pm$ 0.00} & 100.00\tiny{$\pm$ 0.00} & 80.00\tiny{$\pm$ 0.00} & 88.89\tiny{$\pm$ 0.00} & 100.00\tiny{$\pm$ 0.00} & 81.25\tiny{$\pm$ 0.00} & 89.66\tiny{$\pm$ 0.00} \\
5-shot MMLU & SaigaLlama3 8B & 0.00\tiny{$\pm$ 0.00} & 0.00\tiny{$\pm$ 0.00} & 0.00\tiny{$\pm$ 0.00} & 0.00\tiny{$\pm$ 0.00} & 0.00\tiny{$\pm$ 0.00} & 0.00\tiny{$\pm$ 0.00} & 100.00\tiny{$\pm$ 0.00} & 6.25\tiny{$\pm$ 0.00} & 11.76\tiny{$\pm$ 0.00} \\
5-shot MMLU & Vikhr 7B IT 5.4 & 100.00\tiny{$\pm$ 0.00} & 83.33\tiny{$\pm$ 0.00} & 90.91\tiny{$\pm$ 0.00} & 100.00\tiny{$\pm$ 0.00} & 80.00\tiny{$\pm$ 0.00} & 88.89\tiny{$\pm$ 0.00} & 100.00\tiny{$\pm$ 0.00} & 81.25\tiny{$\pm$ 0.00} & 89.66\tiny{$\pm$ 0.00} \\
LoRA & Qwen2 7B IT & 100.00\tiny{$\pm$ 0.00} & 80.56\tiny{$\pm$ 11.45} & 88.79\tiny{$\pm$ 7.00} & 100.00\tiny{$\pm$ 0.00} & 83.33\tiny{$\pm$ 21.34} & 89.15\tiny{$\pm$ 15.15} & 100.00\tiny{$\pm$ 0.00} & 65.62\tiny{$\pm$ 10.05} & 78.80\tiny{$\pm$ 7.34} \\
SFT & RuBioRoBERTa & 100.00\tiny{$\pm$ 0.00} & 86.11\tiny{$\pm$ 11.45} & 92.12\tiny{$\pm$ 6.78} & 100.00\tiny{$\pm$ 0.00} & 50.00\tiny{$\pm$ 10.00} & 66.07\tiny{$\pm$ 8.93} & 100.00\tiny{$\pm$ 0.00} & 58.33\tiny{$\pm$ 4.66} & 73.57\tiny{$\pm$ 3.79} \\
    \hline
    \end{tabular}
    }
    
\caption{Results of best models for DSM dataset for various sub-splits for healthy group (mean $\pm$ std).}
\label{tab:results_dsm_healthy_splits}
\end{table*}
\begin{table*}[t]
\centering
\resizebox{\linewidth}{!}{%
    \begin{tabular}{llllllll}
    \hline
    \multirow{2}{*}{\textbf{Mode}} & \multirow{2}{*}{\textbf{Model}} & \multicolumn{3}{c}{\textbf{BDI Score 30-34}} & \multicolumn{3}{c}{\textbf{BDI Score 34-63}} \\
    \cline{3-8}
     &  & \begin{tabular}[c]{@{}l@{}} \textbf{Precision} \\ \textbf{pathology}\end{tabular} & \begin{tabular}[c]{@{}l@{}} \textbf{Recall} \\ \textbf{pathology}\end{tabular} & \textbf{F1-pathology} & \begin{tabular}[c]{@{}l@{}} \textbf{Precision} \\ \textbf{pathology}\end{tabular} & \begin{tabular}[c]{@{}l@{}} \textbf{Recall} \\ \textbf{pathology}\end{tabular} & \textbf{F1-pathology}  \\
    \hline
    5-shot & SaigaLlama3 8B & 100.00\tiny{$\pm$ 0.00} & 100.00\tiny{$\pm$ 0.00} & 100.00\tiny{$\pm$ 0.00} & 100.00\tiny{$\pm$ 0.00} & 100.00\tiny{$\pm$ 0.00} & 100.00\tiny{$\pm$ 0.00} \\
5-shot & Vikhr 7B IT 0.4 & 100.00\tiny{$\pm$ 0.00} & 52.00\tiny{$\pm$ 4.00} & 68.33\tiny{$\pm$ 3.33} & 100.00\tiny{$\pm$ 0.00} & 37.50\tiny{$\pm$ 0.00} & 54.55\tiny{$\pm$ 0.00} \\
5-shot MMLU & SaigaLlama3 8B & 100.00\tiny{$\pm$ 0.00} & 100.00\tiny{$\pm$ 0.00} & 100.00\tiny{$\pm$ 0.00} & 100.00\tiny{$\pm$ 0.00} & 100.00\tiny{$\pm$ 0.00} & 100.00\tiny{$\pm$ 0.00} \\
5-shot MMLU & Vikhr 7B IT 5.4 & 100.00\tiny{$\pm$ 0.00} & 50.00\tiny{$\pm$ 0.00} & 66.67\tiny{$\pm$ 0.00} & 100.00\tiny{$\pm$ 0.00} & 37.50\tiny{$\pm$ 0.00} & 54.55\tiny{$\pm$ 0.00} \\
LoRA & Qwen2 7B IT & 100.00\tiny{$\pm$ 0.00} & 58.33\tiny{$\pm$ 14.62} & 72.59\tiny{$\pm$ 11.90} & 100.00\tiny{$\pm$ 0.00} & 41.67\tiny{$\pm$ 11.79} & 57.87\tiny{$\pm$ 11.50} \\
SFT & RuBioRoBERTa & 100.00\tiny{$\pm$ 0.00} & 55.00\tiny{$\pm$ 5.00} & 70.83\tiny{$\pm$ 4.17} & 100.00\tiny{$\pm$ 0.00} & 27.08\tiny{$\pm$ 4.66} & 42.42\tiny{$\pm$ 5.42} \\
    \hline
    \end{tabular}
    }
    
\caption{Results of best models for DSM dataset for various sub-splits for pathology group (mean $\pm$ std).}
\label{tab:results_dsm_pathology_splits}
\end{table*}

\section{Textual Examples}
\label{app:de_examples}
For a better understanding of the investigated task, we presented several samples from the DE dataset - one per class. The samples were paraphrased and anonymized, and then translated into English. The examples are presented in \Cref{tab:de_examples}.
\begin{table*}
\centering
\resizebox{1.0\linewidth}{!}{
\begin{tabular}{lp{16cm}}
\toprule
\textbf{Label} &   \textbf{Text (In English)} \\
\midrule
 Pathology &  Hello! My name is NAME, I am already AGE years old. I live with my family: my husband and our wonderful daughter. My profession is a midwife in a women's clinic, which I truly love. The team in our department is small, but we are all very close to each other. If any problems arise, we solve them very quickly. Until recently, I was an active person, with a large circle of friends who delighted me with various entertainments: going to the theater, watching movies, walking around the city or just meeting for a cup of coffee. Communicating with them brought me great pleasure. However, recently I began to distance myself from everyone, both at work and in my personal environment. This was the result of my fear, which turned out to be much more powerful than I expected. Fear takes away not only emotional strength, but also physical. At present, it seems to me that it is better to remain alone, avoiding meetings with other people. Although this makes my life less bright, I cannot cope with this condition yet. Also, I used to be into bead embroidery, but now I can't even bring myself to do a simple task. Luckily, we have our cute house cat, who helps me cope with stress and is a great antidepressant. This is my current lifestyle. \\

\hline
 Healthy & When I was a child, I accidentally found out that there were people who didn’t like me. They were my classmates, and I was very upset when I tried to be friends with them, but my attempt only increased the hostility. I tried to attract their attention with gifts, invitations to visit and other ways, but each new attempt ended in failure and even more trouble. Then I realized that no matter how you behave, there will always be people who don’t like you, sometimes for no apparent reason. But it is important to understand that this is normal and you shouldn’t try to live up to the expectations of others. You should value yourself for who you are. Many years have passed since then, and now I have many friends and acquaintances, some leave, and new ones come. Each person is unique and beautiful in their own way. I especially liked the expression: “Each person is a small cosmos.” It is really profound. Inside each person there is a whole universe consisting of his life experience, mistakes, disappointments, joys, defeats and small victories. If a person allows others to open up, it can be incredibly beautiful and exciting. In a world where we encounter many people every day, I want to believe that everyone respects each other, despite the fact that everyone may not like them. Even those we find unattractive may be dear to someone else. So there is no need to worry about not being loved, because we can find a common language and fill each other with love for life. We are all small universes living together in one big world, trying to get to know each other every day. \\

\bottomrule
\end{tabular}
}\caption{Paraphrased and anonymized examples from DE dataset.}
\label{tab:de_examples}
\end{table*}

\end{document}